\documentclass{cta-author}

\usepackage{stackengine}
\usepackage{paralist}

\usepackage{subfig}
\usepackage{rotating}
\usepackage{multirow}
\usepackage[table]{xcolor}
\usepackage{enumerate}
\usepackage[normalem]{ulem}
\usepackage{gensymb}

\usepackage{amsmath,amssymb,latexsym}
\usepackage{color}
\usepackage{graphicx,array,multirow,algorithm2e}
\usepackage{pgffor}
\usepackage[utf8]{inputenc}
\usepackage{pdfpages}
\usepackage[acronym,toc]{glossaries}
\newacronym{oam}{OAM}{Object-Awareness Model}

\newacronym{csd}{CSD}{Comprehensive Salient Object Detection}

\newacronym{cos}{COS}{Comprehensive Object Saliency}

\newacronym{svr}{SVR}{Support Vector Regression}

\newacronym{mst}{MST}{Minimum Spanning Tree}

\newacronym{hvs}{HVS}{Human Visual System}

\newacronym{FT}{FT}{Frequency-Tuned~\cite{Achanta_frequency_tuned}}

\newacronym{HC}{HC}{Histogram-based Contrast~\cite{Cheng_global_contrast}}

\newacronym{RC}{RC}{Region-based Contrast~\cite{Cheng_global_contrast}}

\newacronym{LR}{LR}{Low-Rank matrix recovery~\cite{Shen_a_unified}}

\newacronym{SF}{SF}{Saliency Filters~\cite{Perazzi_saliency_filters}}

\newacronym{GC}{GC}{Global Cues~\cite{Cheng_efficient_salient}}

\newacronym{AMC}{AMC}{Absorbing Markov Chain~\cite{Jiang_saliency_detection}}

\newacronym{GMR}{GMR}{Graph-based Manifold Ranking~\cite{Yang_saliency_detection}}

\newacronym{HSD}{HSD}{Hierarchical Saliency Detection~\cite{Yan_hierarchical_saliency}}

\newacronym{CH}{CH}{Contextual Hyperhraph~\cite{Li_contextual_hypergraph}}

\newacronym{HR}{HR}{Hierarchical Regression~\cite{Yildirim_saliency_detection}}

\newacronym{FASA}{FASA}{Fast, Accurate, and Size-Aware~\cite{Yildirim_fast_accurate}}

\newacronym{MSF}{MSF}{Multi-Scale Filtering}

\newacronym{MSRA-1000}{MSRA-1000}{MSRA-1000 Dataset~\cite{Achanta_frequency_tuned}}

\newacronym{SED-100}{SED-100}{Segmentation Evaluation Dataset~\cite{Alpert_image_segmentation}}

\newacronym{SOD}{SOD}{Salient Object Dataset~\cite{Movahedi_design_and}}

\newacronym{MSRA}{MSRA}{Microsoft Research Asia Dataset~\cite{Liu_learning_to}}

\newacronym{BSD}{BSD}{Berkeley Segmentation Dataset~\cite{Martin_a_database}}

\newglossaryentry{foveaSize}
{
    name=\ensuremath{\alpha},
    description={Half of the human fovea size in degrees},
    sort=alpha
}

\newglossaryentry{etAccuracy}
{
    name=\ensuremath{\eta},
    description={Accuracy of the eye-tracking equipment in degrees},
    sort=eta
}

\newglossaryentry{subject}
{
    name=\ensuremath{\delta},
    description={Refers to a single subject in a subjective experiment},
    sort=delta
}

\newglossaryentry{experimentalData}
{
    name=\ensuremath{\mathbf{D}},
    description={The data that is obtained via one of the subjective experiments: eye-tracking, point-clicking, or rectangle-drawing},
    sort=data
}

\newglossaryentry{betaSquared}
{
    name=\ensuremath{\beta ^2},
    description={A scalar value that balances the effect of precision and recall values on F-Measure},
    sort=betaSquared
}

\newglossaryentry{mask}
{
    name=\ensuremath{\mathbf{\Gamma}},
    description={A binary masking image, which usually indicate the pixels of an object, a surrounding area, or background},
    sort=mask
}

\newglossaryentry{colorDifference}
{
    name=\ensuremath{\Delta E^*},
    description={Euclidean distance between two CIELa*b* color vectors},
    sort=deltaE
}

\newglossaryentry{saliencyValue}
{
    name=\ensuremath{s},
    description={Saliency value of a pixel or a superpixel},
    sort=tree
}

\newglossaryentry{MST}
{
    name=\ensuremath{\mathbf{T}},
    description={Minimum Spanning Tree of a graph},
    sort=tree
}

\newglossaryentry{image}
{
    name=\ensuremath{\mathbf{I}},
    description={A matrix with either one channel (grayscale) or three channels (color), which represents a digital image},
    sort=image
}

\newglossaryentry{xcoor}
{
    name=\ensuremath{x},
    description={Horizontal coordinate of a matrix or an image},
    sort=x
}

\newglossaryentry{ycoor}
{
    name=\ensuremath{y},
    description={Vertical coordinate of a matrix or an image},
    sort=y
}

\newglossaryentry{salientObjectMap}
{
    name=\ensuremath{\mathbf{S}},
    description={A salient object map},
    sort=salientObjectMap
}

\newglossaryentry{groundTruthMap}
{
    name=\ensuremath{\mathbf{G}},
    description={A ground truth map},
    sort=groundTruthMap
}

\newglossaryentry{MLgroundTruthMap}
{
    name=\ensuremath{\mathbf{M}},
    description={A multi-level ground truth map},
    sort=multilevelGroundTruthMap
}

\newglossaryentry{colorVector}
{
    name=\ensuremath{\mathbf{c}},
    description={Three dimensional vector that represents the color of a pixel or the average color of a superpixel},
    sort=colorVector
}

\newglossaryentry{positionVector}
{
    name=\ensuremath{\mathbf{p}},
    description={Two dimensional vector that represents the position of a pixel or the center of mass of a superpixel},
    sort=positionVector
}

\newglossaryentry{spatialVariance}
{
    name=\ensuremath{V},
    description={A scalar value that represents the spatial variance of the color of a pixel or the average color of a superpixel},
    sort=variance
}

\newglossaryentry{quantizedVector}
{
    name=\ensuremath{\mathbf{q}},
    description={Three dimensional vector that represents a quantized color},
    sort=quantizedVector
}

\newglossaryentry{histogramBin}
{
    name=\ensuremath{h},
    description={A scalar value that represents the number of elements in a histogram bin},
    sort=histogramBin
}

\newglossaryentry{spatialSigma}
{
    name=\ensuremath{\sigma_s},
    description={A scalar value that controls the area of effect of a Gaussian in spatial coordinates},
    sort=sigmaSpatial
}

\newglossaryentry{colorSigma}
{
    name=\ensuremath{\sigma_c},
    description={A scalar value that controls the effect of color contrast},
    sort=sigmaColor
}

\newglossaryentry{weight}
{
    name=\ensuremath{w},
    description={A scalar value that represents the weight of an edge between two nodes of a graph},
    sort=weight
}

\newglossaryentry{truePositive}
{
    name=\ensuremath{t_p},
    description={True positive: number of salient pixels that are estimated as salient},
    sort=truePositive
}

\newglossaryentry{falsePositive}
{
    name=\ensuremath{f_p},
    description={False positive: number of non-salient pixels that are estimated as salient},
    sort=falsePositive
}

\newglossaryentry{falseNegative}
{
    name=\ensuremath{f_n},
    description={False negative: number of salient pixels that are estimated as non-salient},
    sort=falseNegative
}

\newglossaryentry{coefficientOfDetermination}
{
    name=\ensuremath{\text{R}^2},
    description={Coefficient of determination between two sets of values},
    sort=rSquared
}

\newglossaryentry{superpixel}
{
    name=\ensuremath{\mathbf{\Phi}},
    description={A superpixel, which is a set similarly-colored image pixels with proximity},
    sort=phiSuperpixel
}

\newglossaryentry{segmentSaliency}
{
    name=\ensuremath{\gamma},
    description={A binary value that indicates whether corresponding image segment is considered salient or not},
    sort=gammaSaliency
}

\newglossaryentry{gaussianFilter}
{
    name=\ensuremath{F_G},
    description={Two-dimensional Gaussian Filter},
    sort=filterGaussian
}

\newglossaryentry{weakMap}
{
    name=\ensuremath{\mathbf{Z}},
    description={A weak salient object map that is later combined with other maps to obtain a final and more accurate salient object map},
    sort=ZWeakMap
}

\newglossaryentry{histogram}
{
    name=\ensuremath{\mathbf{h}},
    description={CIELa*b* color histogram of a superpixel},
    sort=histogram
}

\newglossaryentry{hogs}
{
    name=\ensuremath{\mathbf{\kappa}},
    description={Histograms of oriented gradients of a superpixel},
    sort=kappa
}

\newglossaryentry{compactness}
{
    name=\ensuremath{K},
    description={A scalar value that indicates the spatial compactness of a salient object map},
    sort=kkk
}

\newglossaryentry{prior}
{
    name=\ensuremath{\mathbf{\Theta}},
    description={A full-resolution map that applies and adaptive center prior to the salient object map},
    sort=prior
}

\newglossaryentry{saliencyProbability}
{
    name=\ensuremath{\Lambda},
    description={The probability of saliency of a pixel, quantized color, or superpixel},
    sort=lambda
}

\newglossaryentry{globalContrast}
{
    name=\ensuremath{\xi},
    description={Global color-contrast value of a pixel, quantized color, or superpixel},
    sort=xsi
}

\newglossaryentry{layer}
{
    name=\ensuremath{l},
    description={A single layer at a hierarchical segmentation},
    sort=layer
}

\begin{document}

\title{Evaluating Salient Object Detection in Natural Images with Multiple Objects having Multi-level Saliency}

\author{\au{G\"{o}khan~Yildirim$^{1}$}, \au{Debashis~Sen$^{2*}$},  \au{Mohan~Kankanhalli$^{3}$}, and \au{Sabine~S\"{u}sstrunk$^{4}$}}

\address{\add{1}{Zalando Research, Berlin, Germany}
	\add{2}{Department of Electronics and Electrical Communication Engineering, Indian Institute of Technology, Kharagpur, India}
	\add{3}{School of Computing, National University of Singapore, Singapore}
	\add{4}{School of Computer and Communication Sciences, \'Ecole Polytechnique F\'ed\'erale de Lausanne, Switzerland}
	\email{dsen@ece.iitkgp.ac.in}}

\begin{abstract}
{Salient object detection is evaluated using binary ground truth with the labels being salient object class and background. In this paper, we corroborate based on three subjective experiments on a novel image dataset that objects in natural images are inherently perceived to have varying levels of importance. Our dataset, named SalMoN (saliency in multi-object natural images), has 588 images containing multiple objects. The subjective experiments performed record spontaneous attention and perception through eye fixation duration, point clicking and rectangle drawing. As object saliency in a multi-object image is inherently multi-level, we propose that salient object detection must be evaluated for the capability to detect all multi-level salient objects apart from the salient object class detection capability. For this purpose, we generate multi-level maps as ground truth corresponding to all the dataset images using the results of the subjective experiments, with the labels being multi-level salient objects and background. We then propose the use of mean absolute error, Kendall’s rank correlation and average area under precision-recall curve to evaluate existing salient object detection methods on our multi-level saliency ground truth dataset. Approaches that represent saliency detection on images as local-global hierarchical processing of a graph perform well in our dataset.}
\end{abstract}

\maketitle


\section{Introduction}
\label{sec:introduction}
Visual saliency signifying the importance of an entity in an image is postulated to guide the deployment of human visual attention \cite{Baluch11}, where both top-down and bottom-up processing are involved. As only the attended entities drive image understanding \cite{Baluch11,Neibur98}, saliency computation acts as a strategy used by humans to judiciously use the limited resources available by considering only pertinent visual sensory data \cite{Neibur98}. Salient object detection uses saliency to perform object detection in images, where objects are detected as important entities as opposed to the background. In recent studies, salient object detection has been evaluated for performance using binary images as ground truth, testing their capability to detect a single salient object class against the background \cite{Han2018}. {As shown in the examples of Figures~\ref{fig:datasetComparisons}\subref{fig:SOD} and \subref{fig:MSRA}, each image in the well-known salient object detection evaluation datasets MSRA-1000~\cite{Achanta_frequency_tuned}, SED-100~\cite{Alpert_image_segmentation}, SOD~\cite{Movahedi_design_and}, PASCAL-S~\cite{Sal_secret}, SOC \cite{fan2018SOC}, and DUT-OMRON~\cite{Yang_saliency_detection} has a corresponding binary ground-truth image /map to be used for evaluation.} For an image with multiple objects, a binary ground truth image for evaluation assumes that all the objects in the image are either of equal importance or some are not salient at all. The important objects fall in the salient object class with others falling in the background class. Note that the works proposing the SOD and PASCAL-S datasets do consider unequal importance of objects in their dataset creation, but do not use it for ground truth generation and evaluation. Assuming equal importance of salient objects, the use saliency for object detection has progressed to object-level abstraction {\cite{Zhou2019}} and video salient object detection {\cite{Chen2018}}.

{Firstly, in this paper, with the help of subjective experiments we corroborate the findings of \cite{Movahedi_design_and, Sal_secret} that objects in natural images are seen and perceived to have varying levels of importance.} The saliency of objects in images with multiple objects is found to be multi-level through a study of spontaneous human visual attention on and visual perception of objects.

{\it Spontaneous Human Attention through Eye Fixation Duration: \quad} According to Henderson et al.~\cite{Henderson_eye_movements} and Yarbus~\cite{Yarbus_eye_movements}, a person spontaneously attends to visually distinctive image regions by fixating on them while free-viewing, and this spontaneous attention is mainly achieved by bottom-up processing involving distinctness computation. The fixation duration (when perception happens), which may differ from person to person, is usually longer for regions that are more informative \cite{Henderson_eye_movements, Yarbus_eye_movements}. Therefore, recording eye fixations at a constant rate yields more fixations at higher informative regions indicating higher importance/saliency. Different fixation durations represented by different fixation densities at different objects shows that object saliency is inherently multi-level. 

{\it Human Perception through Point Clicking and Rectangle Drawing: \quad} While spontaneous attention driven by visual distinction is required for perceiving (during fixation) the information presented by an image region, activities such as point clicking and rectangle drawing are usually performed after perception. Hence, when a person clicks on or draws a rectangle around an image region by attending to them upon asked to do so for important regions, we may assume that the person considers the region to be salient after perception. This consideration of a region to be salient, which guides the attention that leads to image understanding, involves both bottom-up and top-down processing \cite{Baluch11}. Variation in the number of persons who consider different objects as salient shows that object saliency is also inherently multi-level after perception. Note that perception recorded through point clicking and rectangle drawing differ as one is related to marking the location of a salient region while the other requires consideration of a salient region's size and shape.

We created a dataset of 588 images having multiple objects per image to perform the said study of spontaneous attention on and perception of objects, and the three subjective experiments were performed as follows:
\begin{compactitem}
	\item [\it Eye-tracking experiments-]$\ \ $Subjects were asked to free-view the images on a monitor and their eye fixations were recorded using an eye tracker. Here we consider the well-founded assumption that free-viewing images for a short period captures eye movements due to spontaneous attentional shifts~\cite{Peters05}. An example eye fixation density map collected from multiple subjects on a single image is shown in Figure~\ref{fig:multimodalGroundTruth}\subref{fig:eyeFixation}. {In total, 95 people participated in this experiment, where each image is viewed by 24 subjects on an average (min: 13, max: 34).} 
	
	\item [\it Point-clicking experiments-]$\ \ $Subjects recruited through crowd sourcing were asked to view the images and click on the important objects that they notice at first glance. A set of clicked points collected from multiple subjects on a single image is shown in Figure~\ref{fig:multimodalGroundTruth}\subref{fig:pointClicking}. The ``first glance'' phrase was used to avoid multiple perceptual processing (by a subject) of a single image entity. {In this experiment,  each image on an average is viewed by 33 subjects (min: 24, max: 38).}
	
	\item [\it Rectangle-drawing experiments-]$\ \ $Subjects recruited through crowd sourcing were asked to view the images and draw rectangles around the important objects that they notice at first glance. A set of rectangles collected from multiple subjects on a single image is illustrated in Figure~\ref{fig:multimodalGroundTruth}\subref{fig:rectangleDrawing}. {In this experiment, each image on an average is viewed by 32 subjects (min: 15, max: 50).}
\end{compactitem}

\newcommand{\datasetImageWidth}{0.245\columnwidth}
\begin{figure}[t]
	\centering
	\subfloat[Multiple salient objects (MSO); Binary ground truth (BGT)]{\label{fig:SOD}\stackunder[2pt]{\includegraphics[width=0.13\columnwidth, keepaspectratio=true]{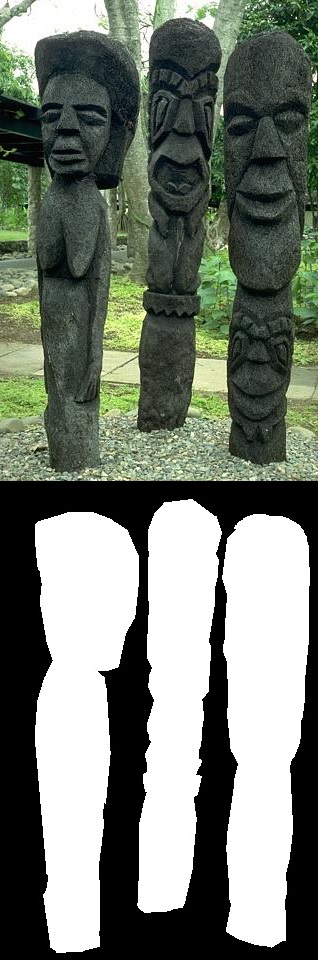}}{SOD}\stackunder[2pt]{\includegraphics[width=0.26\columnwidth, keepaspectratio=true]{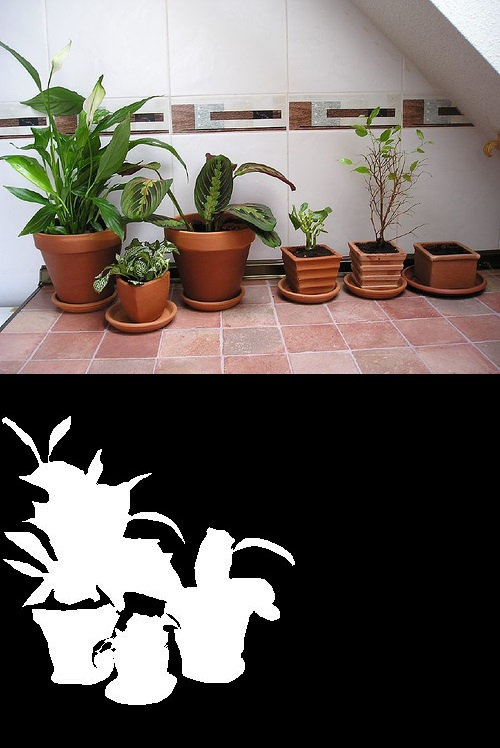}}{PASCAL-S}\stackunder[2pt]{\includegraphics[width=0.26\columnwidth, keepaspectratio=true]{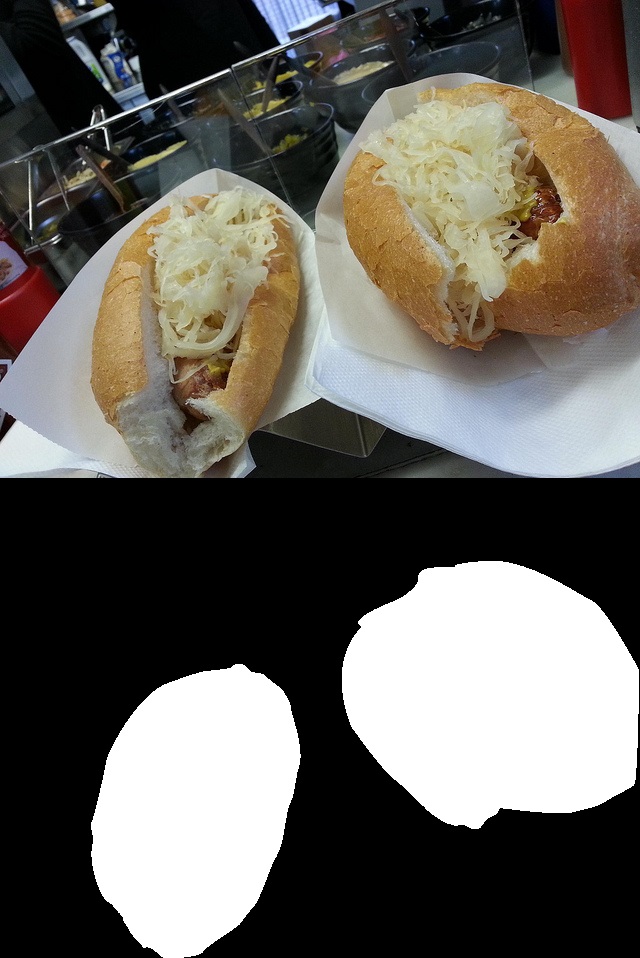}}{SOC}\stackunder[2pt]{\includegraphics[width=0.195\columnwidth, keepaspectratio=true]{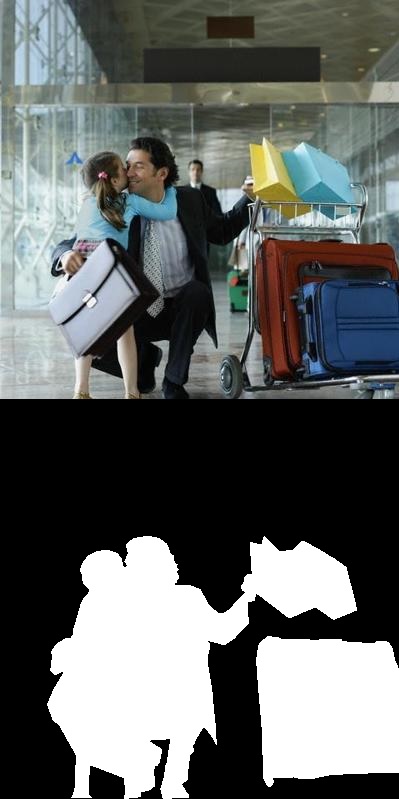}}{DUT-OMRON}}\\
	\subfloat[Single salient object; BGT]{\label{fig:MSRA}\stackunder[2pt]{\includegraphics[width=\datasetImageWidth, keepaspectratio=true]{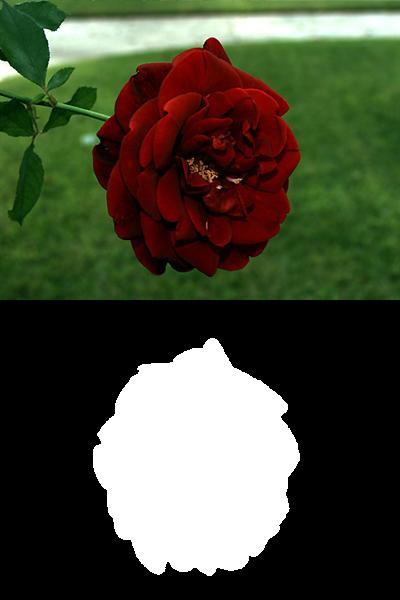}}{MSRA-1000}\stackunder[2pt]{\includegraphics[width=\datasetImageWidth, keepaspectratio=true]{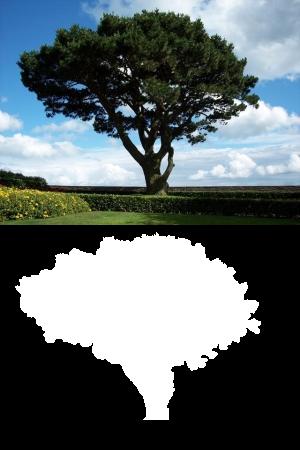}}{SED-100}}
	\subfloat[Ranked MSO]{\label{fig:AugPascalS}\stackunder[0.2pt]{\includegraphics[width=\datasetImageWidth, keepaspectratio=true]{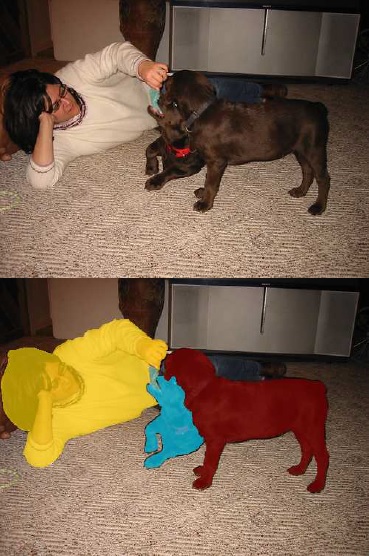}}{AugPASCAL-S}}
    \subfloat[Gray MSO]{\label{fig:OURS}\stackunder[2pt]{\includegraphics[width=\datasetImageWidth, keepaspectratio=true]{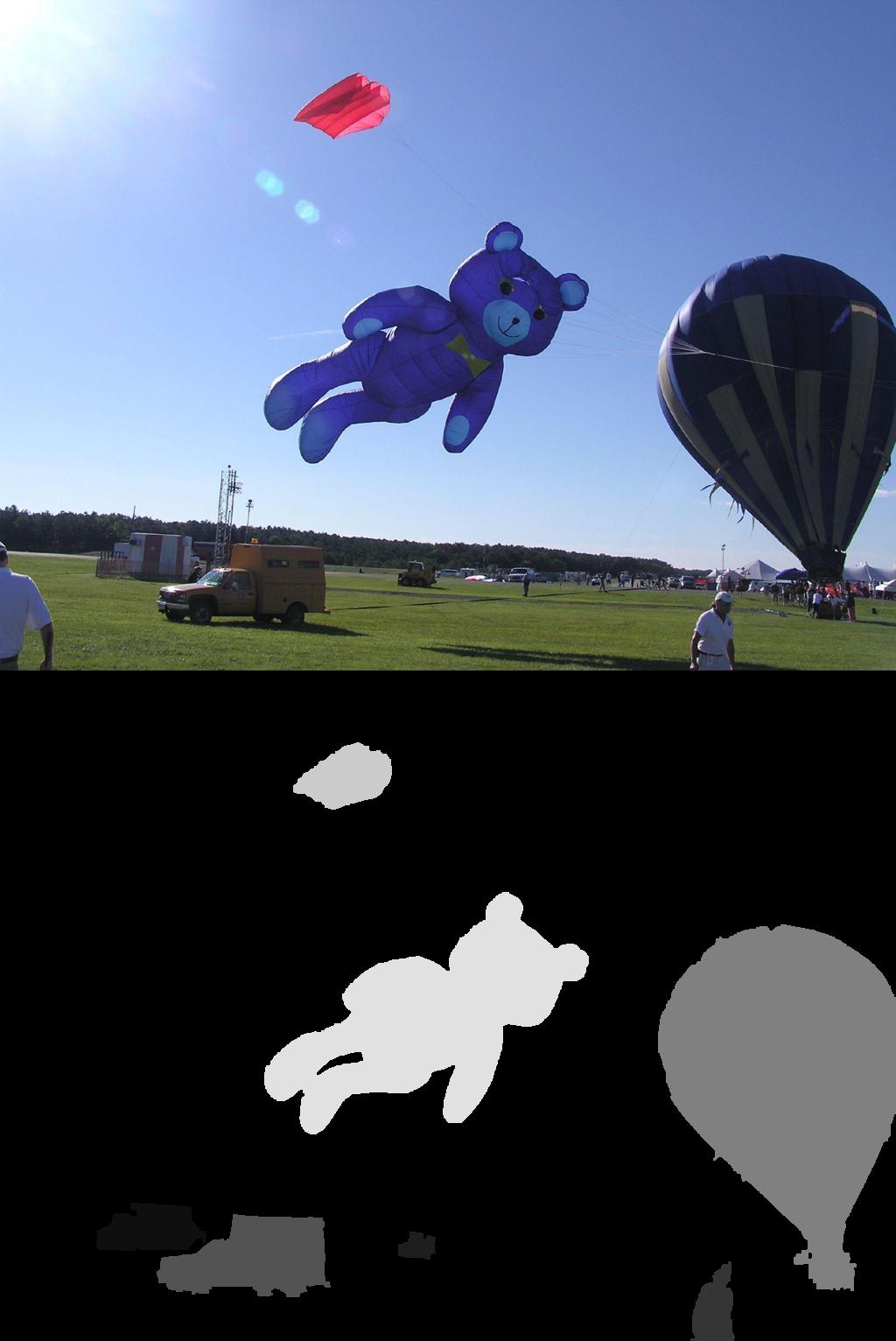}}{SalMoN}} 
	\caption{Sample images, and (a)-(b) binary ground-truth maps from six existing datasets, (c) binary ground-truth map along with saliency rank order from high to low as red, yellow and then blue from AugPASCAL-S dataset and (d) ground-truth maps in our dataset `SalMoN' are multi-level (gray), i.e.  saliency value of each object is between zero and one.}
	\label{fig:datasetComparisons}
\end{figure}

Secondly, in this paper, we propose that as saliencies of objects in images are inherently multi-level, a salient object detection method must be evaluated for its capability to detect all the multi-level salient objects considering their levels of saliency. For this purpose, we generate multi-level (gray-level) images as ground truth using the data collected from the three subjective experiments. The ground truth is generated for each of the 588 images of the new dataset, which we refer to as the SalMoN (saliency in multi-object natural images) dataset, indicating a dataset of natural images with multiple objects having multi-level saliency. As the study of spontaneous attention on and perception of objects is subjective in nature, it is imperative that the ground truth generation must consider the data obtained from many persons collectively to account for inter-person variations \cite{Wolfe_guided_search}. For every image, three multi-level ground truth images are generated corresponding to each subjective experiment with the {help of }manually marked object boundaries (see Figure~\ref{fig:multimodalGroundTruth0}\subref{fig:objectSegmentation}) in the images. The multi(gray)-level ground truth images for the fixation density in Figure~\ref{fig:multimodalGroundTruth}\subref{fig:eyeFixation}, the point-clicking in Figure~\ref{fig:multimodalGroundTruth}\subref{fig:pointClicking} and the rectangle drawing in Figure~\ref{fig:multimodalGroundTruth}\subref{fig:rectangleDrawing} are shown respectively in Figures~\ref{fig:multimodalGroundTruth}\subref{fig:fixationGT}, \subref{fig:pointGT} and \subref{fig:rectangleGT}. Note that the multi-level ground truths should not be confused with normalized fixation density maps used to evaluate saliency map generation algorithms for eye fixation prediction as in \cite{Judd_learning_to,Li_visual_saliency}. Unlike fixation density maps, our ground truths are for evaluating salient object detection results and not saliency map based eye fixation prediction results. Our salient object detection evaluation involves assessment of an approach's capability of giving appropriate object boundaries, unlike evaluation of saliency map based eye fixation prediction. Moreover, our ground truth is based on both human spontaneous attention and perception, unlike fixation density maps that considers only spontaneous attention.

Thirdly and finally, we propose simple yet effective performance evaluation approaches to evaluate detection of multiple objects with multi-level saliency, and consider them for evaluating a few (seventeen) well-known and latest methods on the natural images of our dataset. In this evaluation, we perform comparisons between the methods using the three multi-level ground truths {individually as well as collectively}. We propose specific ways of using them together such that certain biasing can be avoided. One important observation in the evaluation of salient object detection using our dataset is that the approaches found superior in the standard evaluation using existing datasets with binary ground truth maps may not be superior when applied on images with multiple multi-level salient objects.

The rest of the paper is organized as follows. In Section~\ref{sec:relatedWork}, we discuss the data collection procedures and the limitations of a few well-known saliency datasets. In Section~\ref{sec:dataCollection}, we explain our dataset, the subjective experiments, and object saliency measurement from the data collected through the experiments. In Section~\ref{sec:visualSaliencyOfObjects}, we discuss the inherent multi-level nature of object saliency and analyze the different kinds of saliencies measured from the different subjective experiments. In Section~\ref{sec:salientObjectDetectors}, we evaluate state-of-the-art salient-object detectors on our dataset, and in this process, propose suitable approaches to evaluate multi-level salient object detection. Section~\ref{sec:conclusion}, concludes our paper discussing the summary and future scope.

\newcommand{\teaserImageWidth}{4.3cm}

\begin{figure}[!h]
	\centering
	\subfloat[Original image]{\label{fig:multiObject}\includegraphics[width=\teaserImageWidth]{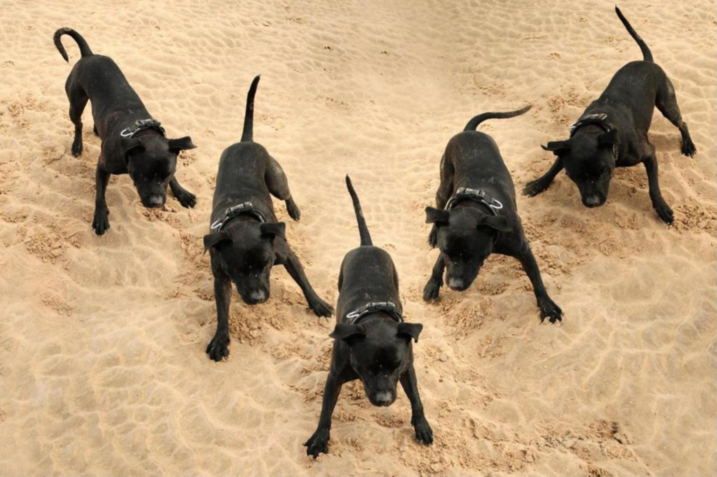}}
	\subfloat[Object segmentation]{\label{fig:objectSegmentation}\includegraphics[width=\teaserImageWidth]{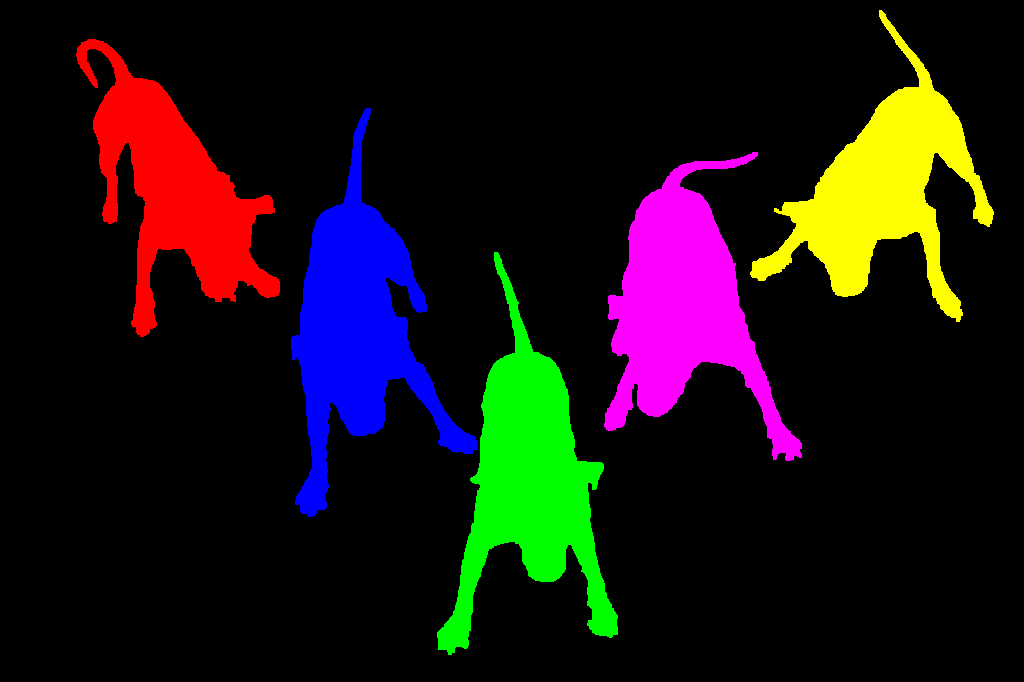}}
	\caption{For (a) each image in our dataset (b) object boundaries are manually marked.}
	\label{fig:multimodalGroundTruth0}
\end{figure}
\begin{figure}[!h]
	\subfloat[Eye-fixation density]{\label{fig:eyeFixation}\includegraphics[width=\teaserImageWidth]{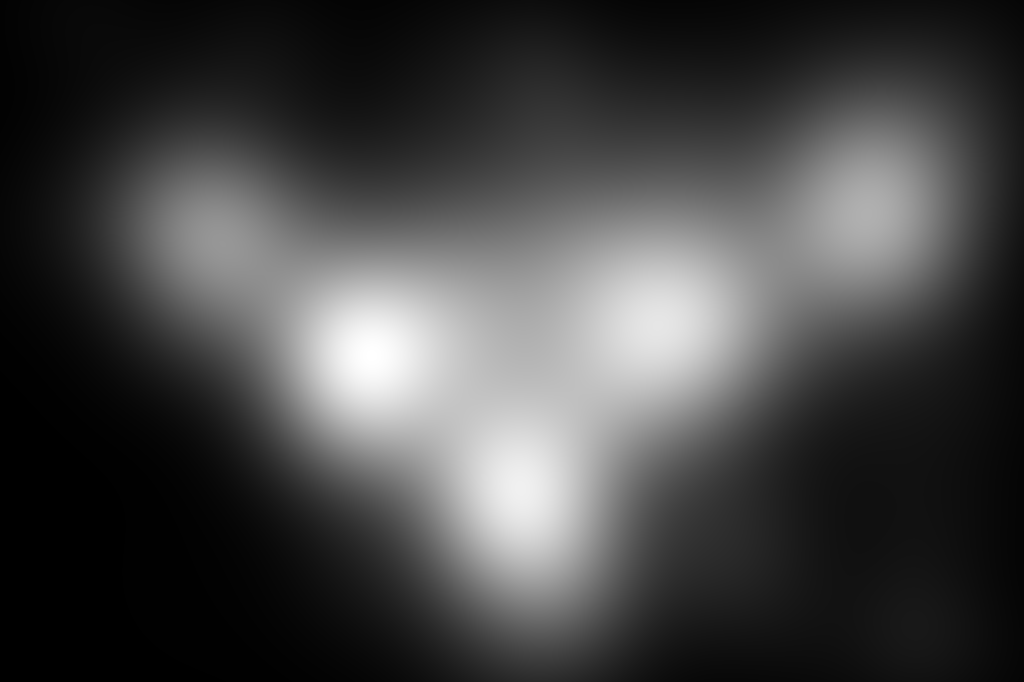}}
	\subfloat[Eye-tracking GT]{\label{fig:fixationGT}\includegraphics[width=\teaserImageWidth]{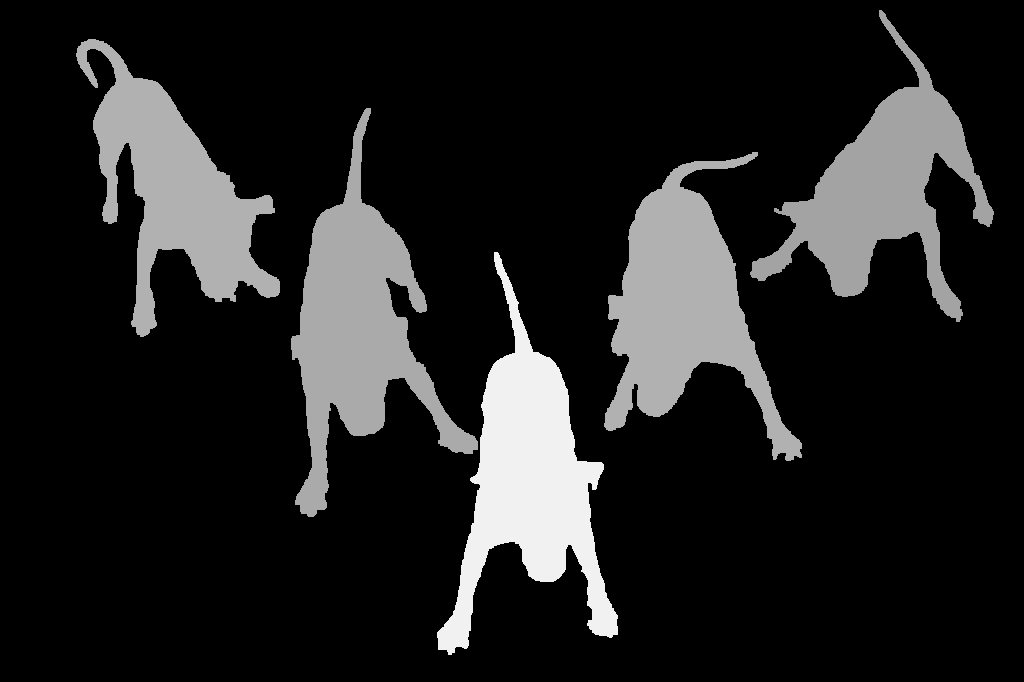}}\\
	\subfloat[Point clicking]{\label{fig:pointClicking}\includegraphics[width=\teaserImageWidth]{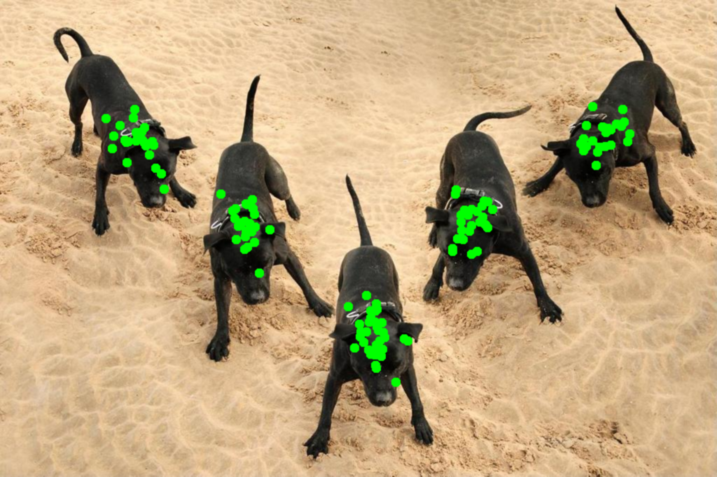}}
	\subfloat[Point-clicking GT]{\label{fig:pointGT}\includegraphics[width=\teaserImageWidth]{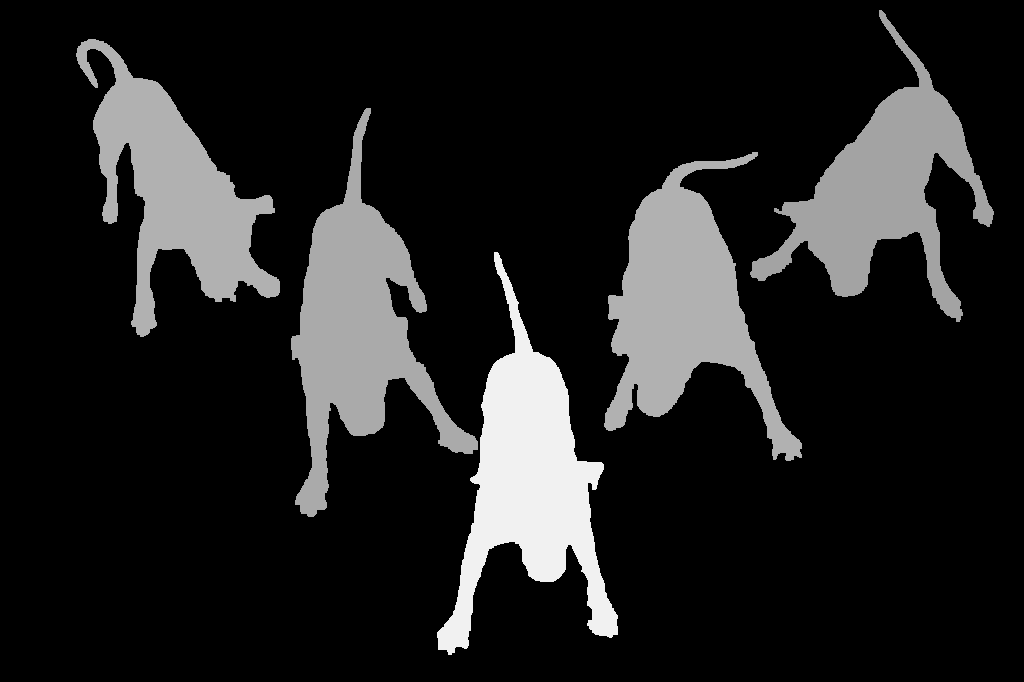}}\\
	\subfloat[Rectangle drawing]{\label{fig:rectangleDrawing}\includegraphics[width=\teaserImageWidth]{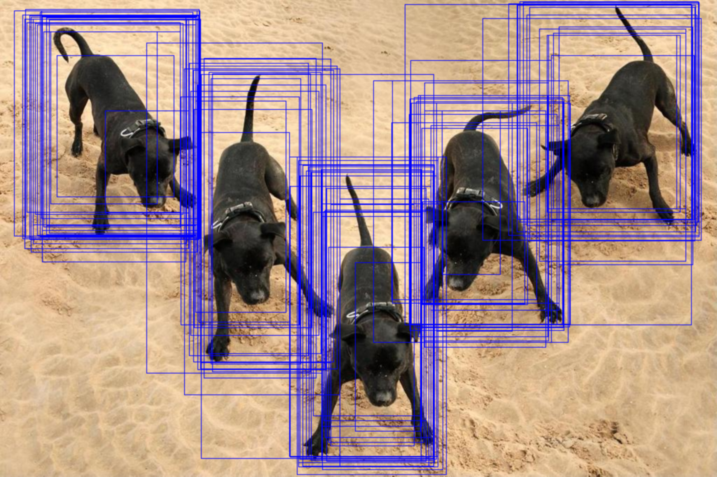}}
	\subfloat[Rectangle-drawing GT]{\label{fig:rectangleGT}\includegraphics[width=\teaserImageWidth]{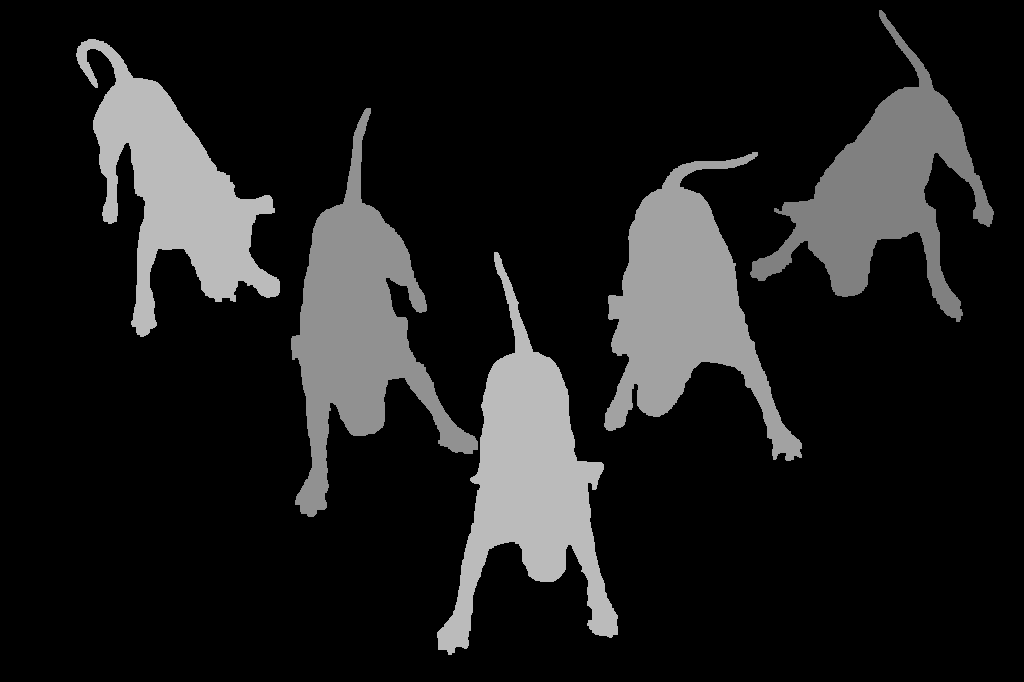}}
	\caption{{(a,c,e) Three types of experimental data are collected to generate (b,d,f) three different multi-level ground truth (GT) maps.}
	}
	\label{fig:multimodalGroundTruth}
\end{figure}

\section{Related Work}
\label{sec:relatedWork}
The existing literature related to our work is on dataset generation for evaluation of salient object detection. Although many datasets for salient object detection evaluation exist in literature, we consider the following widely used datasets,  MSRA-1000~\cite{Achanta_frequency_tuned}, SED-100~\cite{Alpert_image_segmentation}, SOD~\cite{Movahedi_design_and}, PASCAL-S~\cite{Sal_secret} and {DUT-OMRON \cite{Yang_saliency_detection, DUTOMRON}}, which have been used by some of the state-of-the-art salient object detectors among \cite{Achanta_frequency_tuned,Yildirim_fast_accurate,Cheng_global_contrast,Jiang_automatic_salient,Cheng_efficient_salient,Perazzi_saliency_filters,Shen_a_unified,Li_contextual_hypergraph,Jiang_saliency_detection,Yang_saliency_detection,Yildirim_saliency_detection,Zhang2015,Li2015,Peng2017,Huang2017}. {We also consider the datasets SOC \cite{fan2018SOC} and AugPASCAL-S \cite{islamsal18} that have been recently reported.} Even though we consider a sample of seven datasets from many existing ones, we do so without loss of generality in relation to the nature of ground truth maps provided and their use in evaluation. Here, we briefly explain the data collection and ground truth generation procedures of these existing datasets and then discuss their representation of multi-level object saliency in natural images with multiple objects.

\subsection{Data Collection and Ground Truth Generation}

The MSRA-1000 consists of 1000 natural images that are taken from the larger MSRA dataset \cite{Liu_learning_to}. The images in the original MSRA dataset \cite{Liu_learning_to} were specifically selected such that they have a single distinctive foreground object. Then, nine subjects were asked to draw a rectangle around the most salient object in each image. In the derived MSRA-1000 dataset, these rectangles are used only to identify a single salient object segmented manually by a single person to produce a binary ground truth image /map as shown in Figure~\ref{fig:datasetComparisons}\subref{fig:MSRA}.

Just like MSRA, all 100 images in the SED-100 dataset were selected such that they have a single clear foreground object. However, unlike MSRA-1000, rectangle drawing was not considered and the salient object in each image was directly segmented by three subjects. The segmentations of the subjects were combined into a binary ground truth image by considering a pixel to be salient when it is marked so by more than one subject. An example image and its ground truth are the ones given in Figure~\ref{fig:datasetComparisons}\subref{fig:MSRA}.

The SOD dataset was formed using the BSD dataset of \cite{Martin_a_database}. In BSD dataset, 300 images were oversegmented by three persons. That is, the objects and background are divided into multiple sub-parts. To create the SOD dataset, seven subjects were asked to identify salient object/s by combining BSD sub-segments. Unlike the other existing datasets, during the creation of SOD, subjects were required to rank the objects with respect to their saliency, if they detect more than one salient object. As evident, this salient object ranking is an attempt to represent multi-level object saliency in images with multiple objects. Although salient object ranking is available in SOD dataset, state-of-the-art salient object detection evaluation using it ignores the ranking and considers only a binary ground truth image with all salient objects in a single salient object class as shown in Figure~\ref{fig:datasetComparisons}\subref{fig:SOD}.

The PASCAL-S dataset was generated from the validation set of PASCAL 2010 dataset {\cite{VOC2010}} containing 850 images. First, all the objects in all the images were manually segmented to separate them from the background. Then 12 subjects were asked to click on the salient objects. They were free to take their time and click as many objects as salient as they wanted. After the experiment, each object in an image was assigned a saliency value given by the ratio of the number of clicks to the number of subjects. As evident, this assignment of different saliency values to different objects is nothing but representing multi-level object saliency in images with multiple objects. However, this multi-level object saliency was used for a purpose other than evaluating salient object detection performance. For salient object detection evaluation, the ground truth images proposed to be used were generated as binary maps obtained by thresholding the multi-level object saliency values. An example image and it's ground truth are the ones given in Figure~\ref{fig:datasetComparisons}\subref{fig:SOD}. {The available multi-level object saliency has been used in \cite{islamsal18} to augment the ground truth of PASCAL-S dataset with information about the relative ranks of the salient objects in an image. We refer to this augmented dataset as AugPASCAL-S dataset, an example of whose ground truth is shown in  Figure~\ref{fig:datasetComparisons}\subref{fig:SOD}. 

Pointing out that most datasets assume that images contain at least one salient object and have images that usually contain objects in low clutter, the SOC dataset was generated addressing the said issues. This dataset contains 6000 images which are categorized into 80 classes, where real-world conditions such as motion blur, occlusion, cluttered background and image with no salient object are present. 5 viewers were asked to mark salient objects using bounding boxes and annotate them. Only objects annotated by a majority of the viewers were considered salient, which yielded 3000 images with salient objects. Volunteers were then involved to mark precise boundaries (through accurate silhouetting) of the salient objects and validate the annotations. As evident, this dataset contains images with multiple salient objects which are individually annotated, and as shown in Figure~\ref{fig:datasetComparisons}\subref{fig:SOD} the ground truth images provided for salient object detection evaluation are binary maps segregating all the salient objects from those which are not and the image background. Note that for images with no salient object, the ground truth is all zeros.}

{The DUT-OMRON dataset consists of 5172 natural images that are taken from the SUN dataset \cite{SUND} through random selection followed by criteria-based pruning with one of the criteria being the presence of an apparent foreground. So the selected natural images contain at least one foreground object, with many of them having multiple. At least five operators are asked to mark bounding boxes on the salient objects in an image, and the final bounding box for a salient object is taken as the binary bounding mask obtained with a threshold of 0.5 on the average of the operators' binary bounding masks. An eye tracker was also employed and many operators' eye fixation data were collected on the images by showing them the images for 2 seconds each. Processing to remove outliers were carried out following which around 150 fixations were available per image on an average. A saliency map, which is a gray map taking values from $[0, 1]$, was generated from the fixations in an image applying Gaussian masks. Finally, a threshold of 0.1 was used on the saliency map to get a binary map. It was suggested that salient object detection may be evaluated on the two binary maps separately using precision-recall and receiver-operator characteristics curves.} 

\subsection{Representation of Multi-level Object Saliency}
\label{sec:limitations}

A discussion on the representation of multi-level object saliency in the seven existing datasets is given here. The binary ground truths generated in the MSRA-1000 and SED-100 datasets marked a single prominent object as salient in the images. When a single object is considered salient in an image, it helps us understand the visual cues of objects that affect their saliency. However, the notion of single object salient oversimplifies what salient object detection should be evaluated for, especially as natural images can be complex and can include multiple objects. Obviously, in natural images with multiple objects, the MSRA-1000 and SED-100 datasets do not attempt to represent multi-level object saliency.

During the creation of SOD dataset, salient object ranking was performed in images with multiple objects in order to represent multi-level object saliency. For evaluation, ground truth that represents multi-level object saliency is required. However, it is not trivial to generate such a ground truth from the salient object ranking. State-of-the-art salient object detection evaluation using SOD data considers only binary ground truth without involving multi-level object saliency. Salient object ranking does not contain information which can be used to quantify the difference between saliencies of any two objects, which makes the ground truth generation non-trivial. Moreover, there is no single appropriate way of combining the rankings given by the seven subjects. 

{Although the SOC dataset is a very useful one with a substantially large number of images where real-world scenarios are carefully captured and different salient objects in an image are annotated, the ground truth available for salient object detection evaluation is binary. This binarization is done while considering objects marked by a majority as salient, with an object's bounding box determined using a threshold on the IoU of the viewers' bounding boxes. Even though the information about the different categories of objects in an image is available through the annotation, mapping that information to multi-level saliency is not straightforward.}

The PASCAL-S dataset assigned a different saliency value to each object in an image in order to represent multi-level object saliency in images with multiple objects. With all objects in an image already manually segmented, this representation of multi-level object saliency can be used as ground truth for evaluating salient object detection, which unfortunately was not considered. Although the multi-level object saliency representation is readily available to be used as ground truth, certain aspects of it need to be noted. The only task performed by the volunteers to generate the representation was to click on salient objects. Such an approach to mark salient objects follows after the human perception of information at fixations. It is well known that saliency detection by humans involves both bottom-up and top-down processes~\cite{Baluch11}. As human perception involves cognitive entities such as memory, knowledge, and preference, the top-down process would dominate in the clicking on salient objects. Here, the lack of the bottom-up process's role diminishes the direct influence of the visual cues /features of objects in their saliency. The multi-level object saliency representation to be used as ground truth for salient object detection evaluation must relate to both bottom-up and top-down processes, as the interaction between the processes is substantially unknown~\cite{McMains11}. The bottom-up process is more objective in nature while the top-down process is more subjective, and hence, they represent different aspects of saliency detection by humans. In addition, as most state-of-the-art salient object detectors are heavily based on processing object's visual cues, consideration of bottom-up process will evaluate the approaches on what they aim to do and the inclusion of top-down process will help in evaluating what an approach must look to achieve further. Moreover, only 12 persons were used to create the PASCAL-S dataset and allied multi-level object saliency representation, which is substantially less number of samples to capture sufficient variations in subjective top-down influences. {The above discussion is also applicable for the AugPASCAL-S dataset, where saliency ranks are also part of the ground truth along with binary salient object maps, but not multi-level object saliency.} Referring back to the salient object ranking in SOD dataset, it must be noted that the ranking was also performed there in a manner where top-down process dominates and only 7 persons were involved.

{The DUT-OMRON dataset with a large number of images, is the only existing dataset which considers bounding box markings and eye fixations to generate ground truth for evaluating salient object detection. Bounding box marking and eye fixation can respectively capture top-down and bottom-up processes involved in saliency detection by humans. Hence, DUT-OMRON dataset is probably better suited for evaluation of salient object detection at different perceptual levels of humans than PASCAL-S dataset. However, similar to the PASCAL-S dataset, although inherently multi-level object saliency data were collected, they were binarized to yield binary ground truth maps for evaluating salient object detection, which is provided in the dataset. Further, although there are two different ground truth binary maps available related to each image obtained from the two subjective human experiments, no guidelines have been provided on performing a single evaluation using them. In \cite{Yang_saliency_detection}, only the binary ground truth corresponding to  bounding box marking is considered for salient object detection evaluation.}

In the following section, we introduce a new dataset and discuss three different subjective experiments performed using the dataset. We also justify that the data collected through the three experiments can be used together to appropriately represent multi-level object saliency ground truth.

\section{Subjective Experiments and Data Collection}
\label{sec:dataCollection}

In order to form our dataset, we examine a subset of the ImageNet~\cite{Deng_imagenet} images, in which, object bounding boxes are provided. Among those examined, we select 588 natural images that have multiple objects (total 2434 objects in 588 images). Our dataset is judiciously built ensuring sufficient variations among images in terms of the number of objects, the object sizes and positions, the object color content and the color contrast between the object and the rest in the image (see Appendix~1 (Section~\ref{appen}) for further explanations). We retrieved the original high resolution versions of the images for our dataset from Flickr (http://www.flickr.com) and the larger dimension of all of the images is {kept at 1024 pixels}. Our dataset and codes are publicly available at \textbf{https://github.com/gokyildirim/salmon\_dataset}. Compared to the datasets described in Section~\ref{sec:relatedWork}, our dataset is unique in the following aspects and hence more suitable:
\begin{compactitem}
	\item Each image contains multiple salient objects and the image sizes are large enough to properly accommodate objects of substantially different sizes.
	\item The saliency of objects is recorded from a sufficiently large number of subjects using multiple complementary subjective experiments.
\end{compactitem}

These factors help us appropriately investigate the saliency levels of objects with different features and sizes in the same image. As we discussed in Section~\ref{sec:introduction}, the saliency level of an object can be studied from visual attention (spontaneous) and perception. For this study, we conducted three subjective experiments: eye-tracking, point-clicking, and rectangle-drawing. The tasks in the subjective experiments {record} spontaneous attention and different kinds of human perception. The eye-tracking experiments are not associated with any goal and conducted in a free-viewing task, where only spontaneous attention is involved. Point-clicking involves a clicking task, which requires positional awareness of a salient object imparted through low-level perception~\cite{Shenton,Chalmers92}. The rectangle-drawing experiments involve concepts of salient object shape and size/scale, where it is necessary to tightly fit a rectangle, and require higher-level perception~\cite{Shenton,Chalmers92}. These subjective experiments measure object saliency at various perceptual levels of humans that can be arranged in a hierarchy as illustrated in Table~\ref{tab:hierarchy}. Tasks in the three subjective experiments achieved through spontaneous attention and levels of perception are shown in the table indicating the progression from bottom-up to top-down process domination. 

\begin{table}[h]
	\centering
	\caption{Hierarchy of our subjective experiments}
	\begin{tabular}{|c|c|c|c|c|}
		\multicolumn{1}{c}{} & \multicolumn{3}{c}{Bottom-up $\Rightarrow$  $\Rightarrow$ $\Rightarrow$ Top-down}\\
		\hline
		\multicolumn{1}{|c|}{} & \begin{tabular}{@{}c@{}}Free\\Viewing\end{tabular} & \begin{tabular}{@{}c@{}}Object\\Location\end{tabular} & \begin{tabular}{@{}c@{}}Object\\Scale \end{tabular}\\
		\hline
		Eye Tracking & \checkmark &  &\\
		\hline
		Point Clicking & \checkmark & \checkmark &\\
		\hline
		Rectangle Drawing & \checkmark & \checkmark & \checkmark\\
		\hline
	\end{tabular}
	\label{tab:hierarchy}
\end{table}
{In Figures~\ref{fig:multimodalGroundTruth}\subref{fig:fixationGT}, \subref{fig:pointGT}, and \subref{fig:rectangleGT}, we visualize the object-level saliency analogs of eye-fixation, point-clicking and rectangle-drawing data. These maps represent object saliencies in the image at different levels of perception. Although object saliencies at different perceptual levels can vary, in general, one of the maps can not be recognized as more appropriate than the others.} {Instead, they reflect the hierarchical differences in perception (see Table~\ref{tab:hierarchy}) and represent these differences at the object level. This allows us to evaluate and verify the features and assumptions of existing salient object detection methods, which in turn should provide guidelines on how to improve them and on what type of applications we can use them.}

In order to precisely measure object saliency, we manually segment all objects in all the images, and then study object saliency using the experimental data collected.

\subsection{Object Segmentation}
\label{obsegmask}
\begin{figure}[!h]
	\subfloat[Original image]{\frame{\includegraphics[width=4cm]{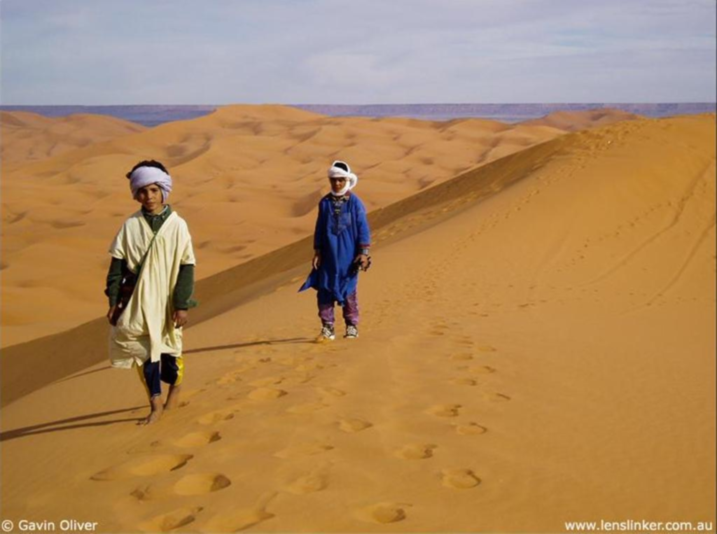}}}
	\hfill
	\subfloat[$\mathbf{M}^o$]{\frame{\includegraphics[width=4cm]{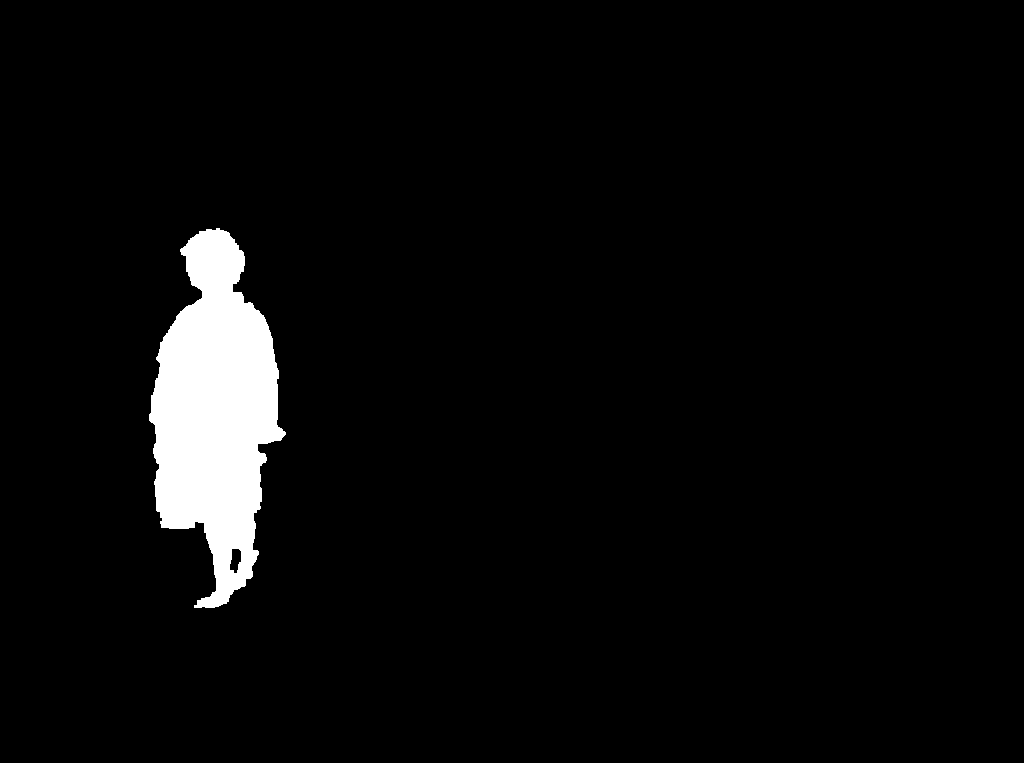}}}\\
	\caption{For each object in (a) an image, we compute a segmentation mask for (b) each individual object (like the one in white = 1).}
	\label{fig:averageColorCalculation}
\end{figure}

As we investigate object saliency, we need to know which pixels belong to which object or background. Objects in images are separated from the background by segmenting them with pixel-precise outlines. The segmentation mask $\mathbf{M}^o$ of an object $o$ is illustrated in Figure~\ref{fig:averageColorCalculation}. We use this mask to measure object saliency.

\subsection{Eye-Tracking Experiments}
\label{sec:eyeTrackingExperiments}

In order to measure the human visual saliency of objects due to their distinct features and information content that attract spontaneous attention, we perform eye-tracking experiments. We used RED250\footnote{{http://www.smivision.com/en/gaze-and-eye-tracking-systems/products/red250-red-500.html}} infrared eye-tracking device with a sampling frequency of 250 Hz and 9-point gaze calibration was considered. In total, we collected eye-fixation data from 95 people (48\% women, 52\% men) within ages 18-34. Each person in the experiment was asked to freely view randomly chosen 200 images (two experiments with 100 images shown in each) on an LCD monitor screen with a resolution of $1680\times 1050$ pixels. Each image was shown to the subjects for five seconds and an empty gray screen was shown for two seconds between two images to destroy persistence. Each image was viewed by 24 subjects on an average (min: 13, max: 34).

\begin{figure}[h]
	\centering
	\includegraphics[width=9cm]{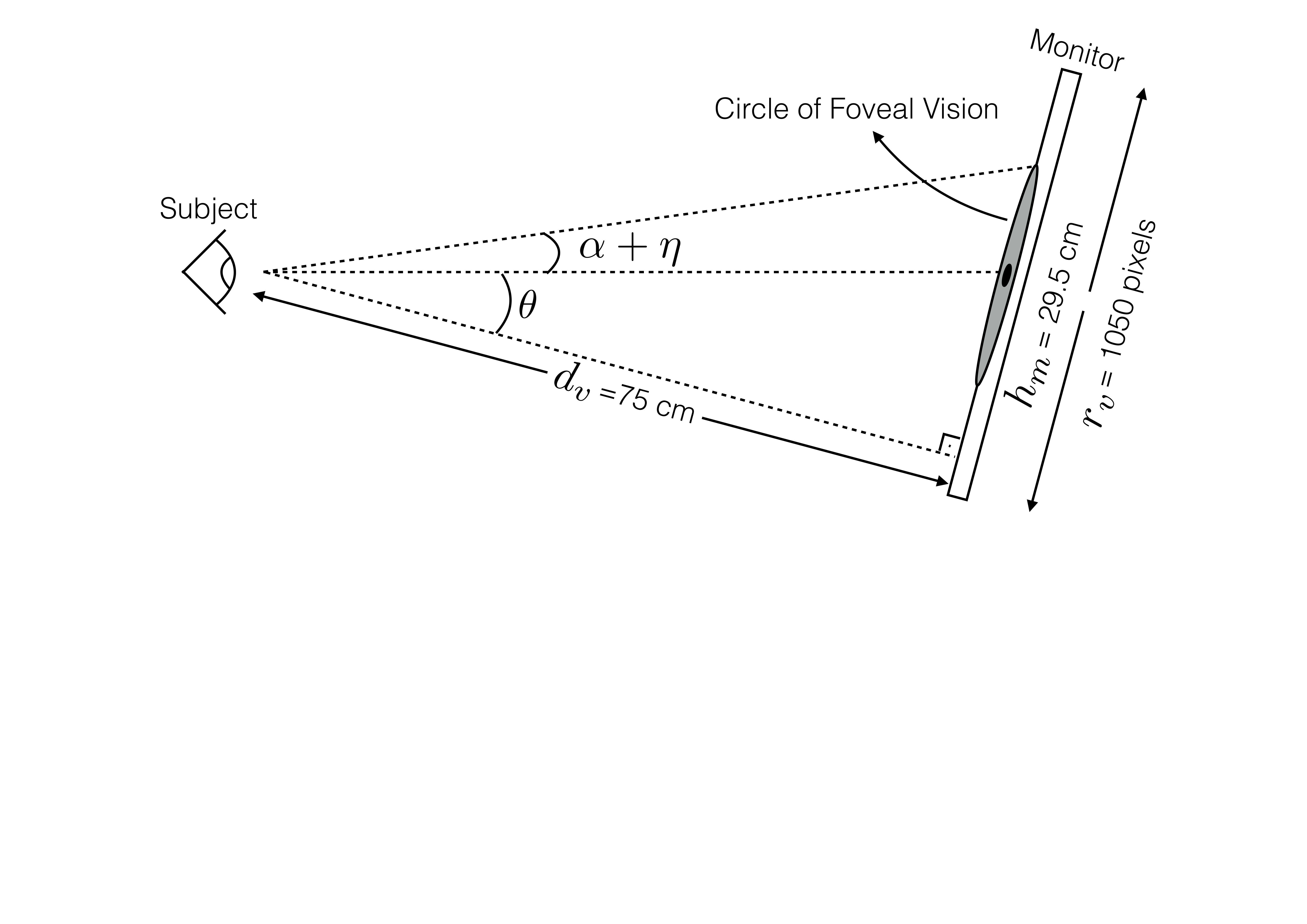}
	\vspace{-3cm}
	\caption{The configuration of the subject and the monitor during the eye-tracking experiments.}
	\label{fig:eyeTrackingSetup}
\end{figure}

For an image, a fixation by a person is recorded as a point at a single pixel location and a fixation map of the image size is formed with each pixel location containing a non-negative integer denoting the number of fixations by the person on it. We convert the fixation points of a person $\delta$ on an image $k$ into a fixation-density map $\mathbf{S}_{et}^{\delta,k}$ using a circular Gaussian filter on the fixation map. The $\sigma$ parameter of this filter is calculated using (\ref{eq:eyeTrackingSigma}), which is equal to the radius of the circle of foveal vision as shown in Figure~\ref{fig:eyeTrackingSetup}.
\begin{equation}
\label{eq:eyeTrackingSigma}
\sigma = d_v \cdot \frac{r_v}{h_m} \cdot \big( \tan(\alpha + \eta + \theta) - \tan(\theta) \big)
\end{equation}

Here, $\alpha = 1\degree$ is the half size of the human fovea, $\eta = 0.4\degree$ is the accuracy of the eye tracker, $d_v = 75\text{ cm}$ is the viewing distance, $r_v = 1050 \text{ pixels}$ is the vertical resolution of the monitor, and $h_m = 29.5 \text{ cm}$ is the height of the monitor. In our experiments $\sigma \approx 66$ pixels. After the Gaussian filtering, the fixation-density maps $\mathbf{S}_{et}^{\delta,k}$ are normalized between 0 and 1. 

\begin{figure}[h]
	\centering
	\subfloat[{Original image}]{\includegraphics[width=4.1cm]{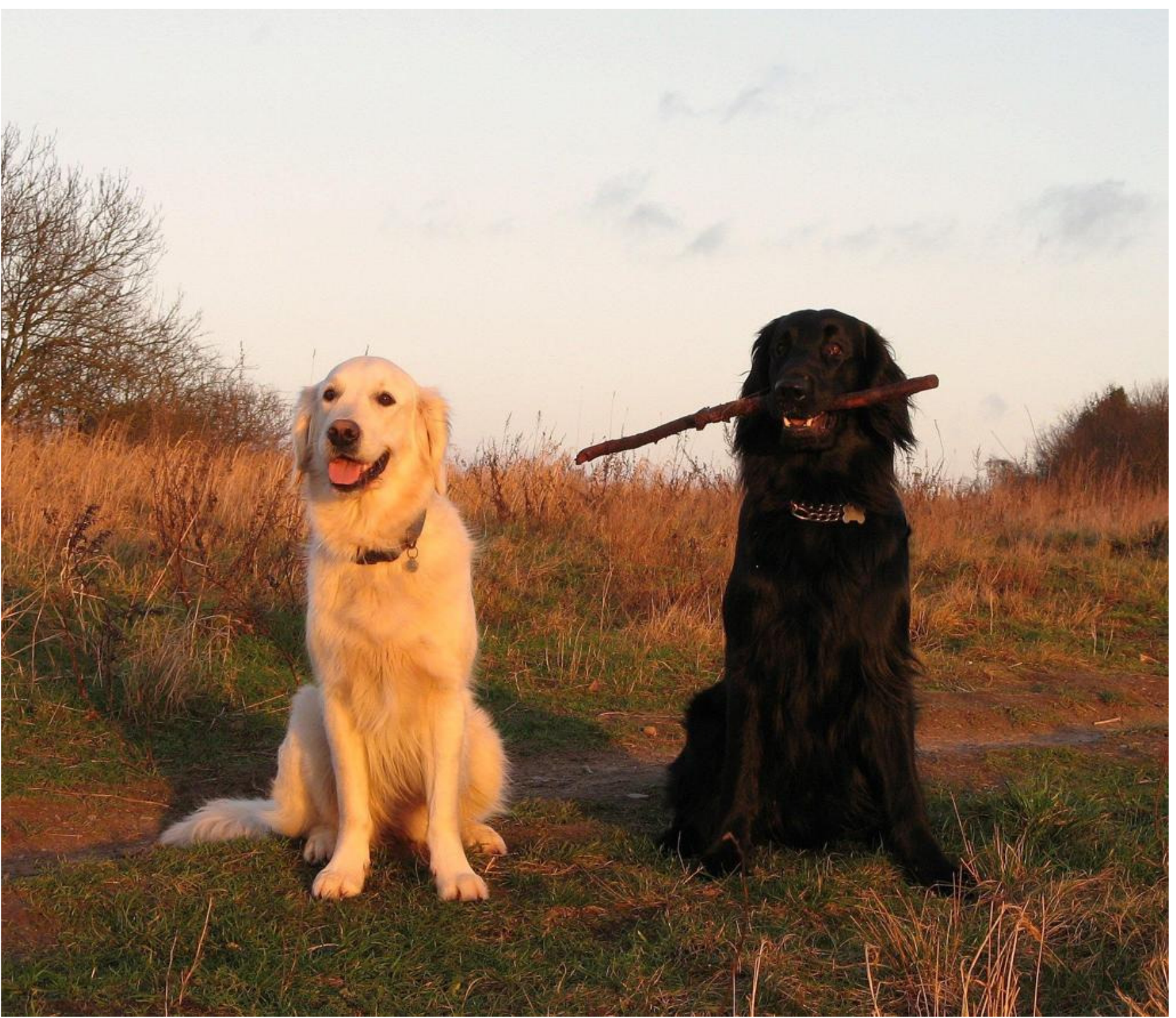}}
	\subfloat[{Eye-fixation density of a subject }]{\includegraphics[width=4.1cm]{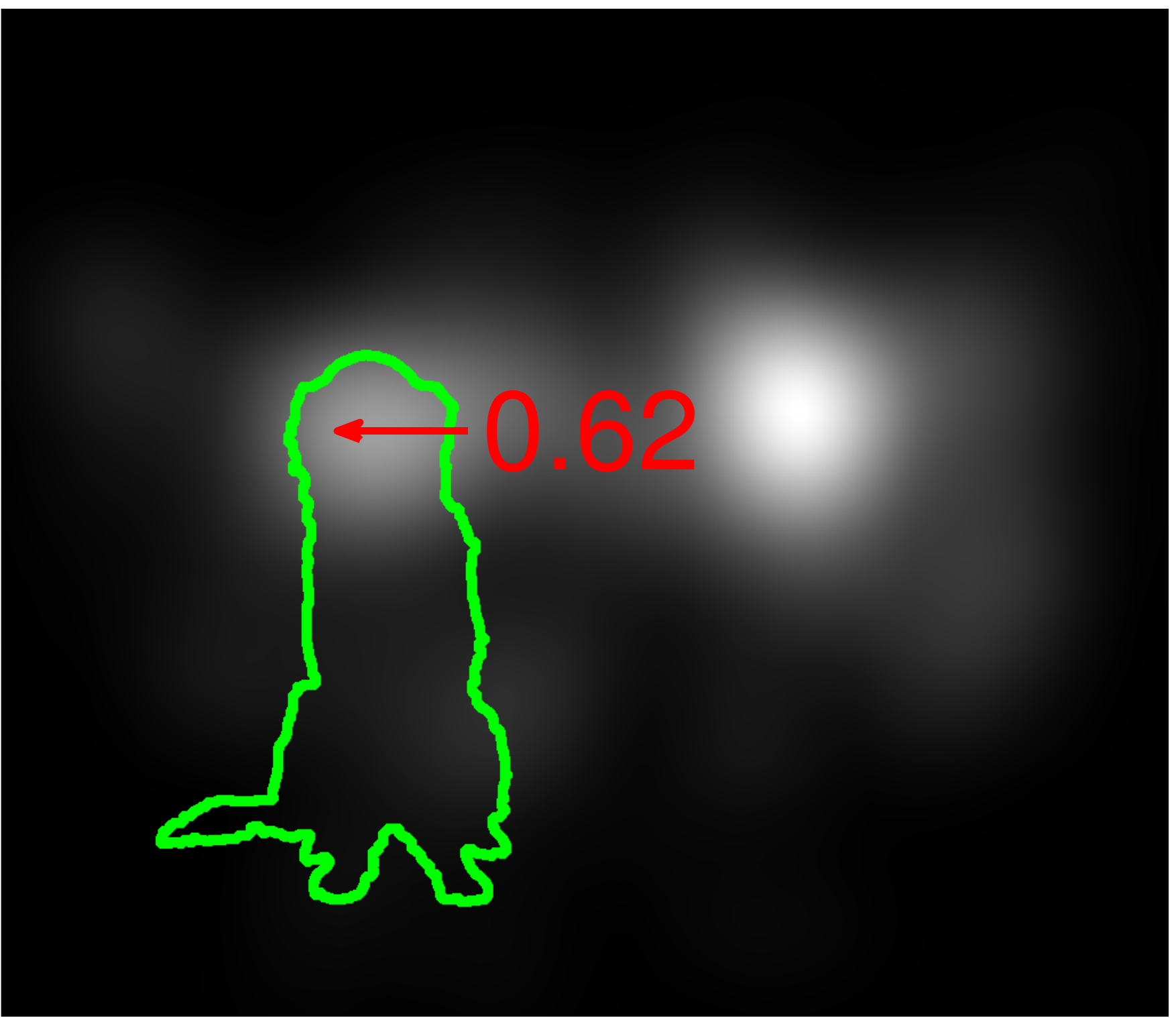}}
	\caption{{Eye-tracking saliency is defined as the average (over $\delta$) of the maximum values ($\displaystyle \max_{i \in \mathbf{M}^o} \big(\mathbf{S}_{et}^{\delta,k}(i)\big)$) inside object boundaries, which is $0.62$ in (b) (green boundary).}
	}
	\label{fig:eyeTrackingExperiments}
\end{figure}

Fixation-density maps indicate the visual saliency level of the object in an image as measured by recording gaze. We use these maps to measure the \lq\lq eye-tracking saliency", namely $s_{et}^o$, of an object $o$ in an image. It is possible that different people fixated at different distinctive parts of the same object which they deemed salient. Considering the fact that objects with a single distinctive part are more likely to be fixated upon \cite{Yarbus_eye_movements}, to measure $s_{et}^o$, we use the maximum fixation-density values within the boundary of object $o$ in image $k$ as follows:
\begin{equation}
s_{et}^o = \frac{\sum_\delta \displaystyle \max_{i \in \mathbf{M}^o} \big(\mathbf{S}_{et}^{\delta,k}(i)\big)}{N_{et}^k}
\end{equation}
Here $N_{et}^k$ is the number of people who viewed the image $k$ during the eye-tracking experiments and $i$ is a pixel inside the object boundary ($i \in \mathbf{M}^o$). An illustration of this operation is given in Figure~\ref{fig:eyeTrackingExperiments}.

{We convert the eye-fixation density {of a person $\delta$ on an image $k$ }into a saliency value for a whole object $o$ via $\max_{i \in \mathbf{M}^o} \big(\mathbf{S}_{et}^{\delta,k}(i)\big)$. We do so considering the fact that in an image where objects lay in front of the background, a person fixating on a part of an object indicates interest on the entire object as a single entity. This enables us to evaluate salient object detection methods against the saliency provided by eye-tracking ground truth, as these methods focus on pixel-wise segmentation of the salient objects, rather than eye-fixation prediction.}

\subsection{Point-Clicking Experiments}
\label{sec:pointClickingExperiments}

In order to measure human visual saliency of objects at various perceptual levels, we perform a couple of experiments involving subjective tasks. First, we consider the measurement of saliency at low-level perception. Correlating low-level perception to the so-called `dorsal' or `where' stream of the two-streams hypothesis related to neural processing of visual signals~\cite{Kandel}, we consider point-clicking experiments, which only require positional awareness of salient objects. In order to perform the point-clicking experiments, we used a crowd-sourcing web site\footnote{http://www.shorttask.com}. We asked people to click on the ``important objects'' that they ``notice at first glance''. The ``first glance'' phrase was used to avoid multiple perceptual processing of a single image entity by a subject to the extent possible. Such avoidance of multiple time perceptual processing makes this experiment akin to the eye tracking one where multiple processing is avoided due to the phenomenon of inhibition of return \cite{Itti_a_model}. The task duration was limited to 30 minutes, and in each task, randomly selected 42 images were shown simultaneously. Therefore, approximately 45 seconds were available to the subjects per image, which is sufficient as far as seeing the image and clicking is concerned. In this experiment, each image is viewed by 33 people on an average (min: 24, max: 38).

\begin{figure}[h]
	\centering
	\subfloat[{Original image with point clicks}]{\includegraphics[width=4.1cm]{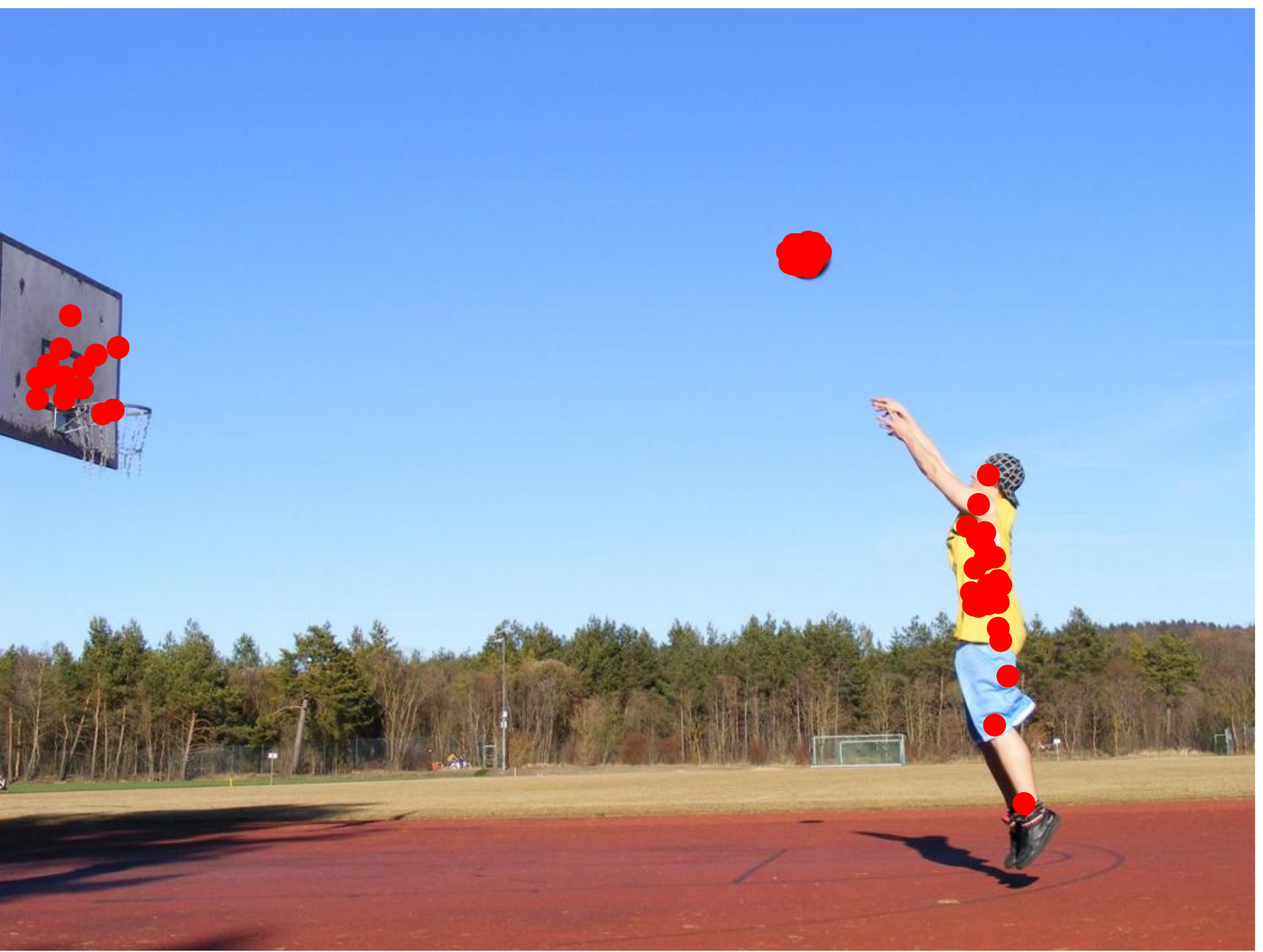}}
	\subfloat[{Clicks on a single object}]{\includegraphics[width=4.1cm]{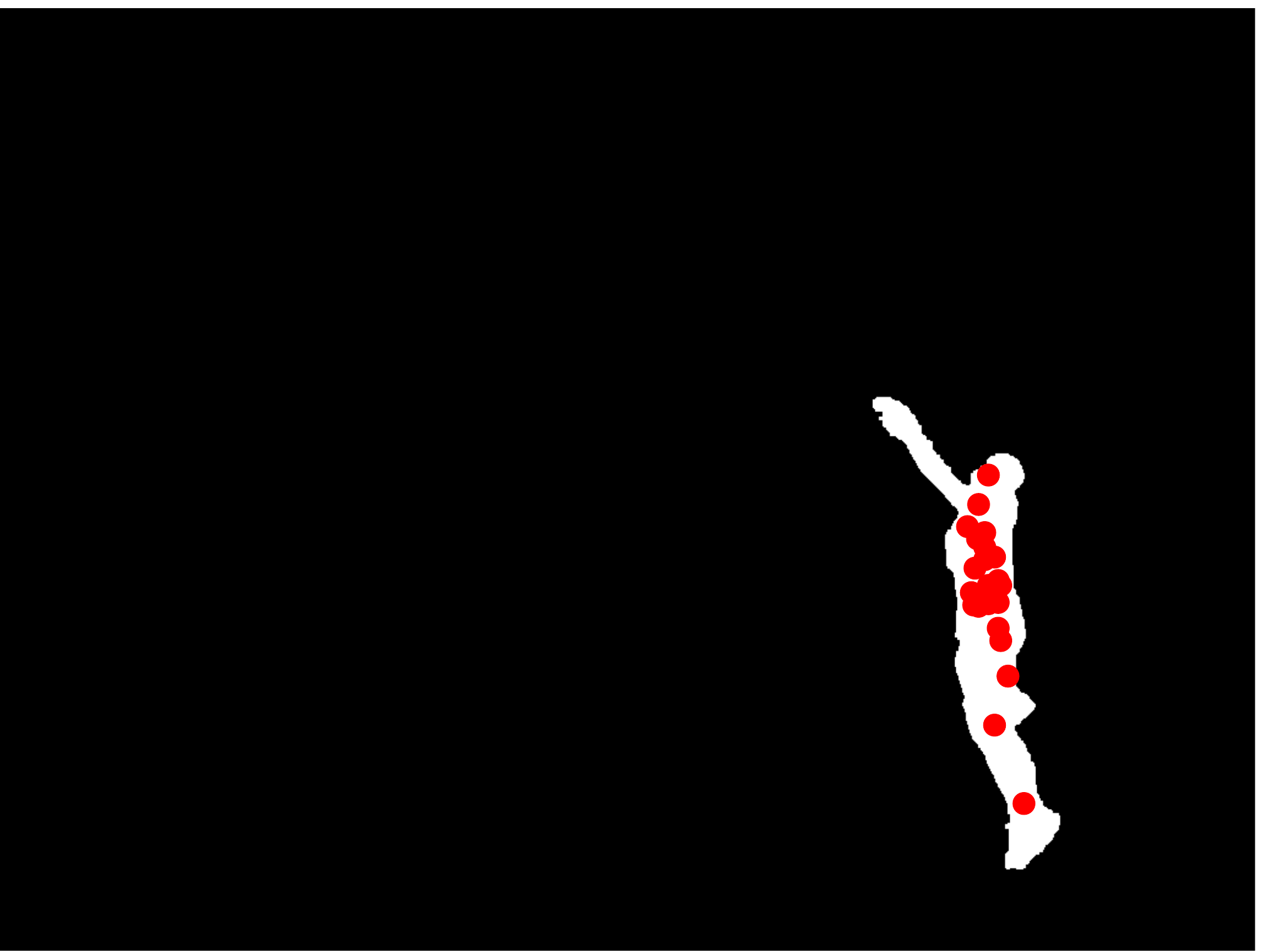}}
	\caption{{23 out of 30 subjects clicked on the object in (b), which makes its point-clicking saliency $s_{pc}^o = 23/30 \approx 0.76$.}
	}
	\label{fig:pointClickingExperiments}
\end{figure}

We represent the set of points (pixels $i$) where a person $\delta$ clicked on image $k$ as $\mathbf{S}_{pc}^{\delta,k}$. In order to measure the \lq\lq point-clicking saliency", namely $s_{pc}^o$, of an object $o$ in image $k$, we count the number of people who clicked an object and normalized it with the number of subjects who viewed the image. {This is very similar to the procedure used in PASCAL-S~\cite{Sal_secret}, where an object is considered as salient by a person when they click within the boundary of that object.} Formally, it can be calculated as follows:
\begin{equation}
\begin{aligned}
s_{pc}^o &= \frac{\sum_\delta f(\mathbf{M}^o, \mathbf{S}_{pc}^{\delta,k})}{N_{pc}^k}\\
f(\mathbf{M}^o, \mathbf{S}_{pc}^{\delta,k}) &= \begin{cases} \exists i \in \mathbf{S}_{pc}^{\delta,k} \colon i \in \mathbf{M}^o\text{,} & 1 \\ \text{else,} & 0\end{cases}
\end{aligned}
\end{equation}
Here $N_{pc}^k$ is the number of people who viewed the image $k$ during point-clicking experiments. An illustration of this operation is given in Figure~\ref{fig:pointClickingExperiments}.

\subsection{Rectangle-Drawing Experiments}
\label{sec:rectangleDrawingExperiments}
We consider a second experiment to measure of saliency at higher-level perception. Correlating higher-level perception to the so-called `ventral' or `what' stream of the two-streams hypothesis, we consider rectangle-drawing experiments, which involve concepts of shape and scale. Similar to the point clicking experiments, rectangle-drawing experiments are performed using crowd sourcing. We asked people to draw a tight rectangle around the ``important objects'' that they ``notice at first glance''. Again, the task duration was limited to 30 minutes and in each task 42 images were shown. In this experiment, each image is viewed by 32 people on an average (min: 15, max: 50).

\begin{figure}[h]
	\centering
	\subfloat[{Original image with rectangles}]{\includegraphics[width=4.1cm]{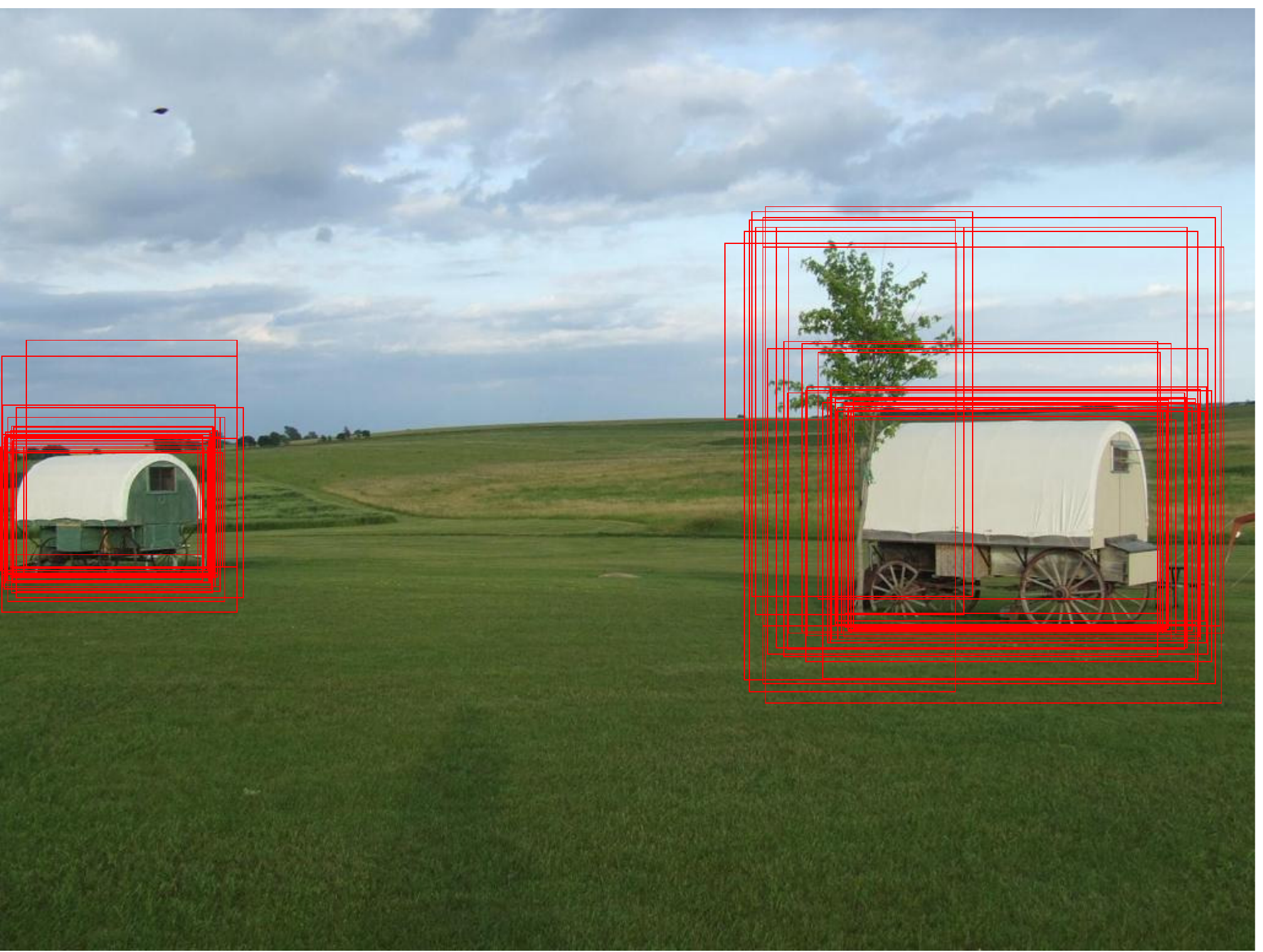}}
	\subfloat[{Rectangles on a single object}]{\label{fig:framingRectangle}\includegraphics[width=4.1cm]{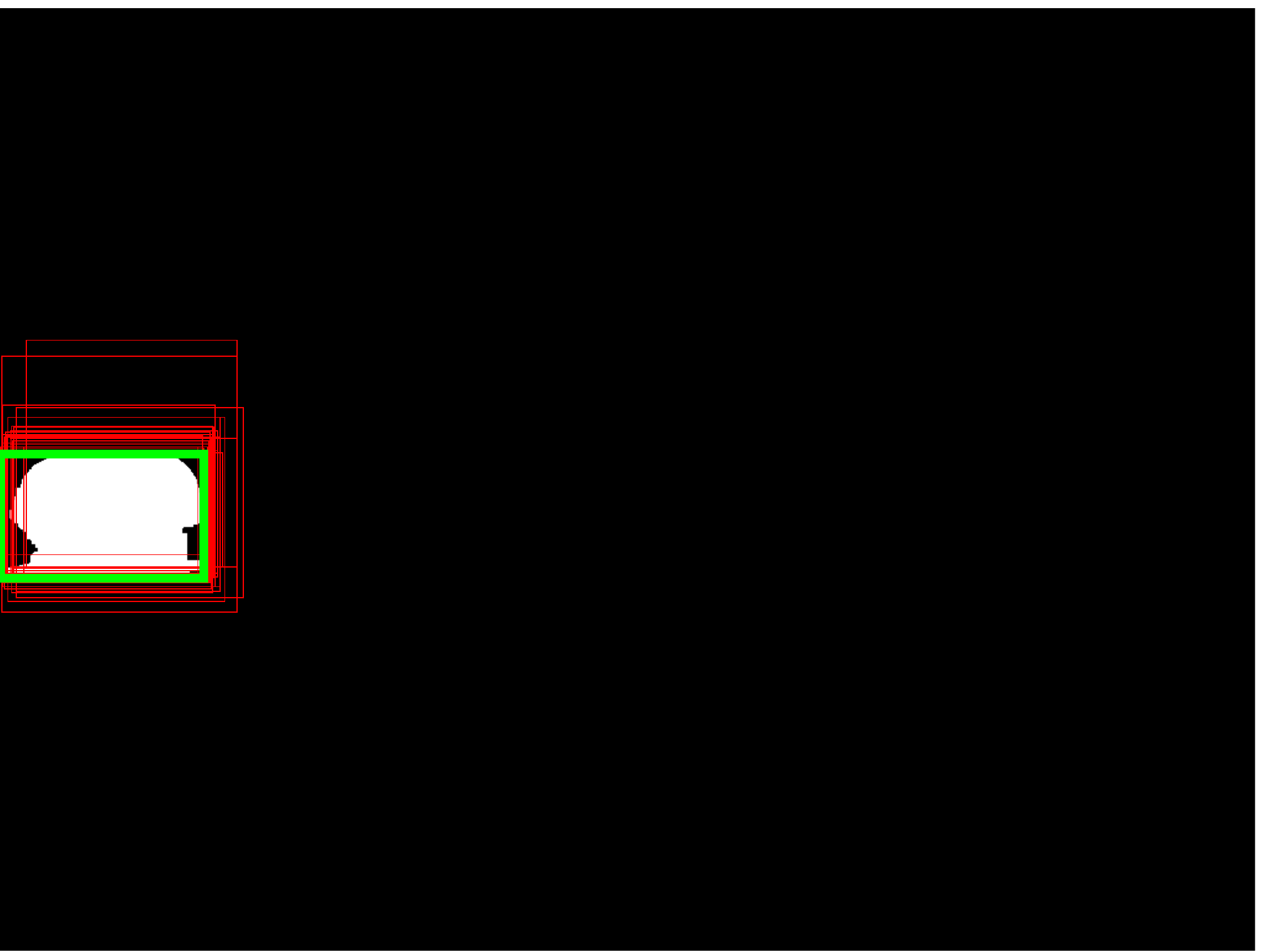}}
	\caption{{27 out of 35 subjects drew overlapping rectangles (IoU $\ge 0.3$ w.r.t. green rectangle) on the object in (b), which makes its rectangle-drawing saliency value $s_{rd}^o = 27/35 \approx 0.77$.}
	}
	\label{fig:rectangleDrawingExperiments}
\end{figure}

We represent the set of rectangles which a person $\delta$ drew on image $k$ as $\mathbf{S}_{rd}^{\delta,k}$. In order to measure the \lq\lq rectangle-drawing saliency", namely $s_{rd}^o$, of an object $o$ in image $k$, we count the number of people, who drew a rectangle on an object with an intersection-over-union {(IoU)} score greater than 0.3 and normalize it using the number of subjects who viewed the image. {In object detection literature, IoU is used to verify whether an estimated rectangle is a correct detection~\cite{Everingham_pascal_visual}. Here, analogously, an object is considered as salient by a person, when they draw a rectangle with IoU $ \ge 0.3$ with respect to the tight rectangle around a segmented object (see Figure~\ref{fig:rectangleDrawingExperiments}\subref{fig:framingRectangle}).} We can calculate this value as follows:  
\begin{equation}
\begin{aligned}
s_{rd}^o &= \frac{\sum_\delta f(\mathbf{M}^o, \mathbf{S}_{rd}^{\delta,k})}{N_{rd}^k}\\
f(\mathbf{M}^o, \mathbf{S}_{rd}^{\delta,k}) &= \begin{cases} \exists r \in \mathbf{S}_{rd}^{\delta,k} \colon g(r,r^o) \geq 0.3\text{,} & 1 \\ \text{else,} & 0\end{cases}
\end{aligned}
\end{equation}
Here $N_{rd}^k$ is the number of people who viewed the image $k$ during rectangle-drawing experiments, $r$ is a rectangle in $\mathbf{S}_{rd}^{\delta,k}$, $g(.,.)$ is the function that computes the intersection-over-union score, and $r^o$ is the reference rectangle, which tightly encloses the object we segmented before. An illustration of this operation is given in Figure~\ref{fig:rectangleDrawingExperiments}.

It is evident from all the three measures of object saliency, that they can be any value between 0 and 1. Certain objects which are considered salient by some, may not be considered salient by others, and this subjectivity is addressed by considering the collective response of a large number of persons. Following this, it becomes clear from the multiple values of object saliency that objects in natural images are seen and perceived to have varying levels of importance. Hence, while evaluating a salient object detection performance quantitatively, it would be imperative for a human to consider the different levels of object importance. Therefore, for an image with multiple objects, a salient object detection algorithm must be evaluated considering the varying level of importance of the objects, making the automatic objective evaluation more meaningful (human-like).

\section{Multi-Level Object Saliency}
\label{sec:visualSaliencyOfObjects}

\begin{figure}[!h]
	\centering
	\subfloat[]{\label{fig:saliencyDistributionsET}\includegraphics[width=4cm]{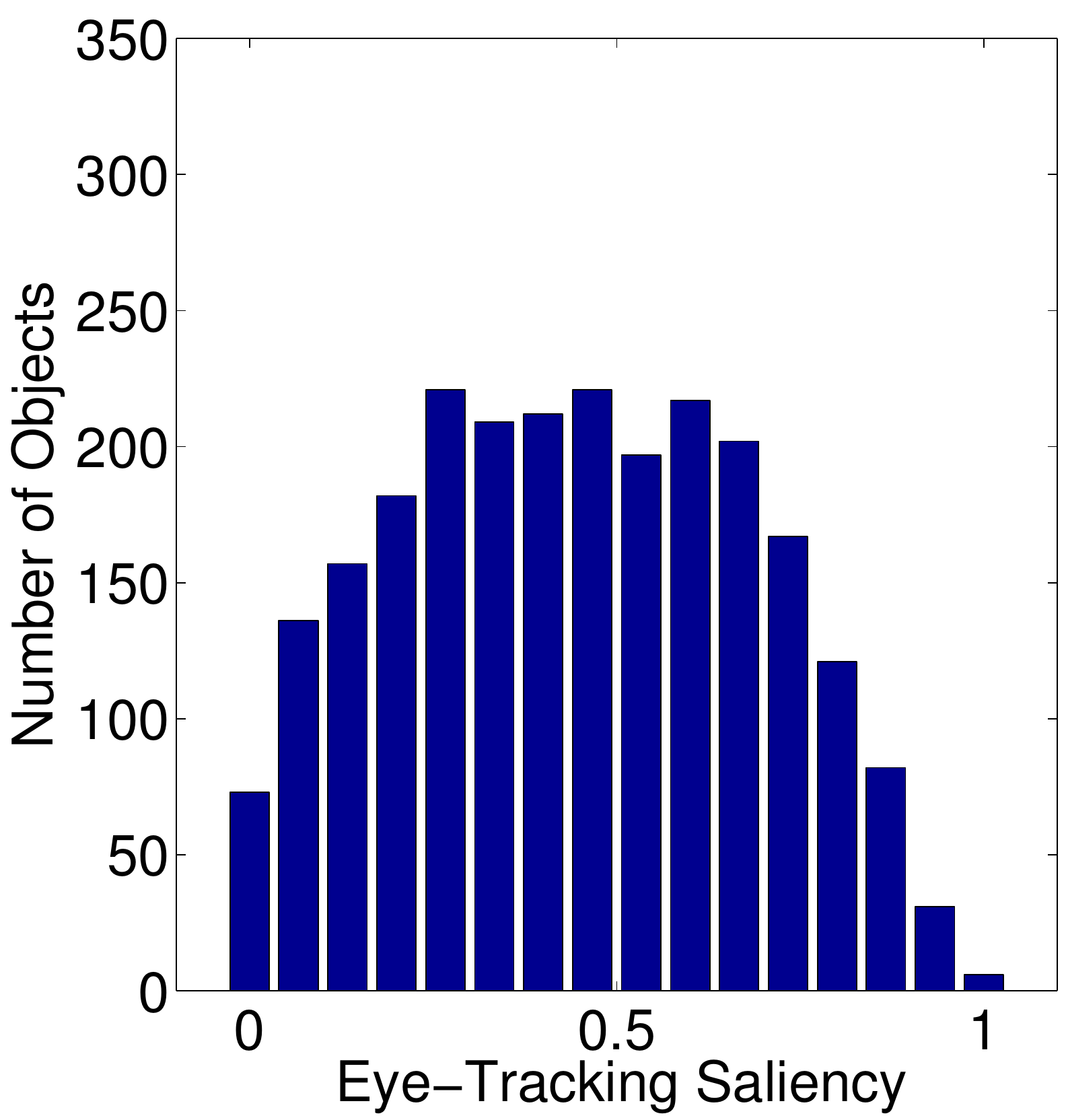}}\hfill
	\subfloat[$R^2$ = 0.64]{\label{fig:fsps}\includegraphics[width=4cm]{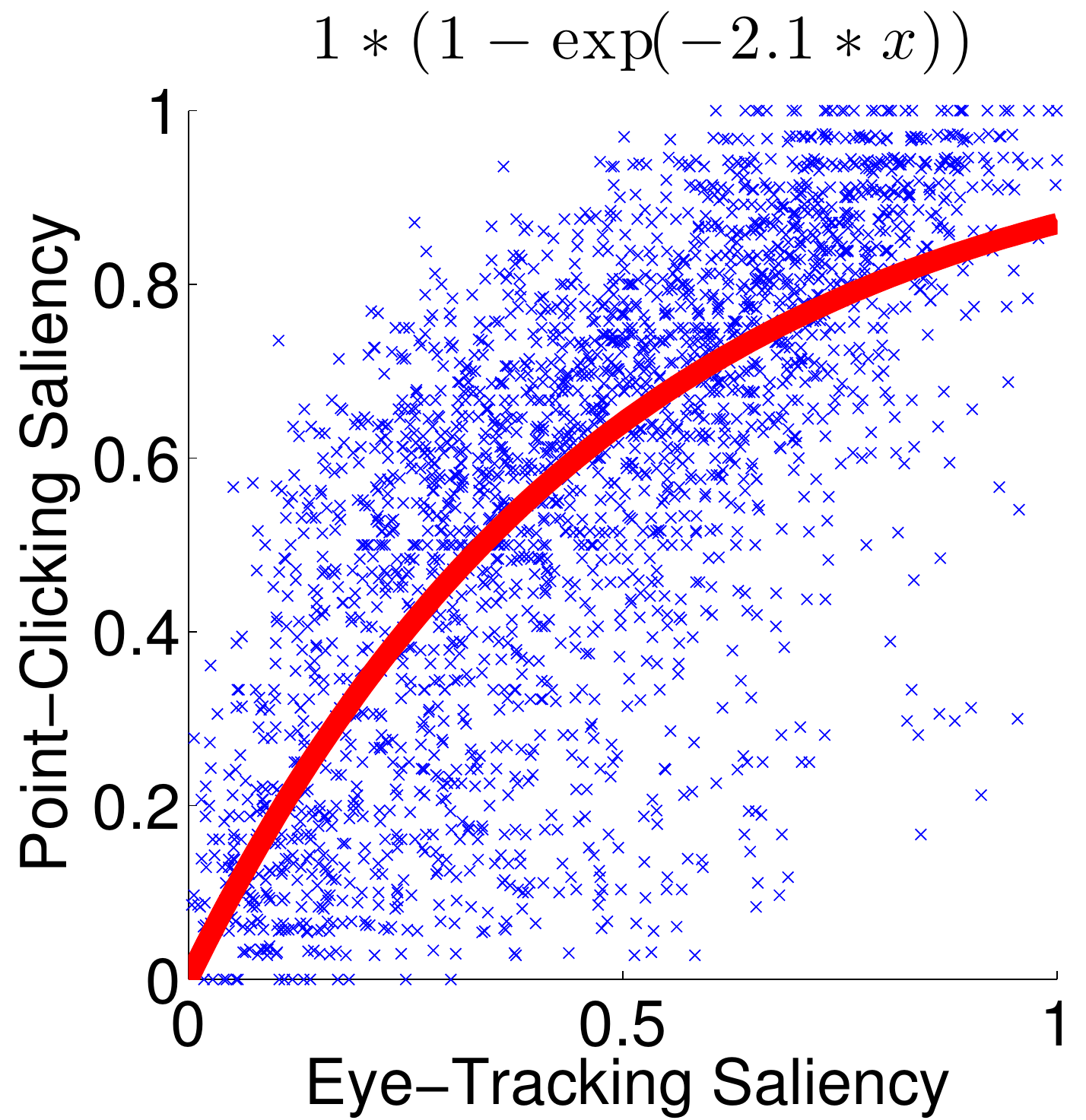}}\\
	\subfloat[]{\label{fig:saliencyDistributionsPC}\includegraphics[width=4cm]{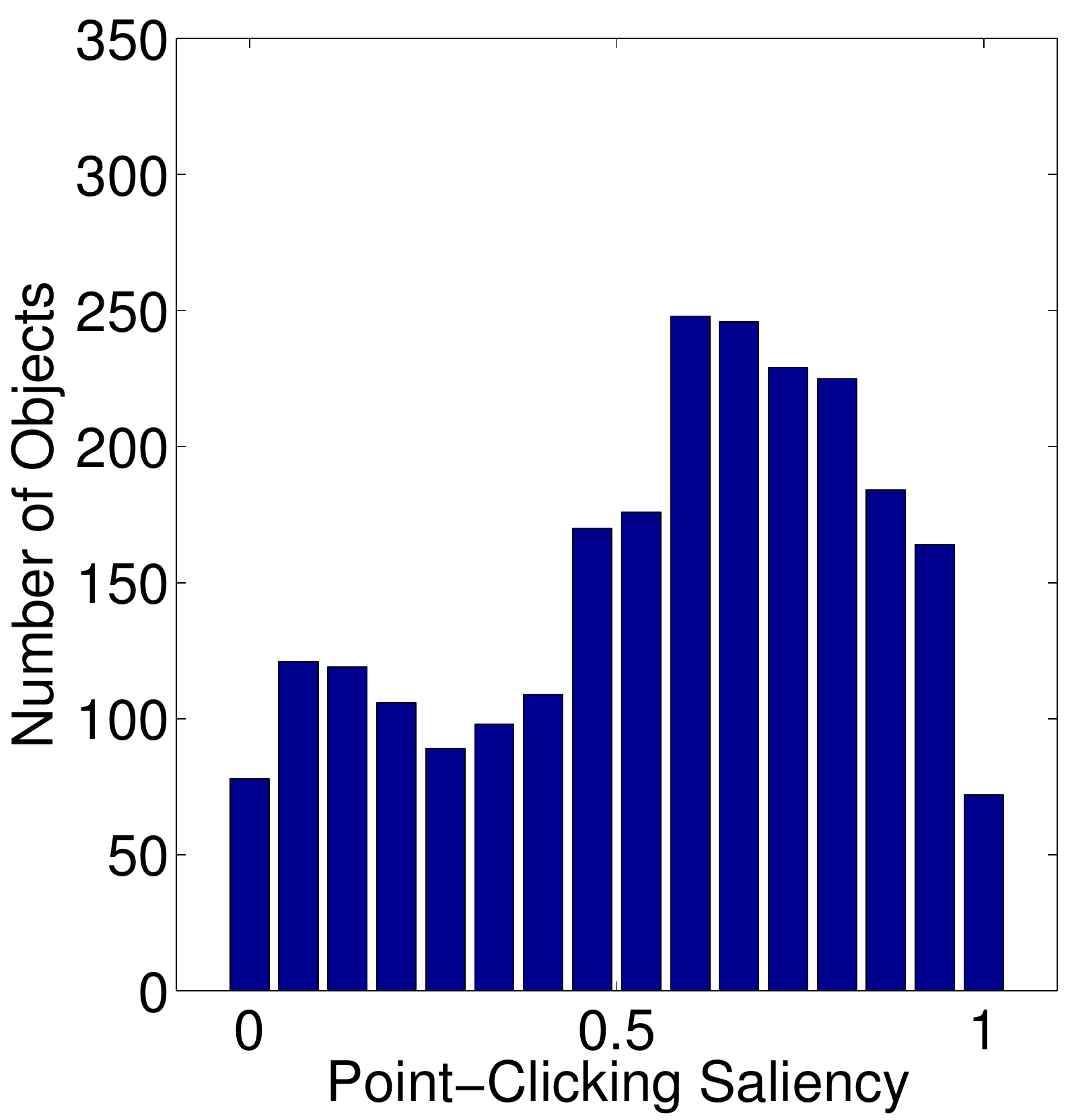}}\hfill
	\subfloat[$R^2$ = 0.48]{\label{fig:fsrs}\includegraphics[width=4cm]{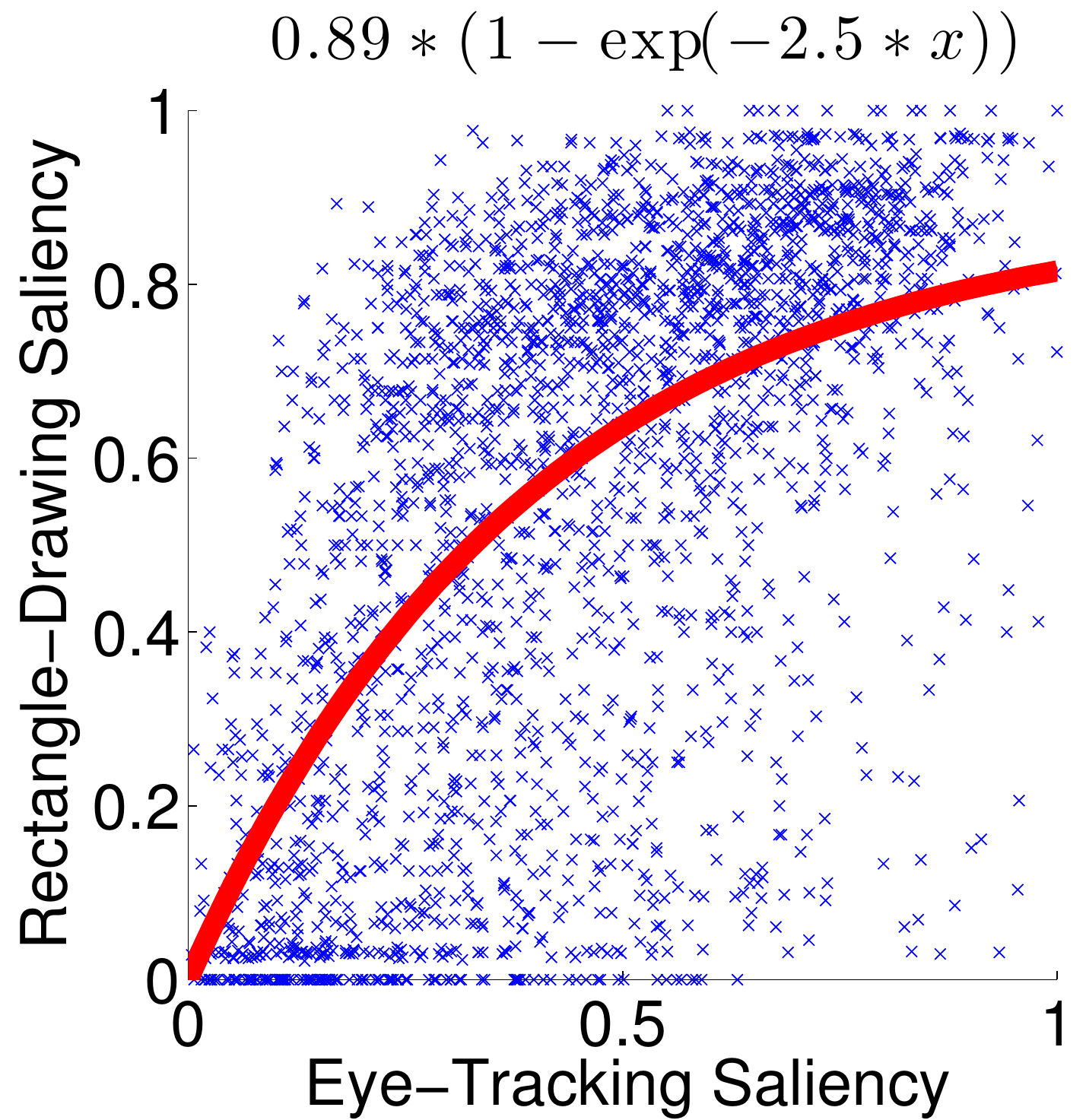}}\\
	\subfloat[]{\label{fig:saliencyDistributionsRD}\includegraphics[width=4cm]{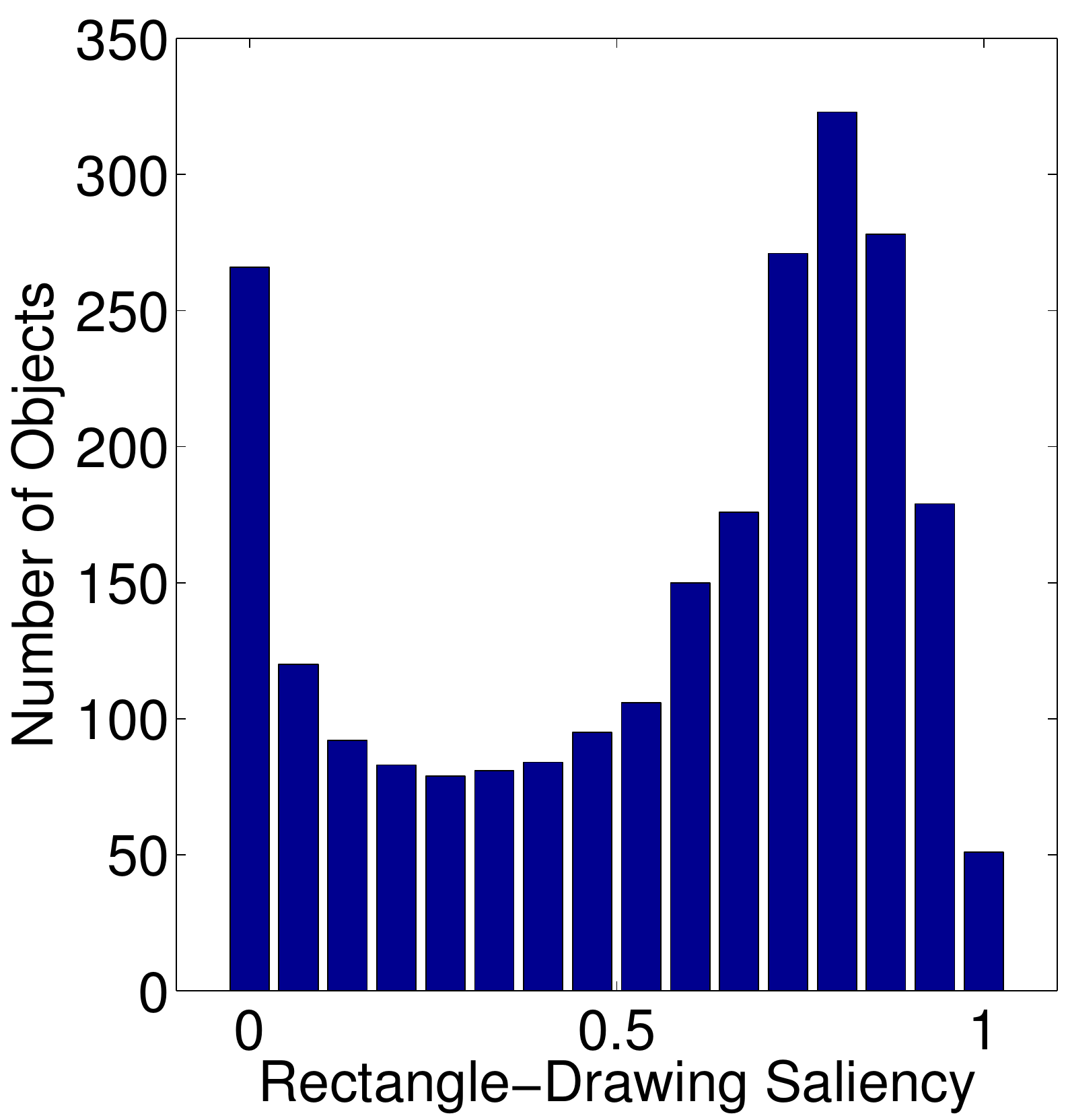}}\hfill
	\subfloat[$R^2$ = 0.79]{\label{fig:psrs}\includegraphics[width=4cm]{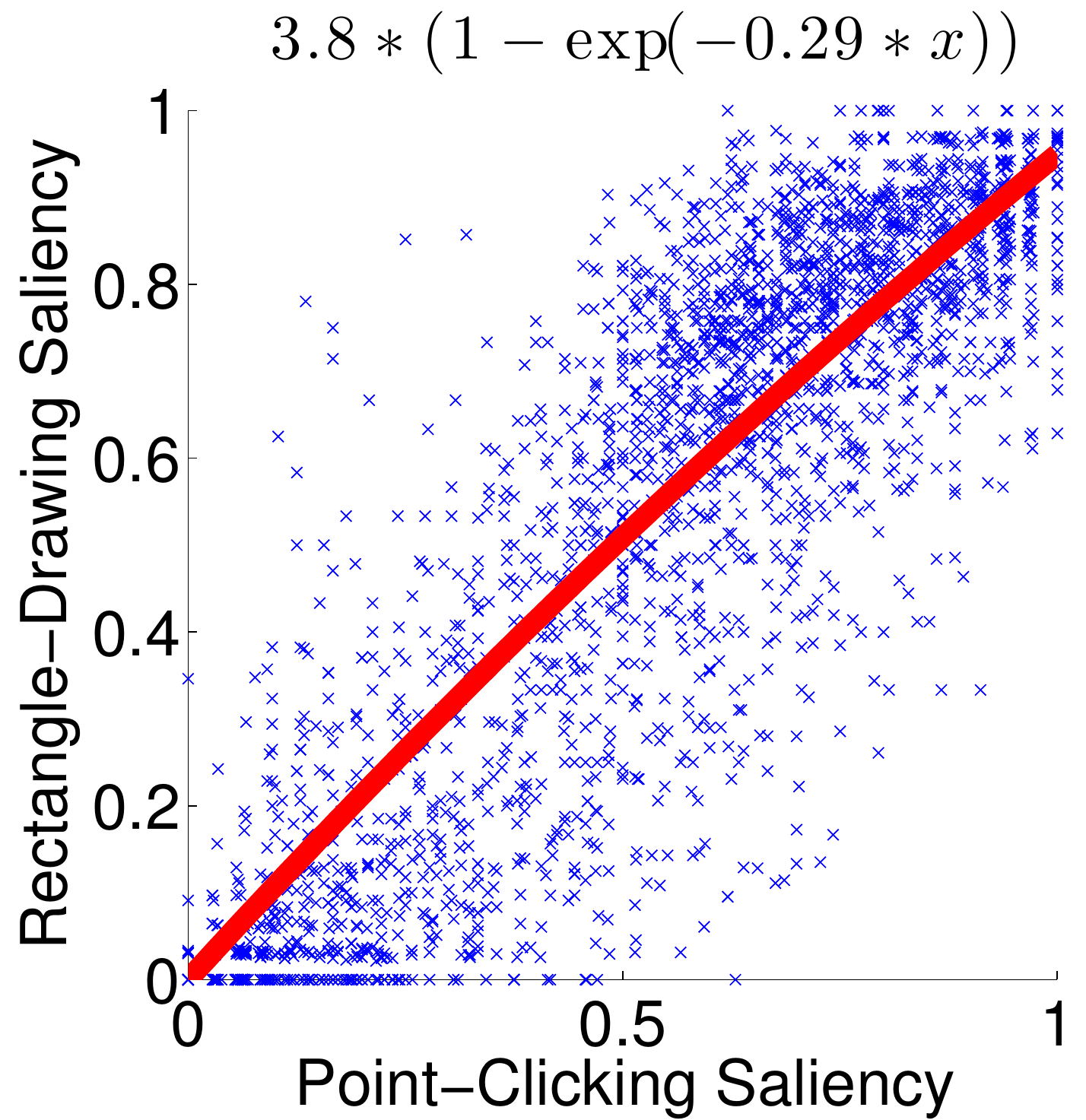}}
	\caption{(a,c,e) The distribution of the eye-tracking ($s_{et}^o$), point-clicking ($s_{pc}^o$), and rectangle-drawing ($s_{rd}^o$) saliency values of the objects in our dataset and (b,d,f) the coefficient of determination between different saliency values (red lines indicate a non-linear fit).}
	\label{fig:objectSaliency}
\end{figure}

Let us consider the varying levels of importance of objects, which makes object saliency inherently multi-level, further here. Multi-level object saliency implies that saliency of objects can take any value between 0 and 1, rather than having two distinct (i.e. binary) values. In Figures~\ref{fig:objectSaliency}(a), (c), and (e), we illustrate the measured saliency values of the 2434 objects in our dataset in the form of distributions. We can clearly see that the objects are not equally salient, i.e. object saliency is multi-level, as far as collective eye-fixation durations and human perception are concerned.

The measured saliency values of an object are different for three subjective experiments. In Figures~\ref{fig:objectSaliency}(b), (d), and (f), we show the relationship between eye-tracking, point-clicking, and rectangle-drawing saliency values of the objects in our dataset in pairs. For each pair, we fit a gamma non-linearity that maps the values from the x-axis to the y-axis. The correlation coefficients (squared) or coefficient of determination for these mappings are given below the plots in Figure~\ref{fig:objectSaliency}. We see that the hierarchy of the subjective experiments given in Table~\ref{tab:hierarchy} is reflected by the coefficients. The eye tracking and rectangle drawing saliencies are more dissimilar from each other than they are from the point clicking saliency. This observation is related to the levels of human perception involved in the experiments, with eye tracking and rectangle drawing being the two opposite ends.

On a different note, depending on the desired kind of object saliency, eye-tracking, point-clicking, and rectangle-drawing saliency values can be employed in different applications. For example, we can use the eye-tracking data as an overall indicator for human spontaneous attention in psychophysical studies~\cite{Engelke_comparative_study,Toet_computational_versus}. In point-clicking experiments, a subject is required to voluntarily move the mouse and to perform the click. So the point-click data does not comprise of components present in eye-tracking data that is due to involuntary eye-movements, experiment fatigue, etc., which are considered unwanted in certain cases. In addition, clicked-points can be very useful for resolving the ambiguity in eye-fixation density maps when two or more objects are spatially very close to each other. Therefore, compared to eye-tracking, point-clicking experiments provide more certain object-level saliency and can be used in areas such as web site design~\cite{Ivory_improving_web}. In rectangle-drawing experiments, subjects needed to fit a rectangle around the objects of interest, which further specifies saliency compared to point-clicking experiment. For example, when a subject clicks on a human face in an image, it is not obvious whether the face or the whole body of that human is considered as salient. In rectangle-drawing experiments, however, a drawn rectangle would remove such uncertainty, which is valuable in objectness measurement~\cite{Alexe_measuring_the}.

In Section~\ref{sec:introduction}, we referred to literature which suggests that humans fixate on (spontaneously attend to) objects that are distinct and more informative for a longer period of time compared to other parts of an image. In~\cite{Alers_how_the}, Alers et al. compare the eye-fixation trends of subjects under two conditions: free-viewing and task-driven. They observe that subjects have a tendency to fixate for more time on the distinct informative objects, when they are not given a certain task (free-viewing condition). As in our free-viewing eye-tracking experiments we take eye-fixation duration as a measure of saliency (importance), we implicitly consider saliency to be correlated to distinctness and informativeness. This is unlike the other two experiments, where the subjects are given explicit instructions to notice important objects. However, as it is well known that human vision performs image understanding by considering only pertinent visual data, the process of spontaneous attention and subsequent fixation duration obviously works to choose the pertinent data, and hence, our implicit consideration is well-founded.

As we discussed in Sections~\ref{sec:introduction} and~\ref{sec:dataCollection}, object saliency is subjective. Some objects which are considered salient by a few, may not be considered salient by others. When we measure object saliency, we take this subjectivity of saliency into account by considering collective human opinion /response. For example, in our point-clicking experiments, we asked a group of people to click on the important objects. The subjects of this experiment followed various strategies, such as clicking on all the objects, only on the most prominent object, or on the objects that are closer to the camera. We measure the collective object saliency by considering the ratio of the number of people who clicked on an object in an image to the number of people who viewed the image. 

We use both point-clicking and rectangle-drawing saliency values as measures based on collective human perception. Note that, the standard deviation of the collective opinion is correlated to the saliency measured. When the majority of the subjects agrees on the significance (or insignificance) of an object in an image, standard deviation of collective opinion is low and the saliency value of that object approaches to 1 (or 0). Whereas, a saliency value near 0.5 implies strong disagreement on object saliency with the subjects being equally divided into two groups and the standard deviation of collective opinion is high.

Even with such careful and conceptually clear measurement of object saliency, there are some, mostly trivial issues that we overlooked in preference to simplicity. However, one important issue that we encountered is the order in which a subject looked at, clicked on, or drew on an object, which we found could be a factor influencing object saliency. For example, during point-clicking experiments, most subjects usually clicked in a horizontal order from left to right. This order of attention introduced a small but undesired bias in saliency measurements making the objects on the left slightly more salient than the objects on the right when no other factor plays a role. {In the future, this bias can possibly be alleviated by applying a random horizontal flip to the displayed images, so that the left to right preference is averaged out.}

\section{Performance Evaluation of Existing Salient Object Detection Approaches}
\label{sec:salientObjectDetectors}
As we have seen till now that image objects are inherently perceived to possess multi-level saliency, salient object detection techniques must be evaluated considering multi-level object saliency ground truth, and hence, we present here such an evaluation of the state-of-the-art. Salient object detectors generate object saliency maps, which are gray-level maps, from which binary object saliency maps are subsequently generated during their standard evaluation against binary ground truth using precision-recall curves. Obviously, if multi-level object saliency ground truth is available, the gray-level object saliency maps generated during salient object detection could be directly used against them for evaluation. We generate multi-level object saliency ground truth corresponding to all the images in our dataset discussed earlier in Section~\ref{sec:dataCollection} using the results of the subjective experiments, and use them here for evaluating a few existing salient object detection approaches.

We get the gray-level object saliency maps generated by the following salient object detectors using their publicly available codes: IT~\cite{Itti_a_model}, FT~\cite{Achanta_frequency_tuned}, FASA~\cite{Yildirim_fast_accurate}, HC~\cite{Cheng_global_contrast}, RC~\cite{Cheng_global_contrast}, CB~\cite{Jiang_automatic_salient}, GC~\cite{Cheng_efficient_salient}, SF~\cite{Perazzi_saliency_filters}, LR~\cite{Shen_a_unified}, CH~\cite{Li_contextual_hypergraph}, AMC~\cite{Jiang_saliency_detection}, GMR~\cite{Yang_saliency_detection}, HR~\cite{Yildirim_saliency_detection}, MB~\cite{Zhang2015}, IILP~\cite{Li2015}, SMD~\cite{Peng2017}, and MIL~\cite{Huang2017}. We use these gray-level saliency maps to compare the salient object detectors using our dataset and ground truth. Note that, among the above detectors, the ones involving machine learning have been trained by the authors using binary ground-truth maps.

The widely-used performance metric, precision-recall curve is applicable with the use of binary ground-truth maps. Multi-level ground-truth maps, however, consist of multiple values. Therefore, we evaluate and compare the performances of the existing methods {on the following three different aspects:
\begin{compactitem} 
\item[\it{Regression:} ] We use a standard regression loss, mean absolute error, to evaluate the accuracy in estimation of multiple saliency values.
\item[\it{Classification:} ] We consider a consistent utilization of the area under precision-recall curve measure to evaluate classification performance at multiple saliency levels.
\item[\it{Ranking:} ] We use a standard ranking correlation measure, Kendall rank correlation coefficient, to evaluate the accuracy in ranking objects based on the estimated multi-level saliency.
\end{compactitem} 
}

\subsection{Forming Multi-Level Ground-Truth Maps}

We generate multi-level ground-truth maps from the eye-tracking ($s_{et}^o$), point-clicking ($s_{pc}^o$), and rectangle-drawing saliency ($s_{rd}^o$) values as follows:
\begin{equation}
\begin{aligned}
\mathbf{F}_{\gamma}^k = \sum_{\forall o \text{ in image } k} s_\gamma^o \cdot \mathbf{m}^o
\end{aligned}
\end{equation}
where
\begin{eqnarray}
\mathbf{m}^o(i,j)&=& 1, \forall (i,j)\in \mathbf{M}^o\\ \nonumber
&=& 0, otherwise
\end{eqnarray}
with $(i,j)$ representing a pixel in image $k$. Here, $\mathbf{F}_{\gamma}^k$ is the multi-level ground-truth map of the image $k$ using $\gamma$ saliency values, where $\gamma \in \{ et, pc, rd\}$ is one of the subjective experiments. This gives us three different ground-truth maps for each image in the dataset, which we will use in our analysis. Note that all the three ground truth maps take values from the interval $[0,1]$.

\subsection{Mean Absolute Error}
\label{MAEss}
Mean absolute error (MAE) measures the absolute difference between estimated and measured (ground truth) object-saliency values.{We employ mean absolute error as it is a well-known regression loss, and computation of saliency values can be directly interpreted as an estimation of regressands based on the input data acting as the regressors.} We calculate the MAE with respect to our dataset as follows:
\begin{eqnarray}
\label{myeq2}
\widehat{\text{MAE}}_{\gamma} &=& \frac{1}{N} \sum_{\forall o} \text{MAE}_{\gamma}^o\\
\nonumber
\text{MAE}_{\gamma}^o &=&  \frac{1}{|\mathbf{M}^o|} \sum_{(i,j) \in \mathbf{M}^o} \big|\mathrm{S}^o - \mathbf{F}_{\gamma}^k(i,j)\big|,\ \mathbf{M}^o \text{ in image } k \\
\label{myeq1}
&=& \big|\mathrm{S}^o - s_\gamma^o\big|\\
\label{myeq0}
\mathrm{S}^o &=& \frac{1}{|\mathbf{M}^o|} \sum_{(i,j) \in \mathbf{M}^o} \mathbf{S}(i,j)
\end{eqnarray}
Here, $\widehat{\text{MAE}}_{\gamma}$ is the average error over all objects, $N = 2434$ is the total number of objects in our dataset and $\text{MAE}_{\gamma}^o$ is the error on an object $o$ for saliency type $\gamma \in \{et,pc,rd\}$.  In (\ref{myeq0}), $\mathbf{S}$ is the estimated gray-level object saliency map (with value $\in [0,1]$) by an algorithm, and in (\ref{myeq1}), $\mathrm{S}^o$ is considered as the estimated saliency value of the object $o$. Note that, $(i,j)$ represents a pixel and $\forall (i,j) \in \mathbf{M}^o$ (object mask), $\mathbf{F}_{\gamma}^k(i,j) = s_\gamma^o$. 

The three average errors that we get in (\ref{myeq2}) for the three saliency types can be used to analyze the performance of an algorithm in terms of how it corresponds to object saliencies at different levels of human perception, and hence, are extremely useful. However, we require a single average error using the three saliency types together, which can be used to directly evaluate an algorithm in comparison to another. We define the error $\text{MAE}^o$ on an object $o$ when all the three ground truth saliencies are considered together as follows:
\begin{equation}
\label{myeq3}
\text{MAE}^o =  \min_{\gamma \in \{ et, pc, rd\}} \text{MAE}_{\gamma}^o
\end{equation}
and this $\text{MAE}^o$ is then used in (\ref{myeq2}) instead of $\text{MAE}_{\gamma}^o$ to get the corresponding average error $\widehat{\text{MAE}}$ over all objects. As can be seen in (\ref{myeq3}), for an object, the minimum of the three errors is considered. This signifies that we do not penalize an approach for not sufficiently conforming with two of the three ground truth saliencies when it does with the third, considering each object independently. In effect, we suggest that a salient object detection approach performs well when it estimates accurately the object saliency at any single human perceptual level, which allows us to fairly compare different algorithms that may contain modules inspired by different phenomena (at different levels) in human vision.

Instead of the average error over all objects in the dataset, we can also obtain the average error for an image by considering only the objects in it while using (\ref{myeq2}) and (\ref{myeq1}), with the number of objects in the image (say $N_I$) in place of $N$. {We perform object-wise MAE computation and not pixel-wise, as we intend to evaluate the single saliency level of an object predicted by an algorithm. In object saliency estimation, an object irrespective of its size is a single entity. So, pixel-wise MAE will be strongly biased towards large objects, and will particularly neglect grave errors in estimating the saliency levels of small but important objects.} In Figures~\ref{fig:multiLevelComparison}(a), (b), and (c), we illustrate the $\widehat{\text{MAE}}_{\gamma}$ values of the mentioned existing salient-object detectors and in Figure~\ref{fig:multiLevelComparison0}(b), we show their $\widehat{\text{MAE}}$ values.

{\subsection{Average area under Precision-Recall Curve}
\label{AUCss}
Evaluation of salient object detection using binary images as ground truth is predominantly done based on precision-recall curves \cite{Han2018}, where the approaches are tested in classification performance. Here, we propose to use area under precision-recall curve (AuPRC) to evaluate classification performance at multiple saliency levels given by our multi-level ground truth images.

{In order to compute AuPRC, we binarize a multi-level ground truth image (of type $\gamma$) into multiple binary images by using the saliency levels of the objects in that image as thresholds. An example of this binarization is illustrated in Figure~\ref{fig:aucVisualization}}.



\begin{figure}[!h]
	\centering
	\subfloat[Original image]{\label{fig:original}\includegraphics[width=4.2cm]{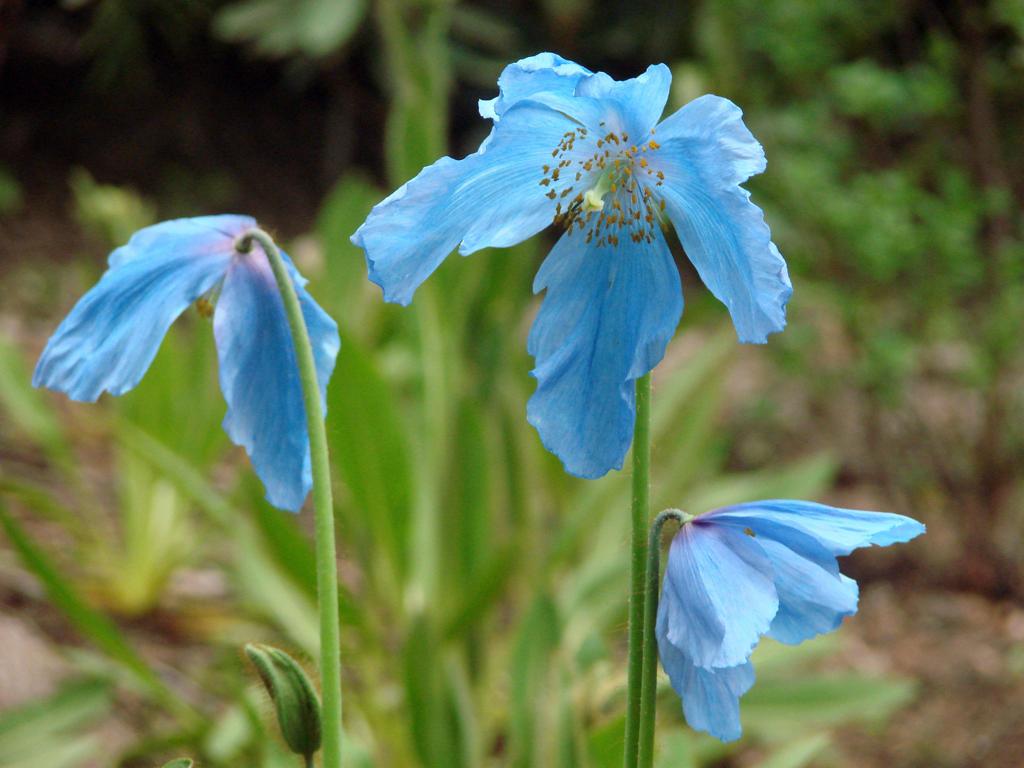}}\hfill
	\subfloat[Eye-Tracking GT]{\label{fig:eyeTrackingGT}\includegraphics[width=4.2cm]{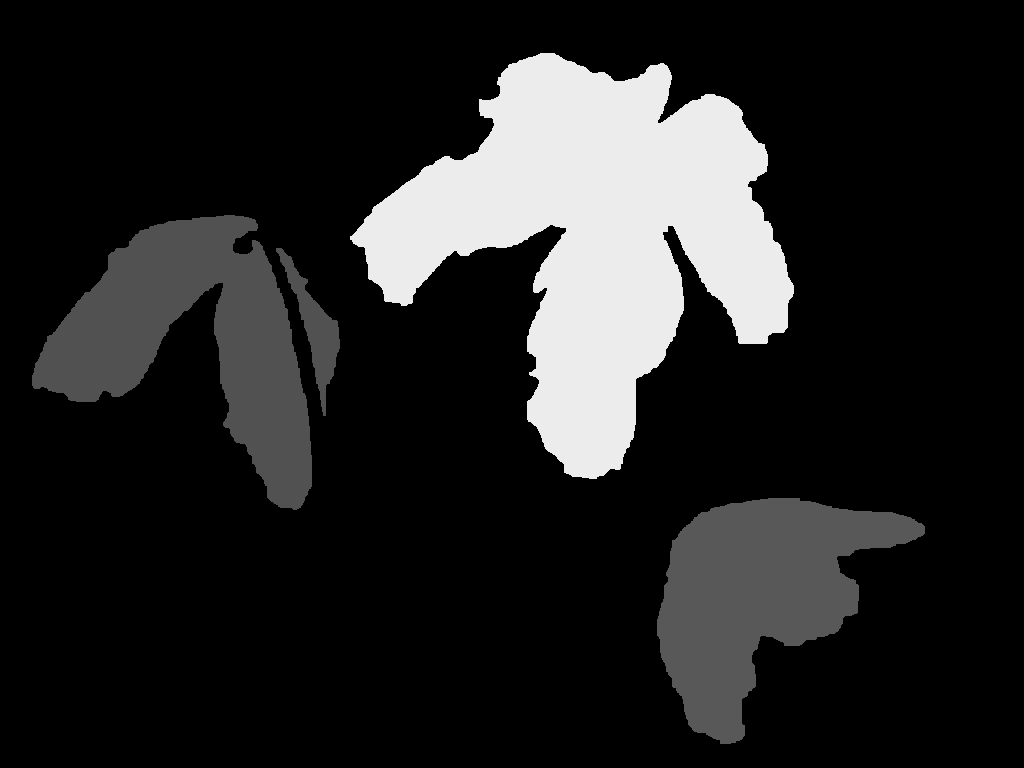}}\\
	\subfloat[Saliency level \#1]{\label{fig:SL1}\includegraphics[width=2.6cm]{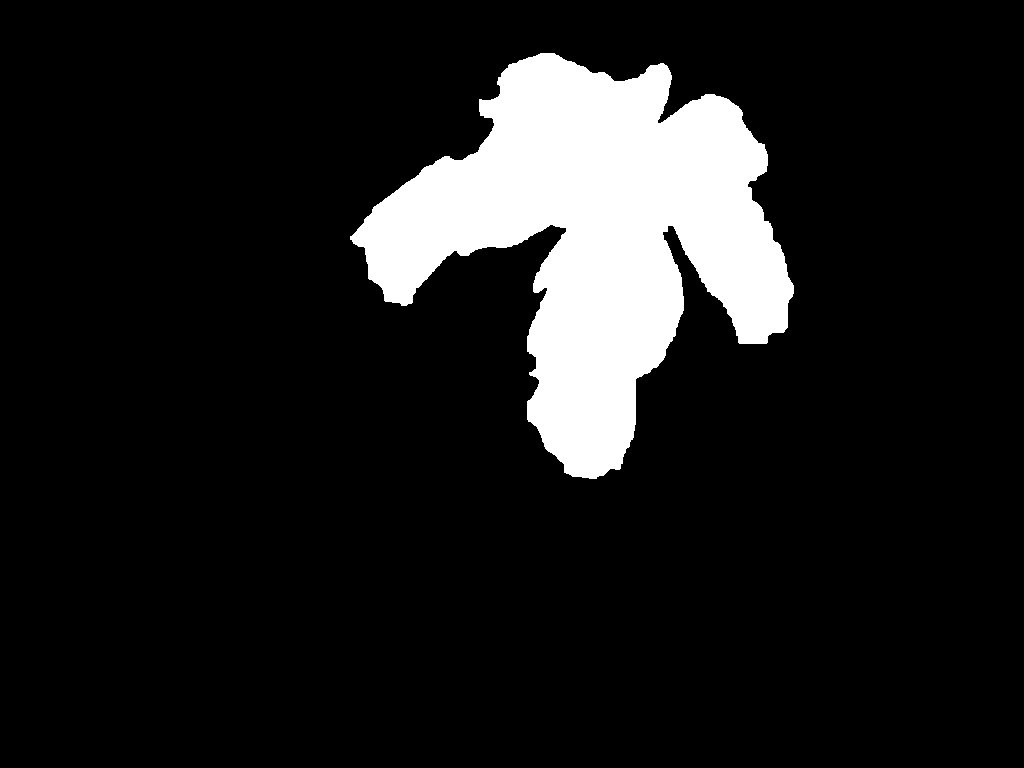}}\hfill
	\subfloat[Saliency level \#2]{\label{fig:SL2}\includegraphics[width=2.6cm]{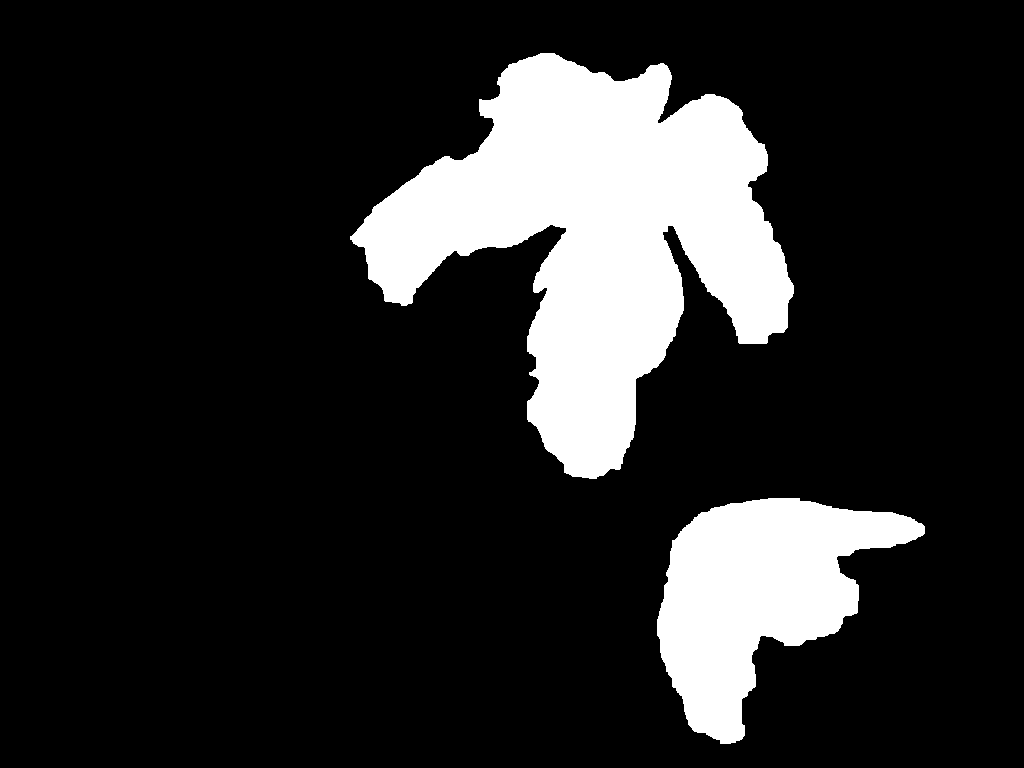}}\hfill
	\subfloat[Saliency level \#3]{\label{fig:SL3}\includegraphics[width=2.6cm]{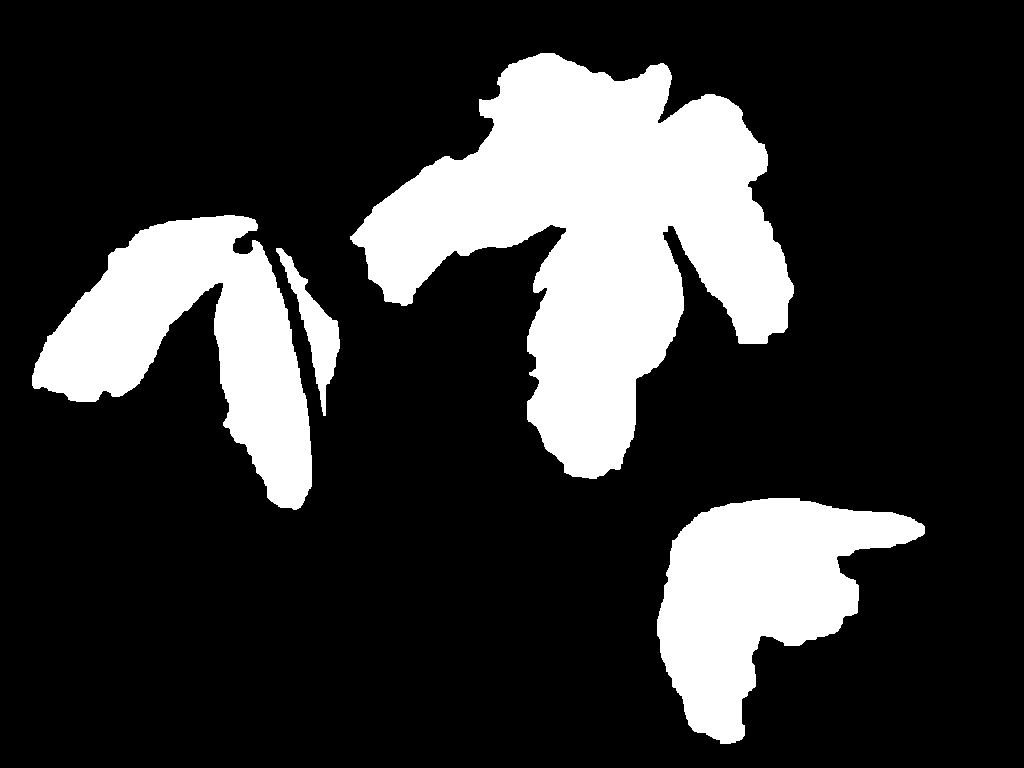}}\hfill
	\caption{Generating binary maps by thresholding a multi-level saliency ground truth. Each image has as many binary maps as there are salient objects.}
	\label{fig:aucVisualization}
\end{figure}

{Let $A_\gamma^{(k,l)}$ be the AuPRC that is calculated using the binary map of type $\gamma$ for image $k$ at saliency level $l$. Then we compute:
\begin{eqnarray}
\label{auc2}
\widehat{A}_\gamma &=& \frac{1}{N} \sum_{o} A_\gamma^{o},\ \ o=(k,l)
\end{eqnarray}}
where, $\widehat{A}_\gamma$ gives the average AuPRC over all objects for ground truth saliency type $\gamma \in \{ et, pc, rd\}$. In a manner similar to the case of MAE computation, we compute a single average AuPRC value $\widehat{A}$ considering all the three ground truth saliency types together as follows:
\begin{equation}
\label{auc3}
    \widehat{A} = \frac{1}{N} \sum_{\forall o}\max_{\gamma \in \{ et, pc, rd\}}A_\gamma^o
\end{equation}
where, we do not penalize an approach for not sufficiently conforming with two of the three ground truths when it does with the third, whose appropriateness has been discussed in Section~\ref{MAEss}.

In Figures~\ref{fig:multiLevelComparison}(a), (b), and (c), the $\widehat{A}_\gamma$ values of the existing salient object detectors are given and in Figure~\ref{fig:multiLevelComparison0}(a), the $\widehat{A}$ values are shown. Just like case of MAE computation, one can check performance on an image by computing average AuPRC for all the objects only in the image using (\ref{auc2})-(\ref{auc3}), accordingly. In the above explanation of AuPRC usage for multi-level ground truth, it should be noted that the consideration of thresholds as shown in {Figure~\ref{fig:aucVisualization}} after ordering is {analogous} with the thresholding performed in the standard precision-recall curve computation.}

\subsection{Salient Object Ranking Correlation}

Salient objects detected by an approach are usually used in further analysis performing image understanding and recognition. In many such cases, the highest to lowest ranking of the salient objects are more important than the absolute saliency values, as preferences in further analysis are given in that order. Therefore, we consider evaluation based on relative importance of salient objects in our dataset by ranking them with respect to their saliencies from highest to lowest.

\begin{figure*}[!h]
	\centering
	\newcommand{\imwidth}{0.33\textwidth}
	\subfloat[]{\label{fig:Ets}\stackunder[2pt]{\includegraphics[width=\imwidth]{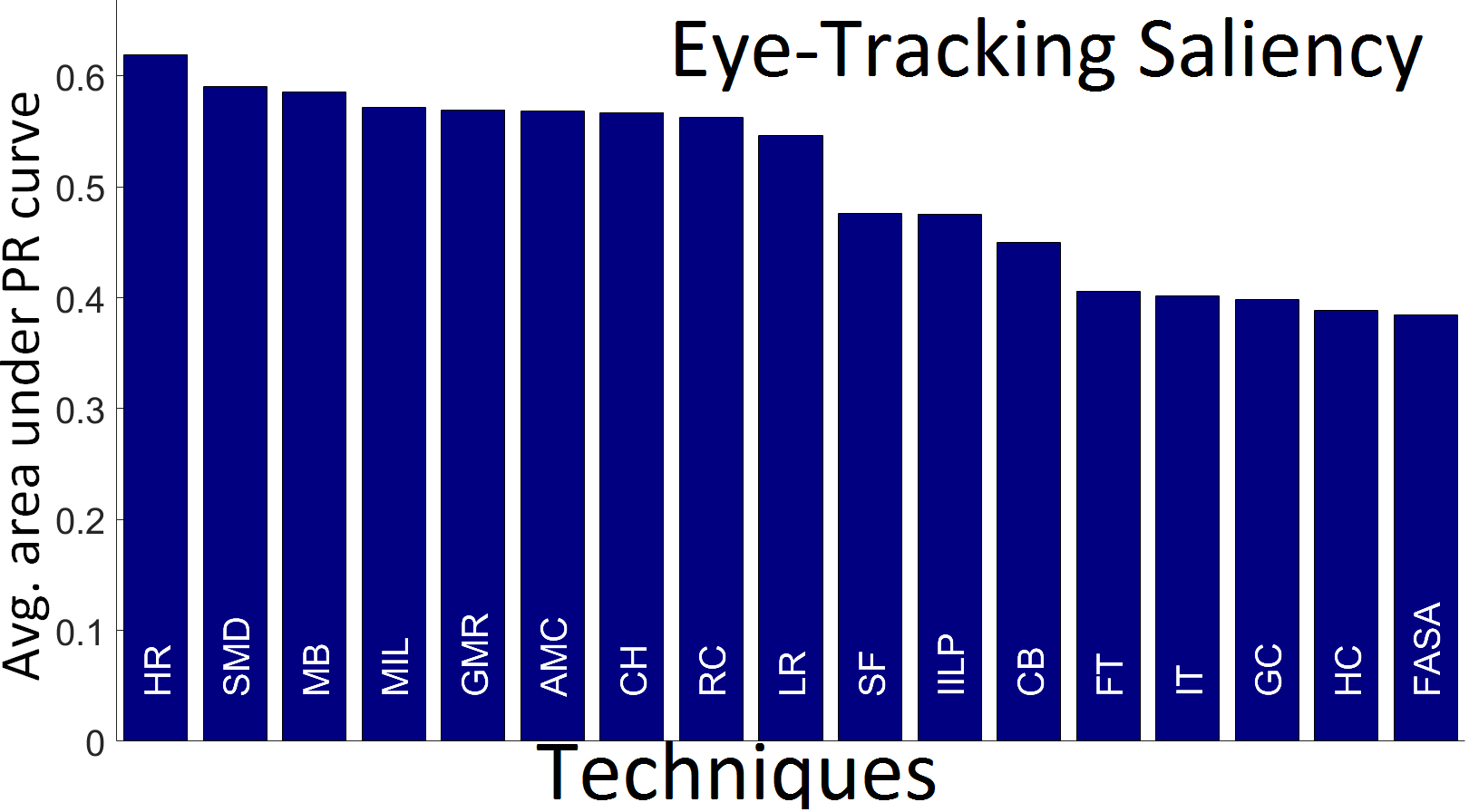}}{AuPRC}\stackunder[2pt]{\includegraphics[width=\imwidth]{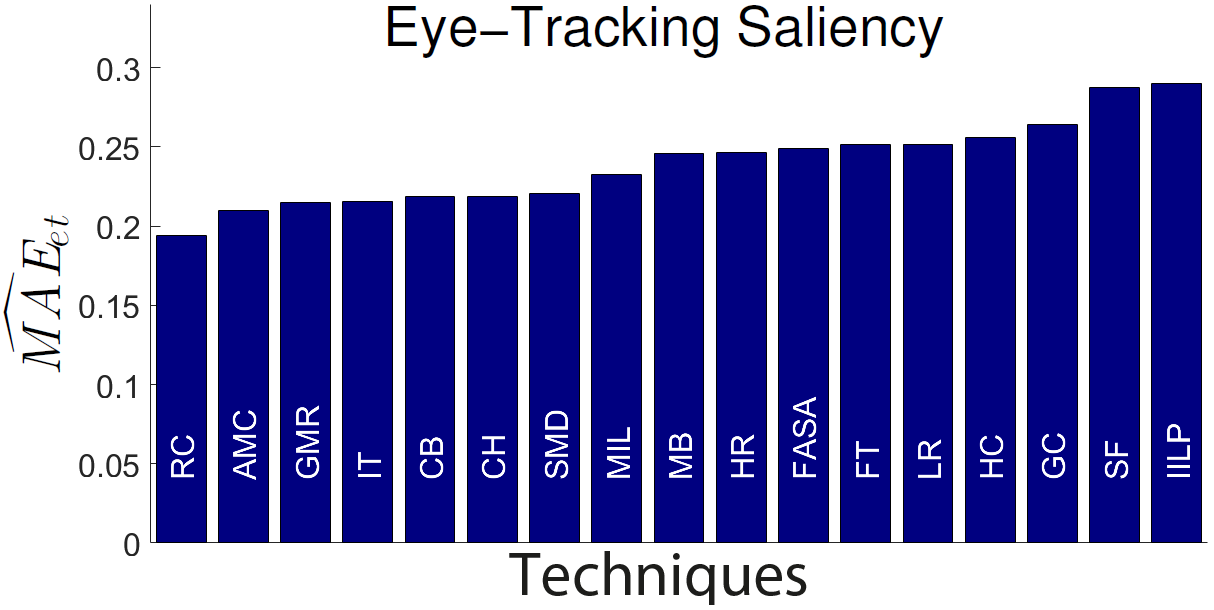}}{MAE}\stackunder[2pt]{\includegraphics[width=\imwidth]{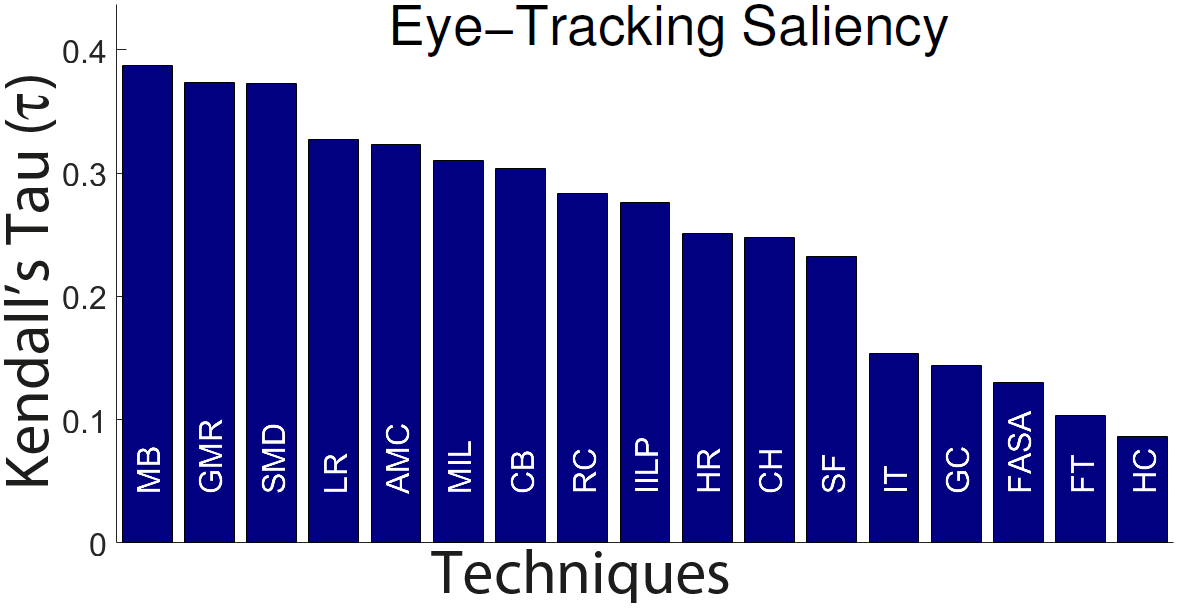}}{Kendall's tau}}\\
	\subfloat[]{\label{fig:Pcs}\stackunder[2pt]{\includegraphics[width=\imwidth]{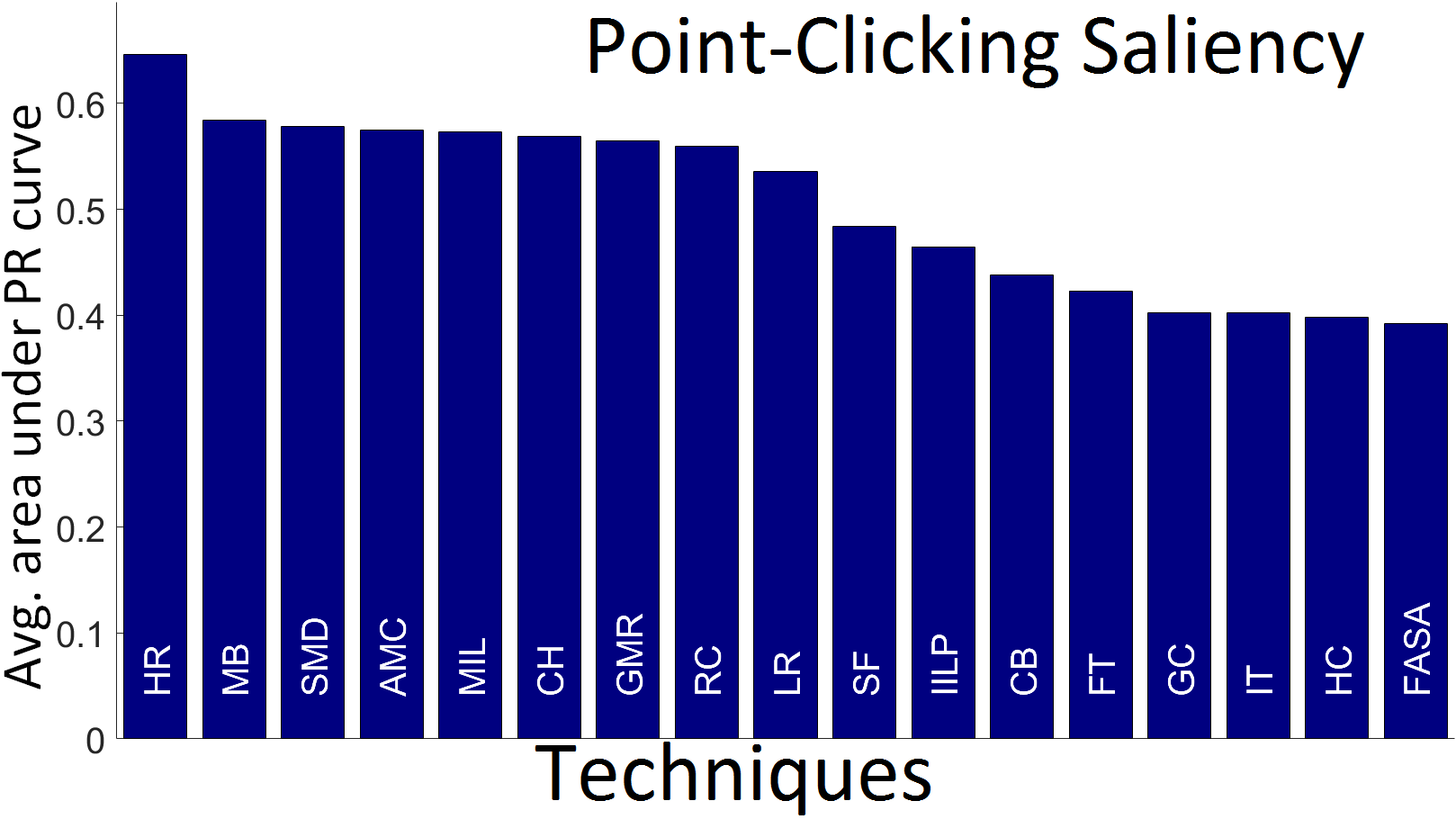}}{AuPRC}\stackunder[2pt]{\includegraphics[width=\imwidth]{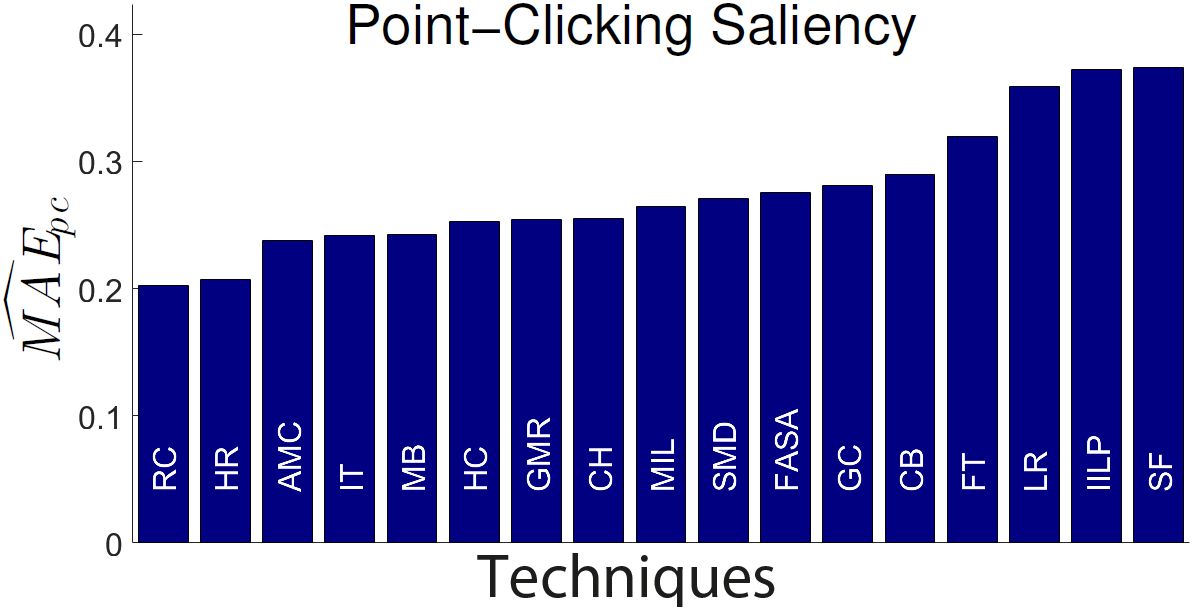}}{MAE}\stackunder[2pt]{\includegraphics[width=\imwidth]{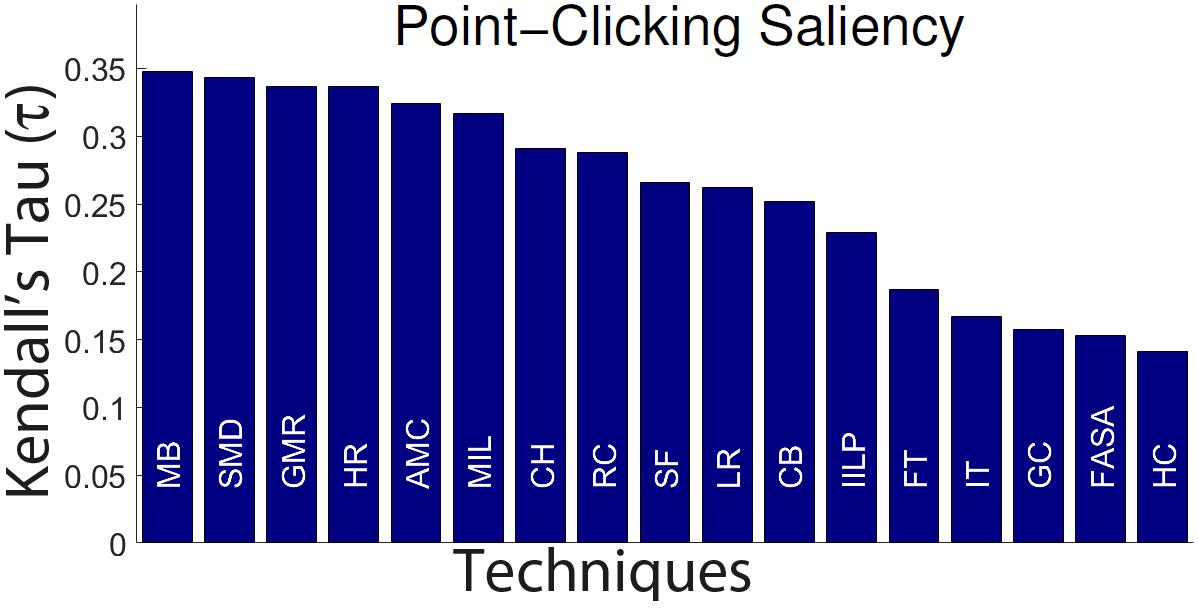}}{Kendall's tau}}\\
	\subfloat[]{\label{fig:Rds}\stackunder[2pt]{\includegraphics[width=\imwidth]{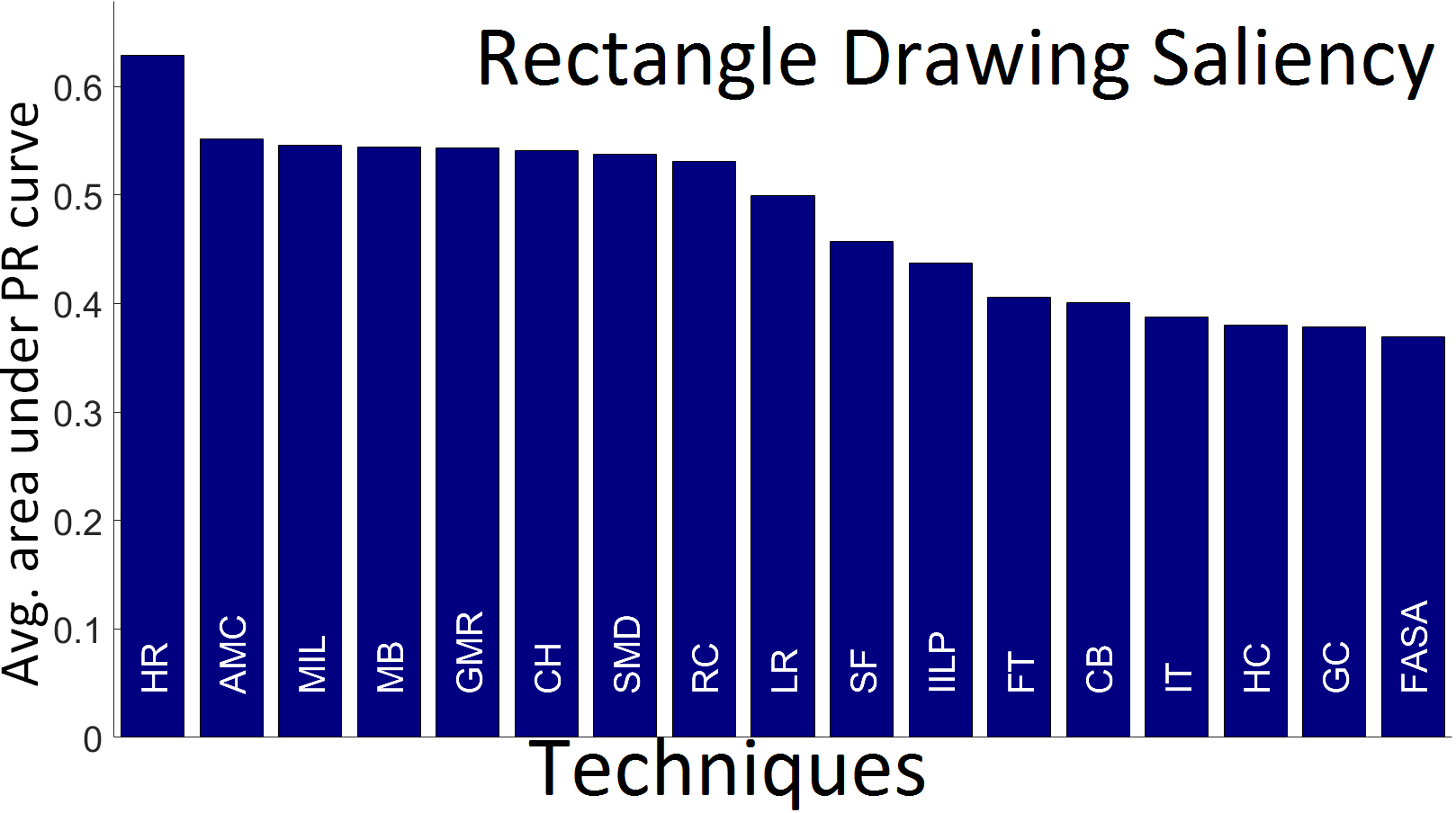}}{AuPRC}\stackunder[2pt]{\includegraphics[width=\imwidth]{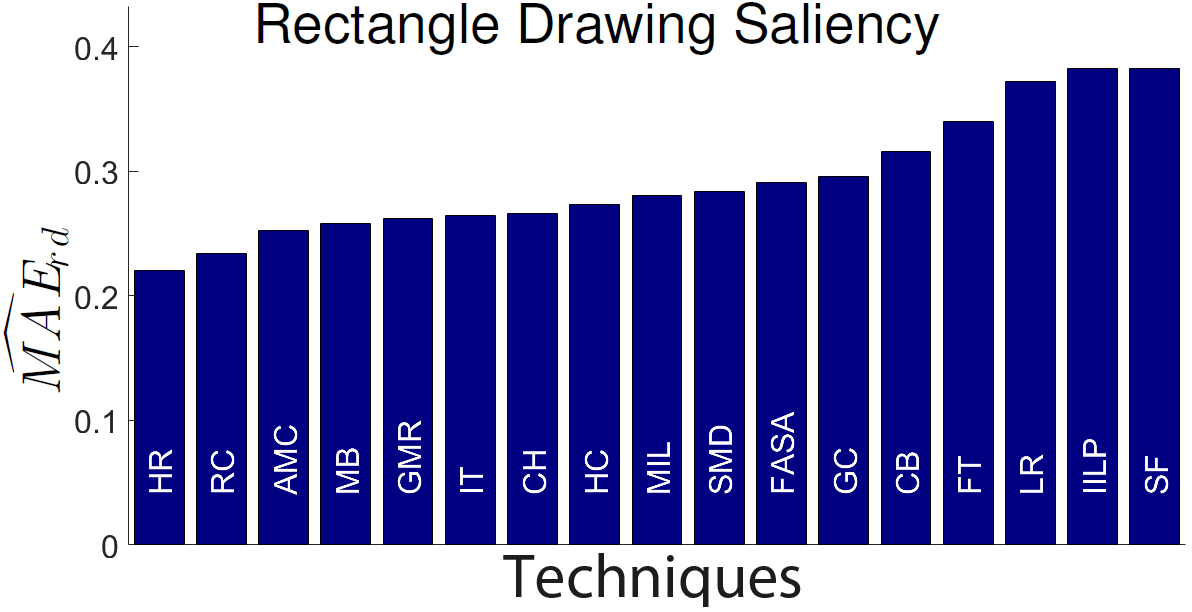}}{MAE}\stackunder[2pt]{\includegraphics[width=\imwidth]{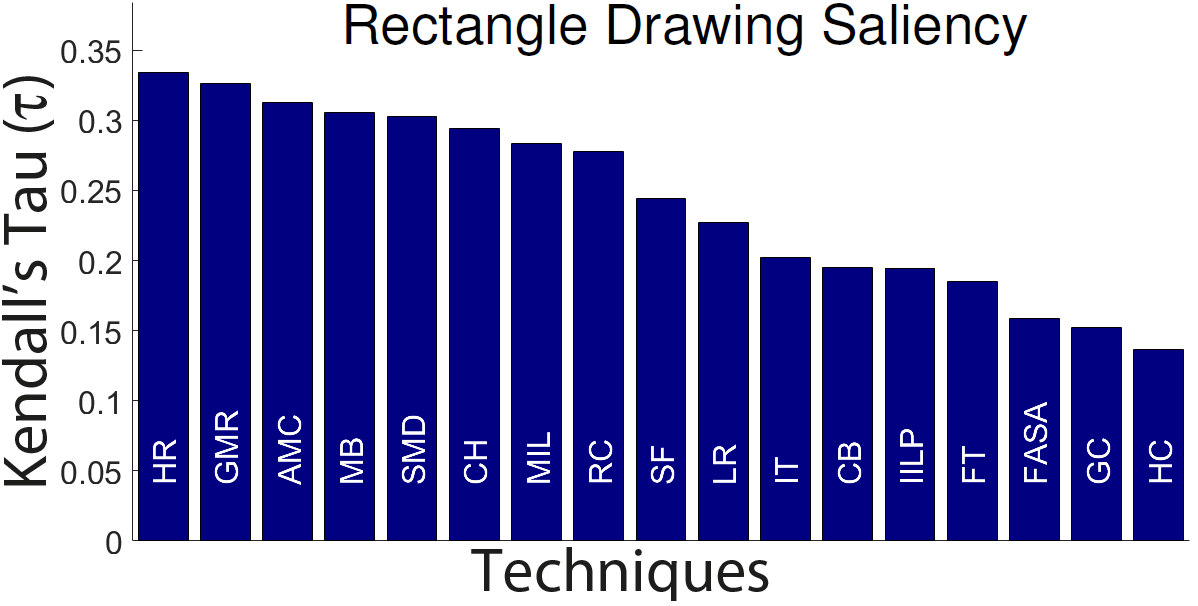}}{Kendall's tau}}
	\caption{Average area under precision-recall curve (AuPRC), mean absolute-errors (MAE) and salient-object ranking correlations (Kendall's tau) of the existing methods on (a) eye-tracking, (b) point-clicking, and (c) rectangle-drawing saliency ground truths}
	\label{fig:multiLevelComparison}
\end{figure*}

For the evaluation, we consider the correlation between the salient object ranking based on the gray-level saliency values provided by an algorithm and that based on the ground truth gray-level saliencies of type $\gamma$ by computing Kendall rank-correlation coefficient (Kendall's $\tau_b$)~\cite{Kendall} between them. In Figures~\ref{fig:multiLevelComparison}(a), (b), and (c), the Kendall's tau {($\tau^\gamma_b$)} values for the existing methods considering the three different ground truths are shown.

Now, similar to {Sections~\ref{MAEss} and~\ref{AUCss}}, we require a single correlation value using the three saliency ground truths together, which can be used to directly evaluate an algorithm in comparison to another. For this, we introduce a slight modification in Kendall's tau computation to get a correlation measure which considers all the three ground truth saliencies together. {Similar to the MAE and AUC computation cases}, the modification of Kendall's tau by considering all the three ground truths together must indicate that a salient object detection approach performs well when it estimates accurately the object saliency rank at any single human perceptual level. That is, an approach is not penalized for not sufficiently conforming with two of the three ground truth saliencies when it does with the third, whose appropriateness has already been pointed out in Section~\ref{MAEss}.

{Let the set of type-$\gamma$ ground truth gray-level saliencies of all objects in our dataset be represented by $\rho_{\gamma}$, that is, $s_\gamma^o\in\rho_{\gamma}, \forall o$. Further, let the set of gray-level saliencies of all the corresponding objects estimated by an algorithm be $R$. 
The gray-level object saliency estimated by an algorithm for an object $o$ is given in (\ref{myeq0}), and therefore, $\mathrm{S}^o\in R, \forall o$. 
{Kendall's tau $\tau^\gamma_b$ for each type of ground truth is calculated between $R$ and $\rho_\gamma$ using the standard way of Kendall's tau computation as given in \cite{Kendall}. 

On the other hand, Kendall's tau ($\tau_b$) considering the three ground truth types together are obtained using a modified Kendall's tau computation as follows: 
\begin{eqnarray}
\label{kennew}
\tau_b=(C-D)\Big{/}\sqrt{(C+D+T_R)(C+D+T_\rho)}
\end{eqnarray}
where, with all estimated saliency pairs  $(\mathrm{S}^{x},\mathrm{S}^{y})$ and all ground truth saliency pairs $(s_\gamma^x, s_\gamma^y)$ employed, the $C$ and $D$ are numbers of concordant and discordant pairs, respectively, and the $T_R$ and $T_\rho$ are numbers of tied pairs in estimated and ground truth saliencies, respectively. The values of the parameters in (\ref{kennew}) are computed as given below:

\newcommand{\CG}[1]{G^s_{x#1y}}
\newcommand{\DG}[1]{G^i_{x#1y}}
\newcommand{\SG}[1]{E_{x#1y}}

\begin{align}
\label{kennew1}
C &= \sum_{x,y} \CG{>} \cdot \SG{>} + \CG{<} \cdot  \SG{<}\\
\label{kennew2}
D &= \sum_{x,y}  \CG{<} \cdot \DG{\le} \cdot \SG{>}+ \CG{>} \cdot \DG{\ge} \cdot \SG{<}\\
\label{kennew3}
T_\rho &=  \sum_{x,y} \DG{=} \cdot (\SG{>} + \SG{<})\\
\label{kennew4}
T_R &=  \sum_{x,y} \CG{\ne} \cdot \SG{=}
\end{align}
with
\begin{align}
\nonumber
\CG{\bullet} &= \max_\gamma \Delta(s_\gamma^x \bullet s_\gamma^y)\\
\nonumber
\DG{\bullet} &= \min_\gamma \Delta(s_\gamma^x \bullet s_\gamma^y)\\
\nonumber
\SG{\bullet} &= \Delta(S^x \bullet S^y)
\end{align}
where, $\bullet \in \{<, >, \le, \ge, =, \ne\}$ is a generic operator and $\Delta(.)$ is an indicator function, which takes the value $1$ when the condition is satisfied and $0$ otherwise.
}


In our modification of Kendall's tau computation, to ensure that an approach is not penalized when it conforms with at least one of the three ground truths, the number of concordant pairs $C$ is incremented in (\ref{kennew1}) when the concordance happens for at least one ground truth type. Further, the number of discordant pairs $D$ is incremented in (\ref{kennew2}) only when there is no agreement with respect to all the three ground truths. The increments in $T_\rho$ and $T_R$ in (\ref{kennew3}) and (\ref{kennew4}) are deduced logically from the increment strategies of $C$ and $D$.}

\begin{figure*}[!h]
	\centering
	\newcommand{\imwidth}{0.33\textwidth}
	\subfloat[]{\label{fig:aucfull}\includegraphics[width=\imwidth]{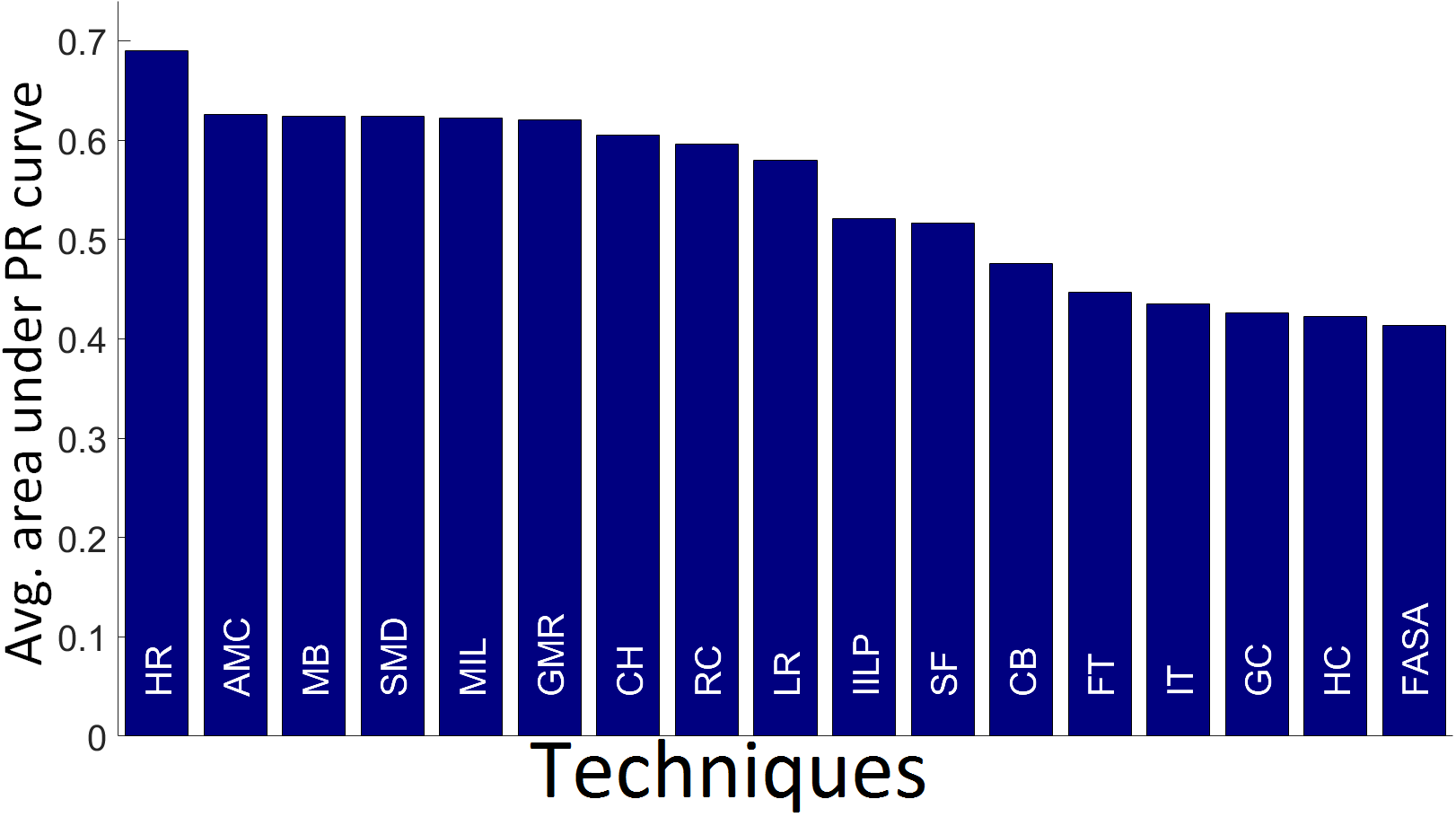}}
	\subfloat[]{\label{fig:maefull}\includegraphics[width=\imwidth]{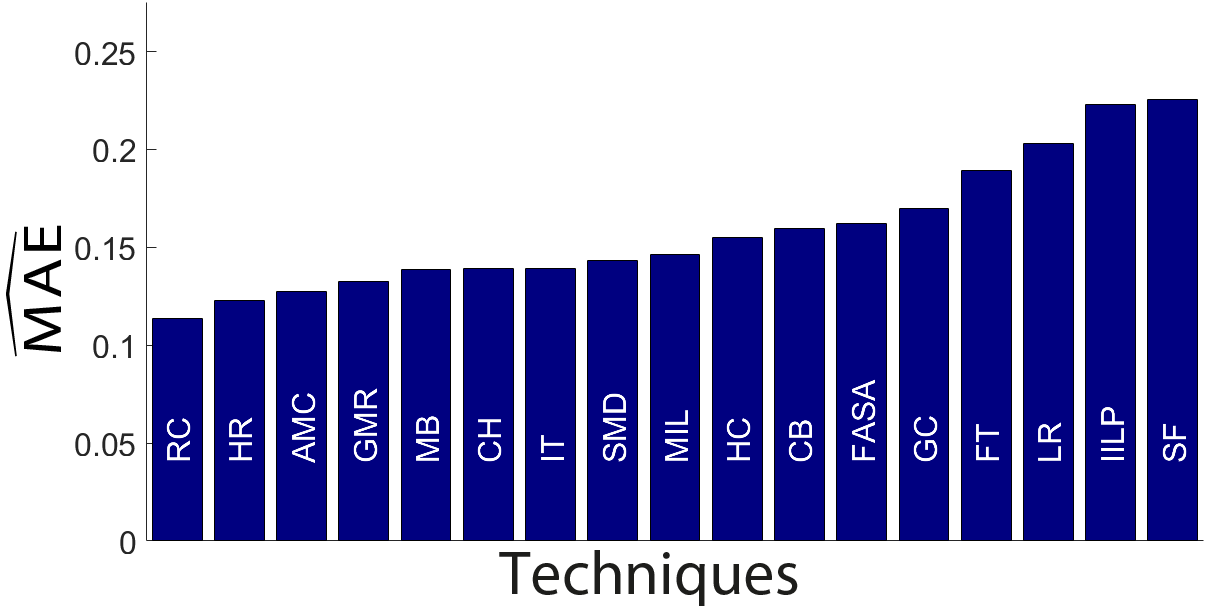}}
	\subfloat[]{\label{fig:corrfull}\includegraphics[width=\imwidth]{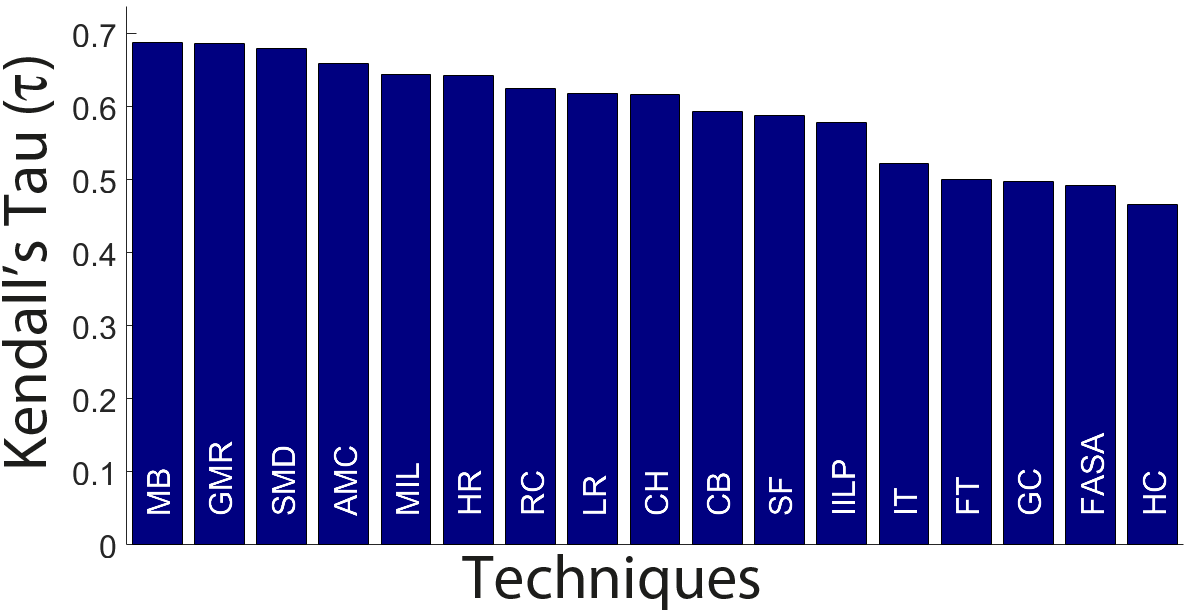}}
	\caption{(a) Average area under precision-recall curves (b) mean absolute errors and (c) salient-object ranking correlations of the existing methods obtained considering the three types of ground truth saliency maps together.}
	\label{fig:multiLevelComparison0}
\end{figure*}

Figure~\ref{fig:multiLevelComparison0}(c) shows the modified Kendall's tau ($\tau_b$) measure values for the existing methods considering the three ground truths together. Instead of computing the correlations considering all objects in the dataset, one can obtain the correlations considering only the objects in an image to check the performance on it.


\newcolumntype{C}[1]{>{\centering}m{#1}}
\newcommand{\tww}{0.7cm}
\newcommand{\tw}{0.56cm}
\begin{figure}[t]
	\centering
	\includegraphics[width=8.4cm]{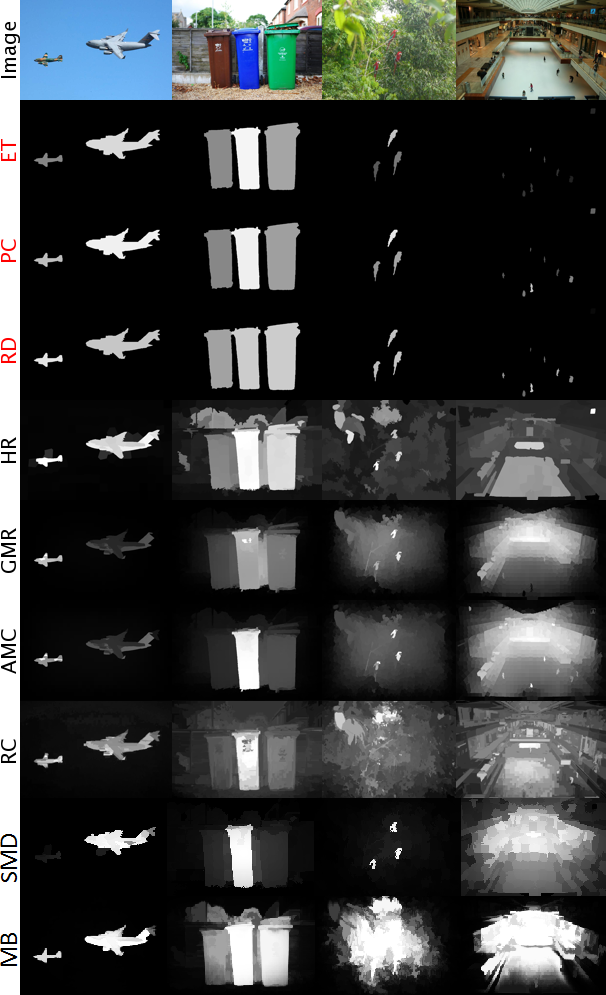}
	\caption{The multi-level ground truth saliency maps and the saliency maps that are estimated by a few well-performing methods on a few images where their qualitative performance significantly vary. Here, ET, PC, and RD indicate the ground truth maps that are generated using the eye-tracking, point-clicking, and rectangle-drawing subjective data, respectively.}
	\label{fig:imageComparison}
\end{figure}

\subsection{Discussion}

In Figures~\ref{fig:multiLevelComparison}(a), (b), (c) and Figure~\ref{fig:multiLevelComparison0}(b), which shows the MAE comparisons, we find that some of the approaches such as RC, HR, AMC do better in most of the cases than the others. Although from the values, the differences seem little, they are significant as they are obtained considering 2434 objects in 588 images. As can be seen, there are a few cases where an approach that performs well with respect to a ground truth, does not do well for the others, and vice versa. For example, HR does not do well for the eye-tracking saliency ground truth unlike for the others, and GMR appears among the top 3 only for the eye-tracking saliency ground truth. This is expected, as intentionally or not, approaches may be designed such that they work similar to that at a narrow band of perception from the entire zero (low) to high spectrum. That is why, Figure~\ref{fig:multiLevelComparison0}(a) showing the MAE results when the three ground truths are considered together is essential. Note that the MAE values of all the approaches in Figure~\ref{fig:multiLevelComparison0}(b) are lower compared to their values in Figures~\ref{fig:multiLevelComparison}(a), (b) and (c) as a consequence of being fair by removing the bias of the evaluation in terms of perceptual level.

In Figures~\ref{fig:multiLevelComparison}(a), (b), (c) and Figure~\ref{fig:multiLevelComparison0}(c), which shows the Kendall's tau comparisons, we again find that some of the approaches such as GMR, MB, SMD do better in most of the cases than the others. Consider the top 3 performing approaches for the rectangle-drawing saliency. We see that two of them, namely HR and AMC appear way below in performance with respect to the other ground truth saliencies. Considering the ground truths separately, we cannot infer whether their good performance for the rectangle-drawing saliency is significant or not. Now, in terms of performance evaluation based on all the ground truths together, HR and AMC does not appear in top 3. This makes us capable to infer that the concerned case of good performance in one of the ground truth saliencies is not significant. This precisely shows how removing the perceptual level bias in the evaluation is appropriate and useful. It can also be noted that the Kendall's tau values of all the approaches in Figure~\ref{fig:multiLevelComparison0}(c) are higher compared to their values in Figures~\ref{fig:multiLevelComparison}(a), (b) and (c), an effect similar to what was seen with the MAE values.

When we compare using MAE and Kendall's tau in Figures~\ref{fig:multiLevelComparison} and~\ref{fig:multiLevelComparison0}, we observe that the order of performance of the existing methods is significantly different. Mean absolute error and salient object ranking correlation measure salient object detection performance very differently. In Table~\ref{tab:metricsDiffer}, two hypothetical cases for multi-level salient-object detection is presented. In each case, there are two salient objects with different ground truth (GT) and estimated saliency values. In the first case, the mean absolute error measures a small error, which indicates good performance. On the other hand, due to the change in the order of saliency values, the salient object ranking correlation evaluates an inverse correlation, that is, a bad performance. In the second case, observation of the evaluations is reversed. This example shows that we should use a measure depending on how we intend to apply the multi-level saliency values in a task.

Given two salient objects, absolute saliency values not only measure which one is more salient, but also quantify the difference in saliency. This approach can be useful in applications such as content-aware image compression, because absolute saliency values are directly used as an indicator to compression ratio. Salient object ranking is useful when the ordering of salient objects is more important than their actual saliency values. For example, in image tagging, processing the most important object would yield more relevant image tags compared to a randomly selected object.

\begin{figure}[!h]
	\centering
	\newcommand{\imwidth}{0.95\columnwidth}
	\includegraphics[width=\imwidth]{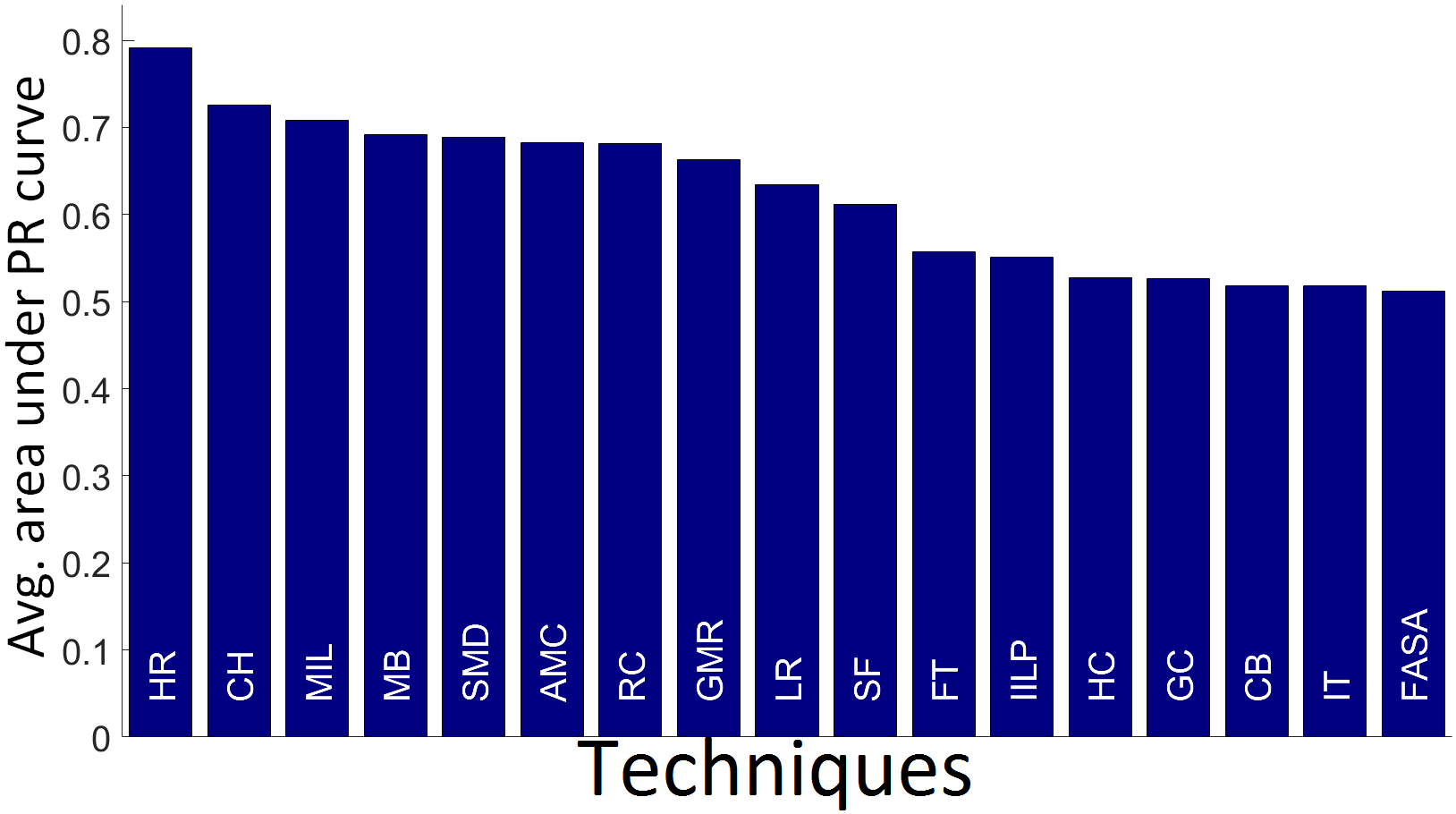}
	\caption{Average area under precision-recall curves  of the existing methods obtained on binary ground truth of our dataset images containing all salient object in one salient class.}
	\label{fig:BGTComparison}
\end{figure}

In literature, a few recently proposed existing approaches which {we have considered} in our analysis have been shown to outperform other existing ones based on binary ground truth datasets, and hence, they represent the state-of-the-art. In fact, most approaches for salient object detection are targeted to perform well on such binary datasets when evaluated using precision and recall measures. {Therefore, the recently developed approaches may not perform well on our multi-level ground truth dataset when evaluated using MAE and Kendall' tau, and hence, it would be appropriate to also evaluate existing approaches on binary maps generated from our multi-level ground truth saliencies. 

The average AuPRC based comparisons on our multi-level ground truth dataset shown in Figures~\ref{fig:multiLevelComparison}(a), (b), (c) and Figure~\ref{fig:multiLevelComparison0}(a) consider performance on binary maps generated from the multi-level ground truth maps. We find that some of the existing approaches such as HR, MB and AMC do, in general, better than others. Similar to the cases of MAE and Kendall's tau based comparisons, we see here as well that most top performing approaches tend to do well in a particular type of ground truth and not in all types. Hence, the average AuPRC values shown in Figure~\ref{fig:multiLevelComparison0}(a) considering the three ground truths together while removing perceptual bias is useful for evaluation. Note that the average AuPRC values of all the approaches
in Figure~\ref{fig:multiLevelComparison0}(a) are higher compared to their values in Figures~\ref{fig:multiLevelComparison}(a), (b), (c), a result of being fair by removing perceptual bias. Like the disagreement between MAE and Kendall's tau demonstrated using Table~\ref{tab:metricsDiffer}, average AuPRC can also differ from both the MAE and Kendall's tau values, which would depend on the comparative estimated saliency values and sizes of the objects in the images.

It is interesting to find that approaches such as HR~\cite{Yildirim_saliency_detection}, AMC~\cite{Jiang_saliency_detection}, MB~\cite{Zhang2015} and GMR~\cite{Yang_saliency_detection}, which consider salient object detection in images as a local to global hierarchical processing of a graph, in general, perform well in our multi-level saliency dataset.} However, a couple of the recently proposed existing approaches do not perform well on our ground truth dataset, {even when evaluated using AuPRC}, and hence, may not be suitable for images with objects having multi-level saliency. {In Figure~\ref{fig:BGTComparison}, we provide the average AuPRC results that evaluate the approaches considered in Figure~\ref{fig:multiLevelComparison0}(a) on the binary ground truth obtained for our dataset images by considering all objects in them to be equally salient. As can be seen, some of the approaches perform well on the binary ground truth in comparison to others, although they did not do so on our multi-level saliency ground truth, and vice-versa. Note that, as expected, the average AuPRC results are higher for the binary ground truth than the multi-level ground truth.} Now, as our three evaluation measures can be used to assess multi-level salient object detection performance, designing of approaches in future can focus on detecting multiple salient objects with varying saliencies, which can be followed with further processing to yield binary (salient/non-salient) salient object detection outputs, if required, and this would be akin to how humans perform ``salient object detection''. Gray-level object saliency maps obtained using a few approaches, which have performed well in our dataset, on a few images are shown in Figure~\ref{fig:imageComparison} along with the corresponding multi-level ground truth saliency maps.

\begin{table}[h]
	\centering
	\caption{Two hypothetical cases, where two evaluation measures significantly differ}
	\label{tab:metricsDiffer}
	\begin{tabular}{|l|c|c|}
		\hline
		& Case \#1 & Case \#2\\
		\hline
		GT Saliency Value \#1 & 0.48 & 0.3 \\
		GT Saliency Value \#2 & 0.52 & 0.8 \\
		\hline
		Estimated Saliency Value \#1 & 0.51 & 0.0 \\
		Estimated Saliency Value \#2 & 0.49 & 0.5 \\
		\hline
		Mean Absolute Error & 0.03 & 0.3\\
		\hline
		Salient-Object Ranking & -1 & 1\\
		\hline
	\end{tabular}
\end{table} 

\vspace{-0.1cm}
\section{Conclusion}
\label{sec:conclusion}

In this paper, subjective experiments have been conducted on natural images to {confirm} that all objects are inherently not equally important and salient object detection should be evaluated considering this aspect. For this, we have introduced an image dataset with each image having multiple objects, and for each image there are three types of multi-level ground-truth maps. For generating the ground truths, the saliency level of an object has been measured by conducting three subjective experiments: eye tracking, point clicking, and rectangle drawing. While eye-tracking experiments capture the effect of spontaneous attention and eye-fixation duration on object saliency, point-clicking and rectangle-drawing experiments measure the role of human perception at different levels. From this, we have shown that object saliency is inherently multi-level. Further, performance evaluation measures are given to evaluate multi-level salient object detection, and existing saliency detection approaches, some of which represent the state-of-the-art, are compared. {Approaches that perform a local to global hierarchical processing of images considering them as graphs are found to perform well in our dataset.} The dataset and measures reported in this paper add a different dimension to the evaluation of salient object detection designed for application in object localization, generic target detection, visual media description, compression and segmentation.

\bibliographystyle{iet}
\bibliography{literature}

\begin{thebibliography}{10}

\bibitem{Baluch11}
Baluch, F., Itti, L.: `Mechanisms of top-down attention', \emph{Trends in
  Neurosciences},  2011, \textbf{34}, (4), pp.~210--224

\bibitem{Neibur98}
Niebur, E., Koch, C.
\newblock `Computational architectures for attention'.
\newblock In: Parasuraman, R., editor. The Attentive Brain. (Cambridge, MA,
  USA: MIT Press,  1998.

\bibitem{Han2018}
{Han}, J., {Zhang}, D., {Cheng}, G., {Liu}, N., {Xu}, D.: `Advanced
  deep-learning techniques for salient and category-specific object detection:
  A survey', \emph{IEEE Signal Processing Magazine},  2018, \textbf{35}, (1),
  pp.~84--100

\bibitem{Achanta_frequency_tuned}
Achanta, R., Hemami, S., Estrada, F., S{\"u}sstrunk, S.
\newblock `Frequency-tuned salient region detection'.
\newblock In: Proceedings of IEEE CVPR. (,  2009. pp.~ 1597 -- 1604

\bibitem{Alpert_image_segmentation}
Alpert, S., Galun, M., Basri, R., Brandt, A.
\newblock `Image segmentation by probabilistic bottom-up aggregation and cue
  integration'.
\newblock In: Proceedings of IEEE CVPR. (,  2007. pp.~ 1--8

\bibitem{Movahedi_design_and}
Movahedi, V., Elder, J.H.
\newblock `Design and perceptual validation of performance measures for salient
  object segmentation'.
\newblock In: Proceedings of IEEE CVPR Workshops. (,  2010. pp.~ 49--56

\bibitem{Sal_secret}
Li, Y., Hou, X., Koch, C., Rehg, J.M., Yuille, A.L.
\newblock `The secrets of salient object segmentation'.
\newblock In: IEEE Conference on Computer Vision and Pattern Recognition. (,
  2014. pp.~ 280--287

\bibitem{fan2018SOC}
Fan, D.P., Cheng, M.M., Liu, J.J., Gao, S.H., Hou, Q., Borji, A.
\newblock `Salient objects in clutter: Bringing salient object detection to the
  foreground'.
\newblock In: European Conference on Computer Vision (ECCV). (Springer,  2018.

\bibitem{Yang_saliency_detection}
Yang, C., Zhang, L., Lu, H., Ruan, X., Yang, M.
\newblock `Saliency detection via graph-based manifold ranking'.
\newblock In: Proceedings of IEEE CVPR. (,  2013. pp.~ 3166--3173

\bibitem{Zhou2019}
{Zhou}, Y., {Mao}, A., {Huo}, S., {Lei}, J., {Kung}, S.: `Salient object
  detection via fuzzy theory and object-level enhancement', \emph{IEEE
  Transactions on Multimedia},  2019, \textbf{21}, (1), pp.~74--85

\bibitem{Chen2018}
{Chen}, Y., {Zou}, W., {Tang}, Y., {Li}, X., {Xu}, C., {Komodakis}, N.: `Scom:
  Spatiotemporal constrained optimization for salient object detection',
  \emph{IEEE Transactions on Image Processing},  2018, \textbf{27}, (7),
  pp.~3345--3357

\bibitem{Henderson_eye_movements}
Henderson, J.M., Hollingworth, A.
\newblock `Eye movements during scene viewing: An overview'.
\newblock In: Eye Guidance while Reading and While Watching Dynamic Scenes. (,
  1998. pp.~ 269--293

\bibitem{Yarbus_eye_movements}
Yarbus, A.L.: `Eye Movements and Vision'.
\newblock (Springer,  1967)

\bibitem{Peters05}
Peters, R.J., Iyer, A., Itti, L., Koch, C.: `Components of bottom-up gaze
  allocation in natural images', \emph{Vision Research},  2005, \textbf{45},
  (18), pp.~2397 -- 2416

\bibitem{Wolfe_guided_search}
Wolfe, J.M.: `Guided search 2.0 a revised model of visual search',
  \emph{Psychonomic Bulletin \& Review},  1994, \textbf{1}, (2), pp.~202--238

\bibitem{Judd_learning_to}
Judd, T., Ehinger, K., Durand, F., Torralba, A.
\newblock `Learning to predict where humans look'.
\newblock In: Proceedings of IEEE ICCV. (,  2009. pp.~ 2106--2113

\bibitem{Li_visual_saliency}
Li, J., Levine, M.D., An, X., Xu, X., He, H.: `Visual saliency based on
  scale-space analysis in the frequency domain', \emph{IEEE Transactions on
  PAMI},  2013, \textbf{35}, (4), pp.~996--1010

\bibitem{DUTOMRON}
Ruan, X., Tong, N., Lu, H.
\newblock `How far we away from a perfect visual saliency detection -
  {DUT-OMRON}: a new benchmark dataset'.
\newblock In: Korea-Japan Joint Workshop on Frontiers of Computer Vision. (,
  2014.

\bibitem{Yildirim_fast_accurate}
Yildirim, G., S\"{u}sstrunk, S.: `{FASA}: Fast, accurate, and size-aware
  salient object detection', \emph{Proceedings of ACCV},  2014,

\bibitem{Cheng_global_contrast}
Cheng, M., Zhang, G., Mitra, N.J., Huang, X., Hu, S.
\newblock `Global contrast based salient region detection'.
\newblock In: Proceedings of IEEE CVPR. (,  2011. pp.~ 409--416

\bibitem{Jiang_automatic_salient}
Jiang, H., Wang, J., Yuan, Z., Liu, T., Zheng, N.: `Automatic salient object
  segmentation based on context and shape prior', \emph{Proceedings of BMVC},
  2011, pp.~ 1--12

\bibitem{Cheng_efficient_salient}
Cheng, M., Warrell, J., Lin, W., Zheng, S., Vineet, V., Crook, N.
\newblock `Efficient salient region detection with soft image abstraction'.
\newblock In: Proceedings of IEEE ICCV. (,  2013.

\bibitem{Perazzi_saliency_filters}
Perazzi, F., Krahenbuhl, P., Pritch, Y., Hornung, A.
\newblock `Saliency filters: Contrast based filtering for salient region
  detection'.
\newblock In: Proceedings of IEEE CVPR. (,  2012. pp.~ 733--740

\bibitem{Shen_a_unified}
Shen, X., Wu, Y.
\newblock `A unified approach to salient object detection via low rank matrix
  recovery'.
\newblock In: Proceedings of IEEE CVPR. (,  2012. pp.~ 853--860

\bibitem{Li_contextual_hypergraph}
Li, X., Li, Y., Shen, C., Dick, A., van~den Hengel, A.: `Contextual hypergraph
  modelling for salient object detection', \emph{Proceedings of IEEE ICCV},
  2013,

\bibitem{Jiang_saliency_detection}
Jiang, B., Zhang, L., Lu, H., Yang, M., Yang, C.: `Saliency detection via
  absorbing markov chain', \emph{Proceedings of IEEE ICCV},  2013,

\bibitem{Yildirim_saliency_detection}
Yildirim, G., Shaji, A., S\"{u}sstrunk, S.: `Saliency detection using
  regression trees on hierarchical image segments', \emph{Proceedings of IEEE
  ICIP},  2014,

\bibitem{Zhang2015}
Zhang, J., Sclaroff, S., Lin, Z., Shen, X., Price, B., Mech, R.
\newblock `Minimum barrier salient object detection at 80 fps'.
\newblock In: IEEE International Conference on Computer Vision. (,  2015. p.
  1404–1412

\bibitem{Li2015}
Li, H., Lu, H., Lin, Z., Shen, X., Price, B.: `Inner and inter label
  propagation: Salient object detection in the wild', \emph{IEEE Transactions
  on Image Processing},  2015, \textbf{24}, (10), pp.~3176--3186

\bibitem{Peng2017}
Peng, H., Li, B., Ling, H., Hu, W., Xiong, W., Maybank, S.J.: `Salient object
  detection via structured matrix decomposition', \emph{IEEE Transactions on
  Pattern Analysis and Machine Intelligence},  2017, \textbf{39}, (4),
  pp.~818--832

\bibitem{Huang2017}
Huang, F., Qi, J., Lu, H., Zhang, L., Ruan, X.: `Salient object detection via
  multiple instance learning', \emph{IEEE Transactions on Image Processing},
  2017, \textbf{26}, (4), pp.~1911--1922

\bibitem{islamsal18}
Islam, A., Kalash, M., Bruce, N.D.B.
\newblock `Revisiting salient object detection: Simultaneous detection,
  ranking, and subitizing of multiple salient objects'.
\newblock In: Computer Vision and Pattern Recognition (CVPR). (,  2018.

\bibitem{Liu_learning_to}
Liu, T., Yuan, Z., Sun, J., Wang, J., Zheng, N., Tang, X., et~al.: `Learning to
  detect a salient object', \emph{IEEE Transactions on PAMI},  2011,
  \textbf{33}, (2), pp.~353--367

\bibitem{Martin_a_database}
Martin, D., Fowlkes, C., Tal, D., Malik, J.
\newblock `A database of human segmented natural images and its application to
  evaluating segmentation algorithms and measuring ecological statistics'.
\newblock In: in Proceedisng of IEEE ICCV. vol.~2. (,  2001. pp.~ 416--423

\bibitem{VOC2010}
Everingham, M., Gool, L.V., Williams, C.K.I., Winn, J., Zisserman, A.. `The
  {PASCAL} {V}isual {O}bject {C}lasses {C}hallenge 2010 {(VOC2010)} {R}esults'.
  (, .
\newblock
  http://www.pascal-network.org/challenges/VOC/voc2010/workshop/index.html

\bibitem{SUND}
Xiao, J., Ehinger, K.A., Hays, J., Torralba, A., Oliva, A.: `Sun database:
  Exploring a large collection of scene categories', \emph{International
  Journal of Computer Vision},  2016, \textbf{119}, (1), pp.~3--22

\bibitem{McMains11}
McMains, S., Kastner, S.: `Interactions of top-down and bottom-up mechanisms in
  human visual cortex', \emph{The Journal of Neuroscience},  2011, \textbf{31},
  (2), pp.~587–597

\bibitem{Deng_imagenet}
Deng, J., Dong, W., Socher, R., Li, L.J., Li, K., Fei.Fei, L.
\newblock `{ImageNet: A Large-Scale Hierarchical Image Database}'.
\newblock In: Proceedings of IEEE CVPR. (,  2009.

\bibitem{Shenton}
Shenton, M.E., Turetsky, B.I.: `Understanding Neuropsychiatric Disorders:
  Insights from Neuroimaging'.
\newblock (Cambridge University Press,  2011)

\bibitem{Chalmers92}
Chalmers, D.J., French, R.M., Hofstadter, D.R.: `High-level perception,
  representation, and analogy: A critique of artificial intelligence
  methodology', \emph{Journal of Experimental \& Theoretical Artificial
  Intelligence},  1992, \textbf{4}, (3), pp.~185--211

\bibitem{Kandel}
Kandel, E.R., Schwartz, J.H., Jessell, T.M.
\newblock 26.
\newblock In: `Principles of neural science'. 4th ed. (New York, USA:
  McGraw-Hill,  2000.

\bibitem{Itti_a_model}
Itti, L., Koch, C., Niebur, E.: `A model of saliency-based visual attention for
  rapid scene analysis', \emph{IEEE Transactions on PAMI},  1998, \textbf{20},
  (11), pp.~1254--1259

\bibitem{Everingham_pascal_visual}
Everingham, M., Eslami, S.M.A., Van.Gool, L., Williams, C.K.I., Winn, J.,
  Zisserman, A.: `The {Pascal} visual object classes challenge: A
  retrospective', \emph{International Journal of Computer Vision},  2014, pp.~
  1--39

\bibitem{Engelke_comparative_study}
Engelke, U., Liu, H., Wang, J., Le.Callet, P., Heynderickx, I., Zepernick, H.,
  et~al.: `Comparative study of fixation density maps', \emph{IEEE Transactions
  on Image Processing},  2013, \textbf{22}, (3), pp.~1121--1133

\bibitem{Toet_computational_versus}
Toet, A.: `Computational versus psychophysical bottom-up image saliency: A
  comparative evaluation study', \emph{IEEE Transactions on PAMI},  2011,
  \textbf{33}, (11), pp.~2131--2146

\bibitem{Ivory_improving_web}
Ivory, M.Y., Hearst, M.A.: `Improving web site design', \emph{IEEE Internet
  Computing},  2002, \textbf{6}, (2), pp.~56--63

\bibitem{Alexe_measuring_the}
Alexe, B., Deselaers, T., Ferrari, V.: `Measuring the objectness of image
  windows', \emph{IEEE Transactions on PAMI},  2012, \textbf{34}, (11),
  pp.~2189--2202

\bibitem{Alers_how_the}
Alers, H., Bos, L., Heynderickx, I.
\newblock `How the task of evaluating image quality influences viewing
  behavior'.
\newblock In: in Proceedings of International Workshop on Quality of Multimedia
  Experience. (,  2011. pp.~ 167--172

\bibitem{Kendall}
Puka, L.
\newblock In: Lovric, M., editor. `Kendall's tau'. (Springer Berlin Heidelberg,
   2011. pp.~ 713--715

\end{thebibliography}

\section{Appendices}
	\subsection{Appendix 1. Variations in the Images and Objects of the New Dataset}
	\label{appen}
	
\begin{figure}[h]
		\newcommand{\imwidth}{0.25\columnwidth}
		\newcommand{\imheight}{0.189\columnwidth}
		\centering
		\subfloat[]{\includegraphics[width=0.44\columnwidth]{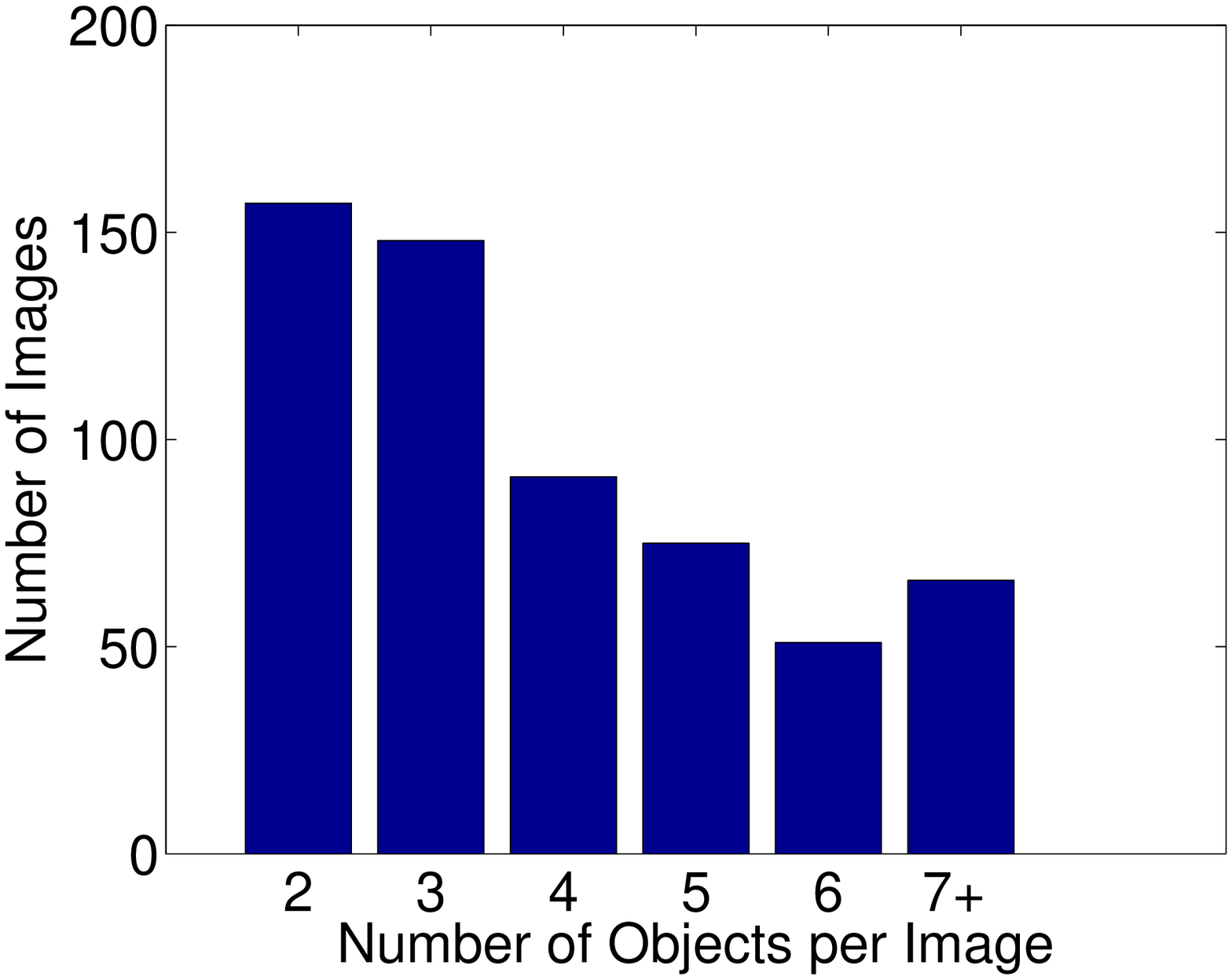}}
		\hfill
		\subfloat[]{\includegraphics[width=0.55\columnwidth]{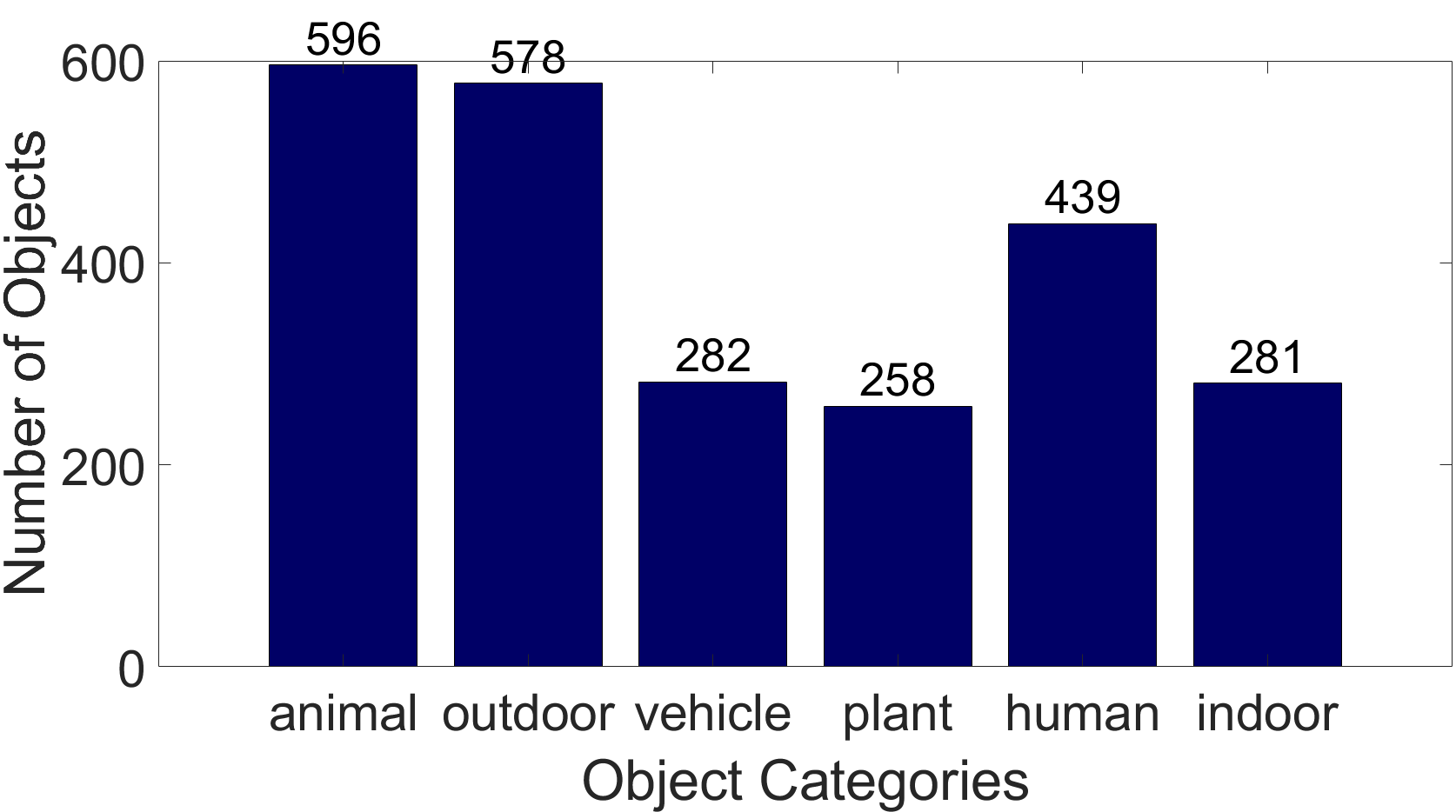}}\\
		\subfloat[]{\includegraphics[width=\imwidth,height=\imheight]{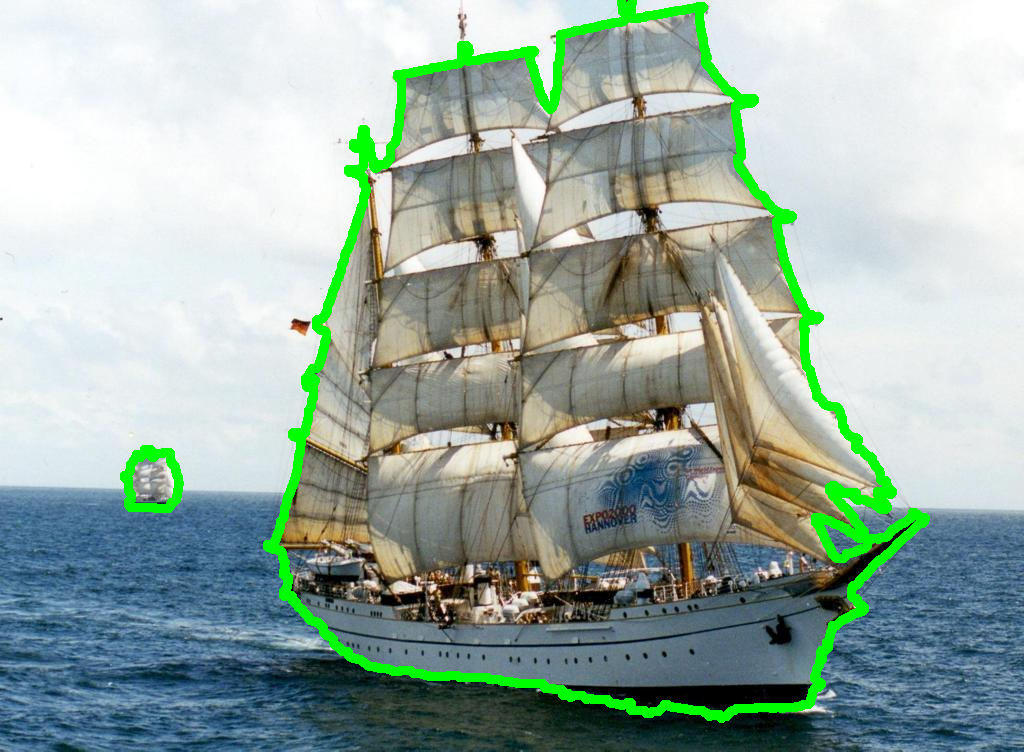}}
		\hfill
		\subfloat[]{\includegraphics[width=\imwidth,height=\imheight]{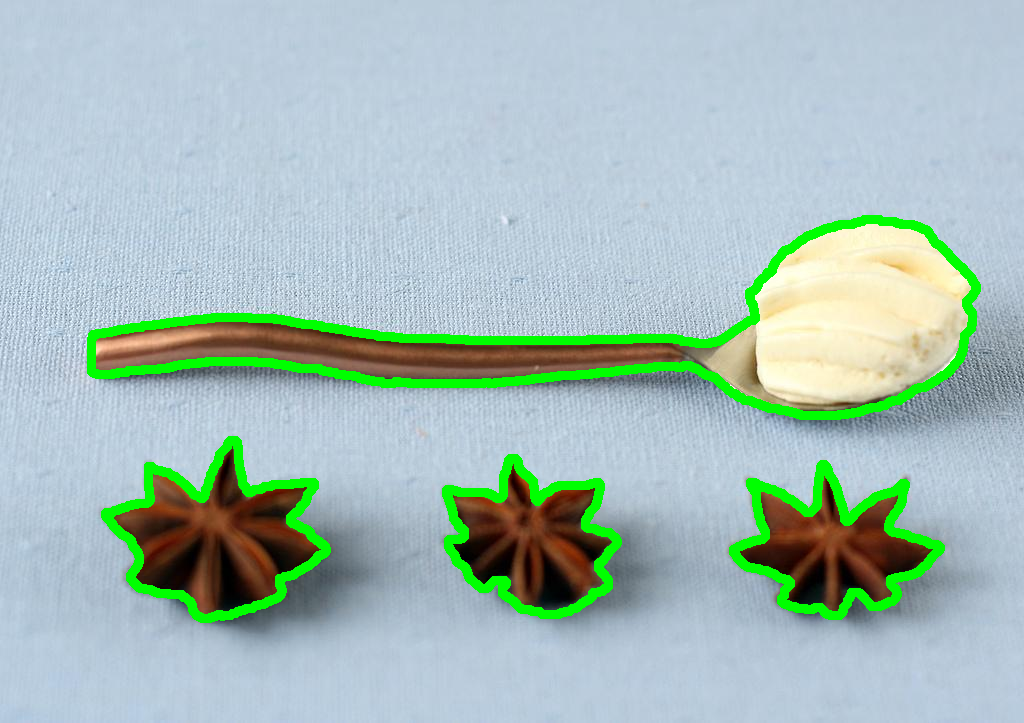}}
		\hfill
		\subfloat[]{\includegraphics[width=\imwidth,height=\imheight]{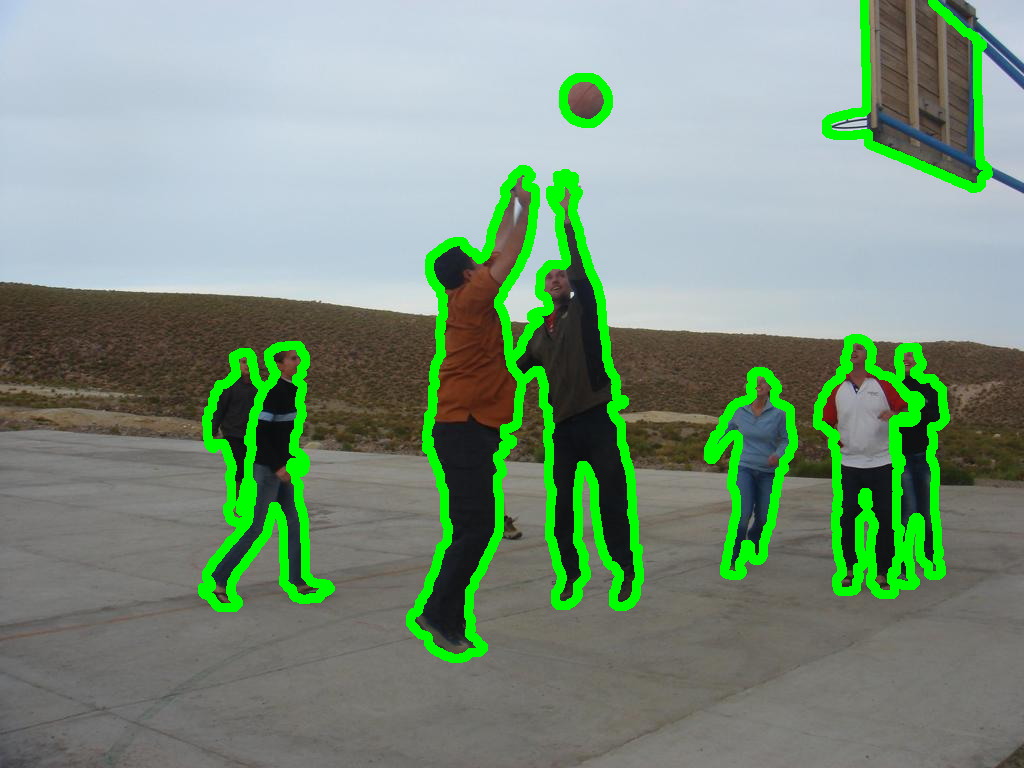}}
		\caption{(a) Number of images with their number of salient objects, {(b) object categories}, and some example images from our dataset with (c) 2, (d) 4, and (e) 7+ salient objects (Enclosed with a green border for clarity). }
		\label{ch3-numberOfObjects}
	\end{figure}

\begin{figure}[h!]
		\centering
		\newcommand{\imwidth}{0.49\columnwidth}
		\subfloat[]{\includegraphics[width=\imwidth]{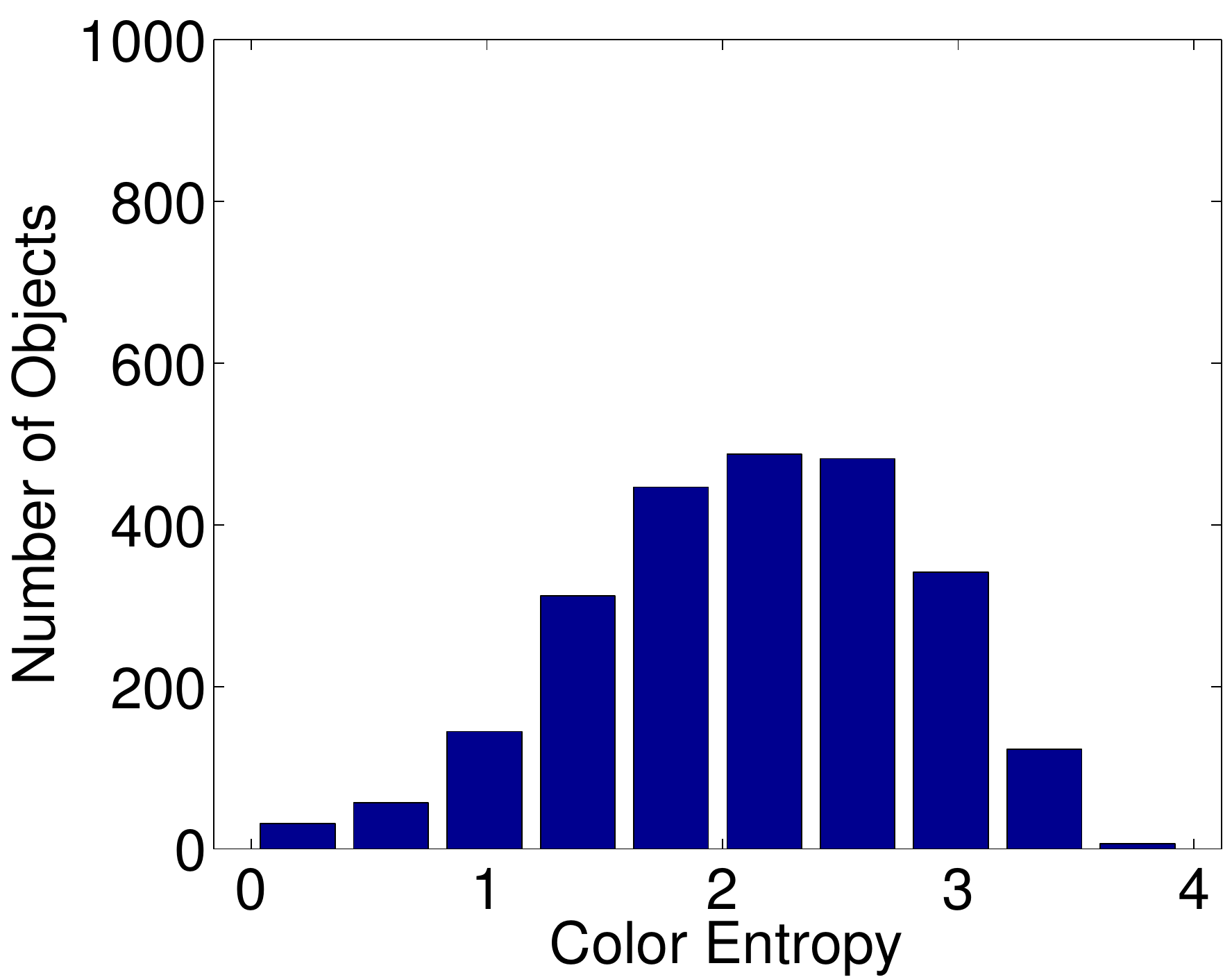}}
		\hfill
		\subfloat[]{\includegraphics[width=0.42\columnwidth]{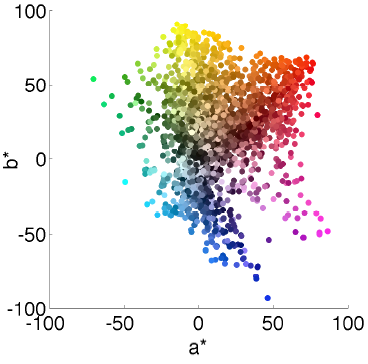}}\\
		\subfloat[]{\includegraphics[width=\imwidth]{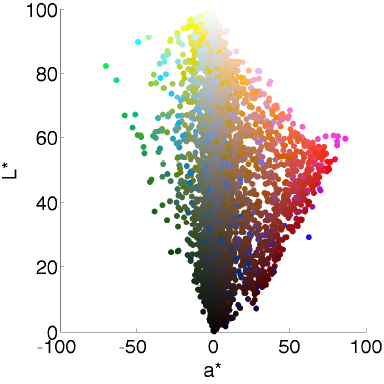}}
		\hfill
		\subfloat[]{\includegraphics[width=\imwidth]{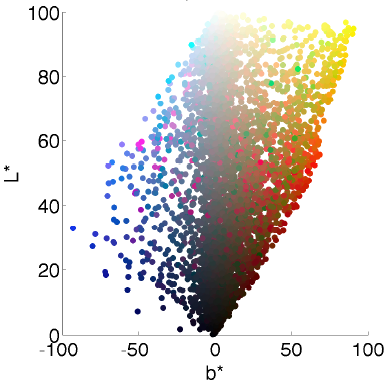}}
		\caption{(a) The distribution of color entropy of the objects in the new dataset are shown. 2D plots showing (b) a$^*$(vs)b$^*$, (c) L$^*$(vs)a$^*$ and (d) L$^*$(vs)b$^*$ distributions of object color values are also given.}
		\label{ch3-objectDistributions1}
	\end{figure}

\begin{figure}[t]
		\centering
		\newcommand{\imwidth}{0.49\columnwidth}
		\subfloat[]{\includegraphics[width=\imwidth]{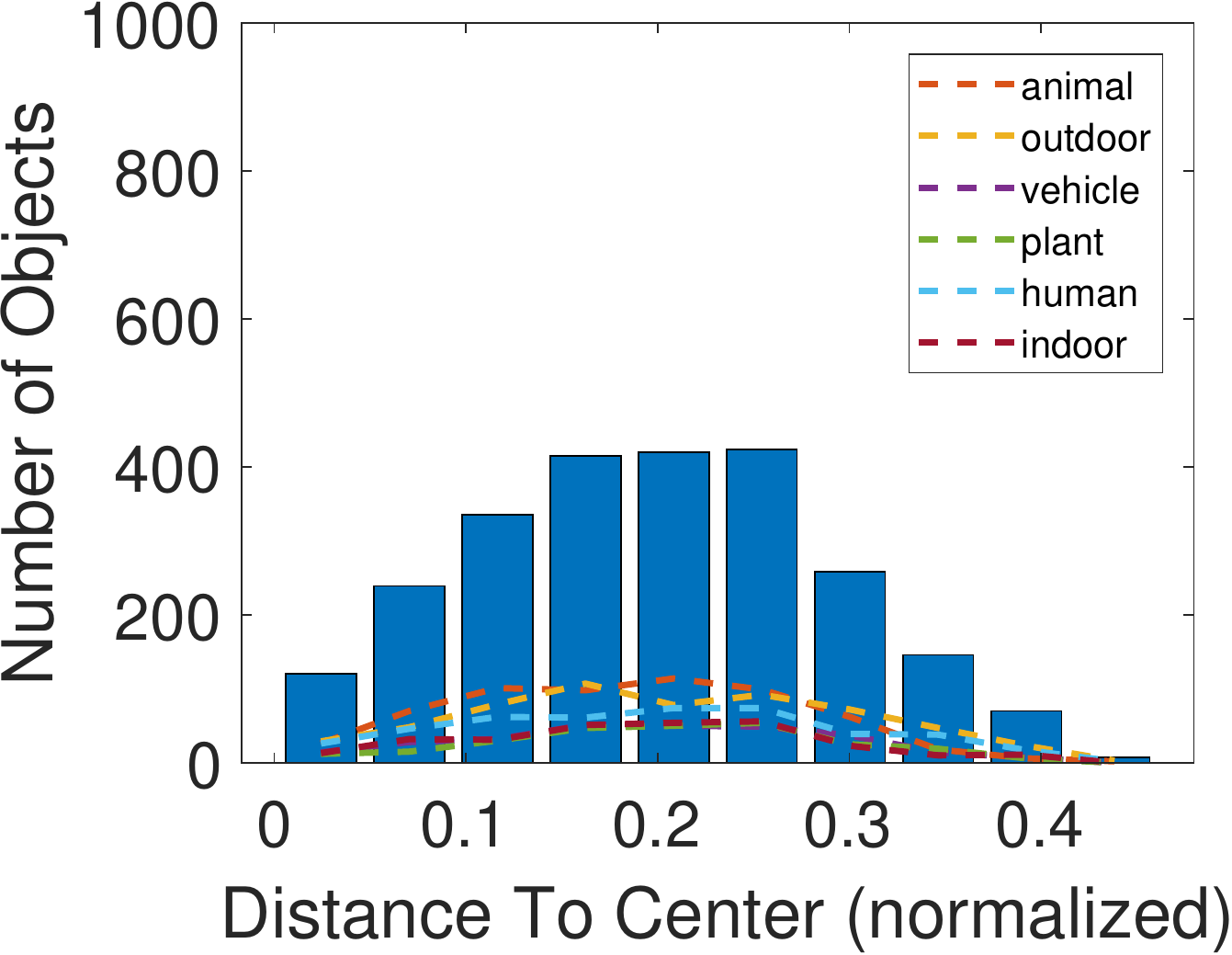}}
		\subfloat[]{\includegraphics[width=\imwidth]{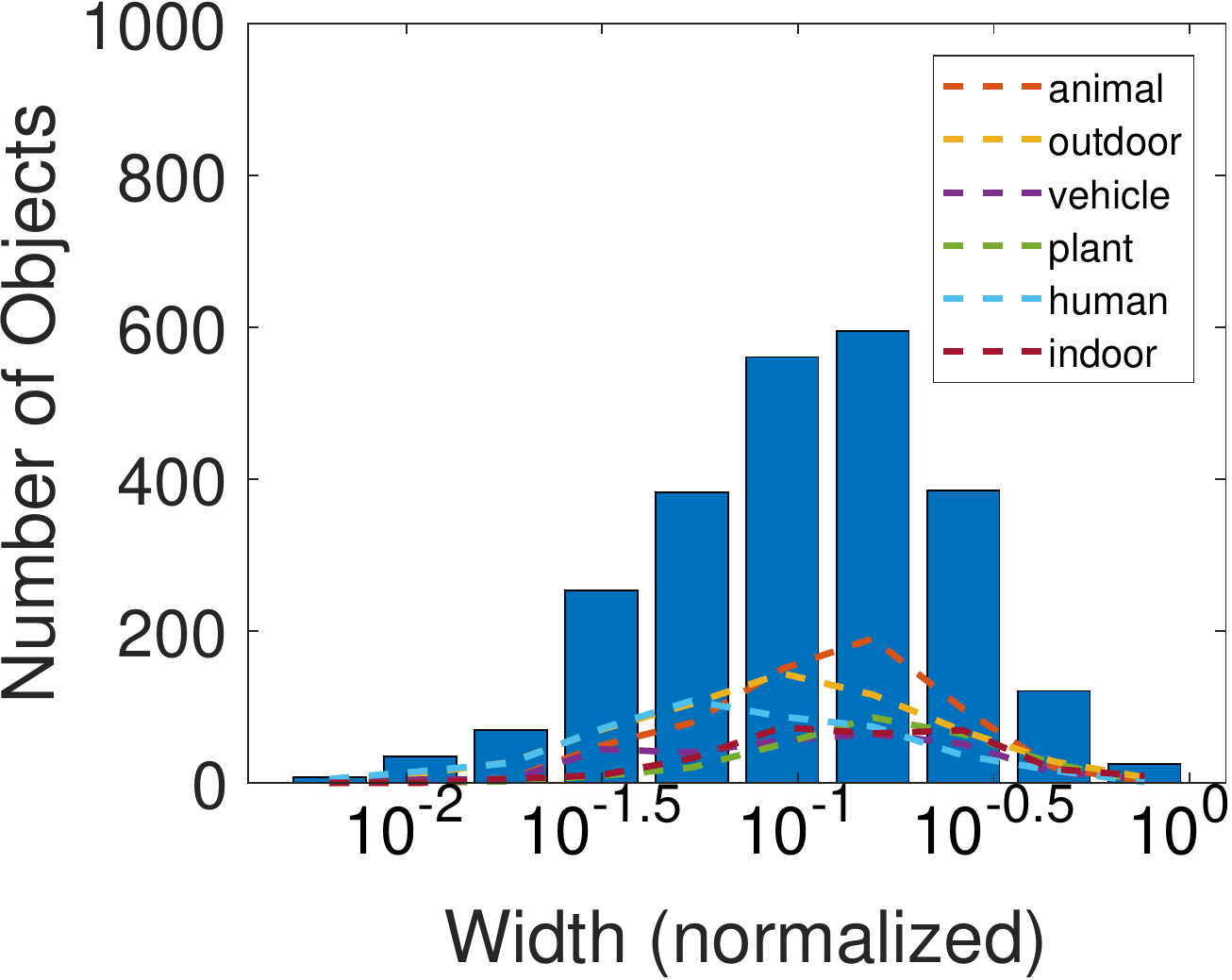}}\\
		\subfloat[]{\includegraphics[width=\imwidth]{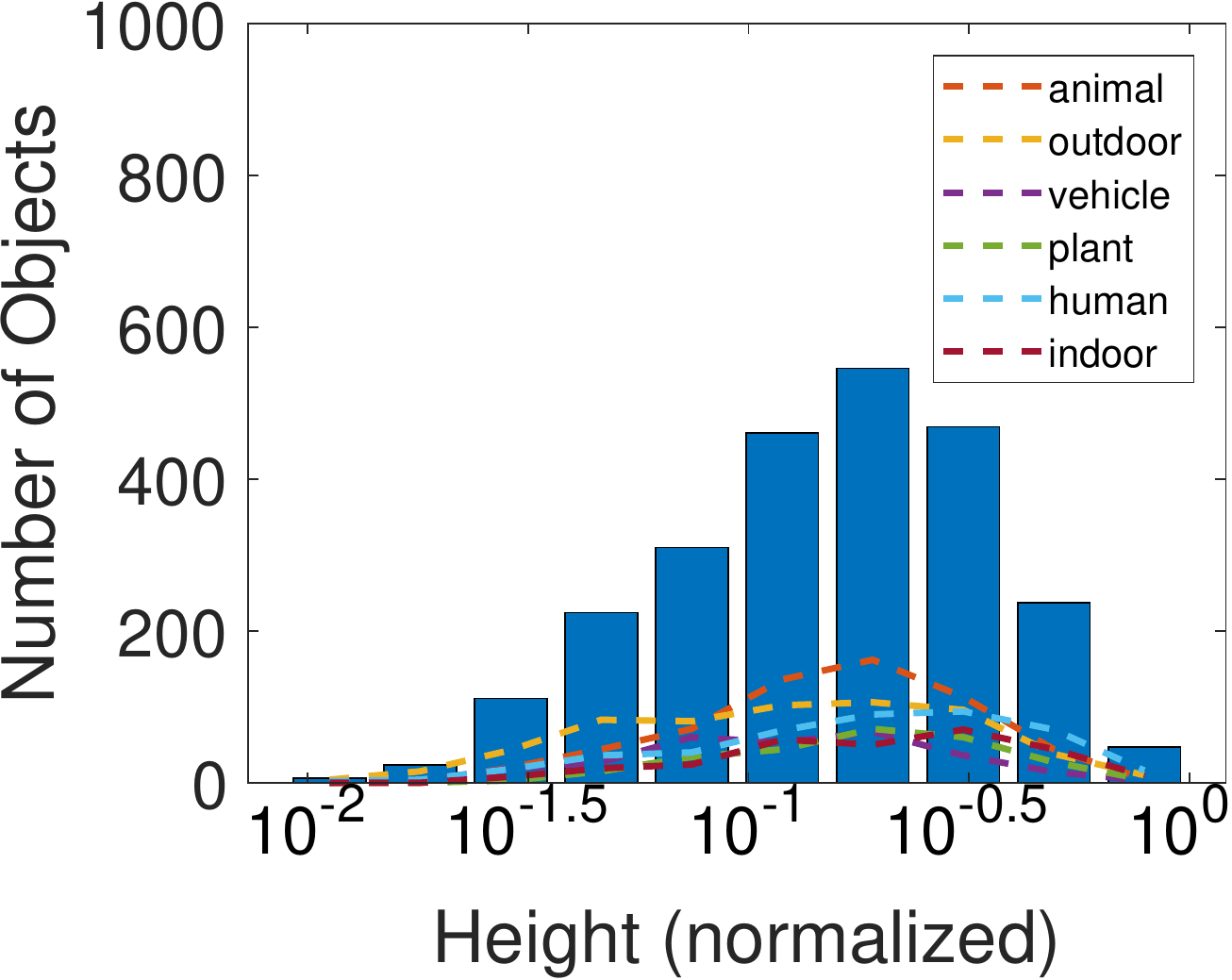}}
		\subfloat[]{\includegraphics[width=\imwidth]{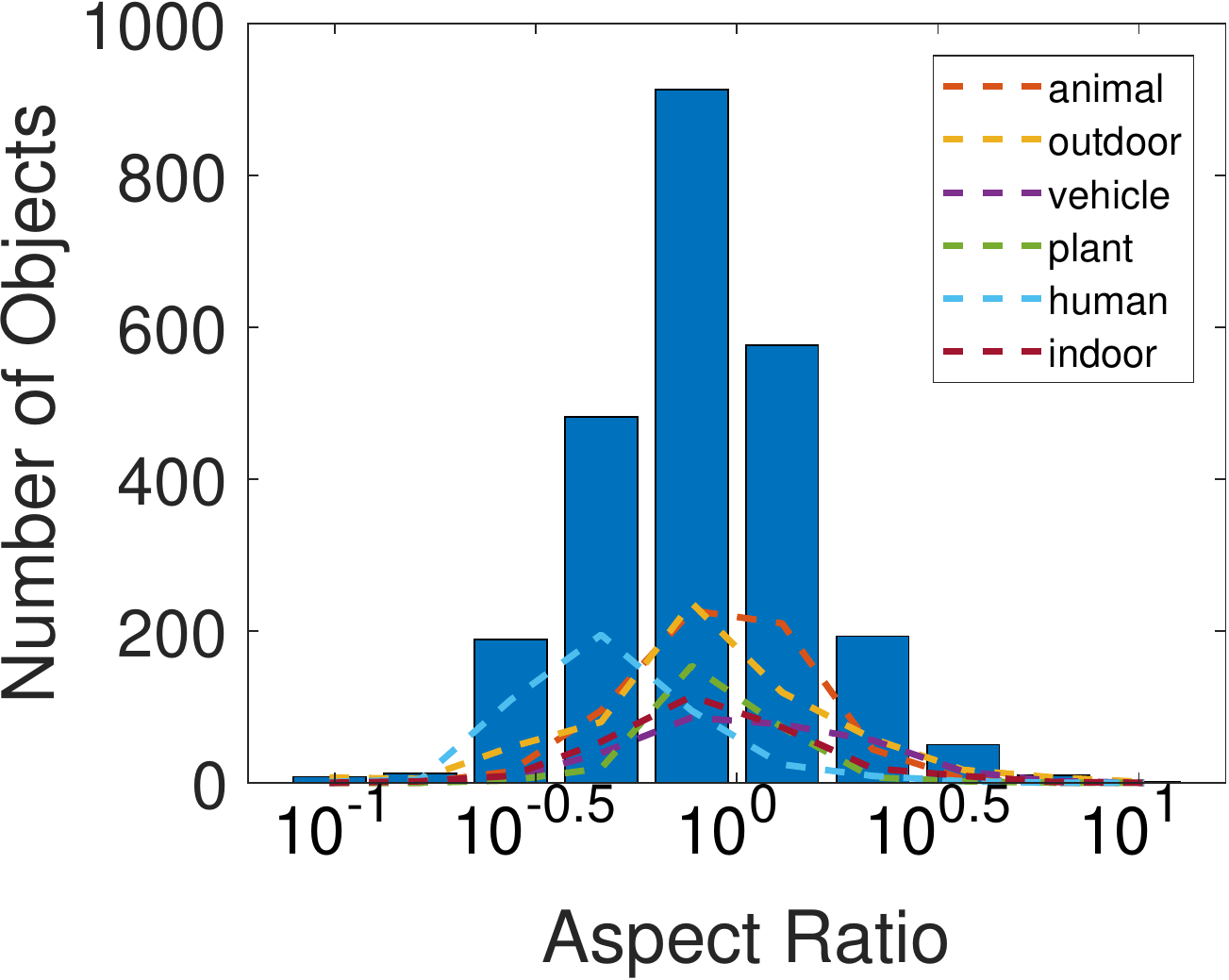}}\\
		\subfloat[]{\includegraphics[width=\imwidth]{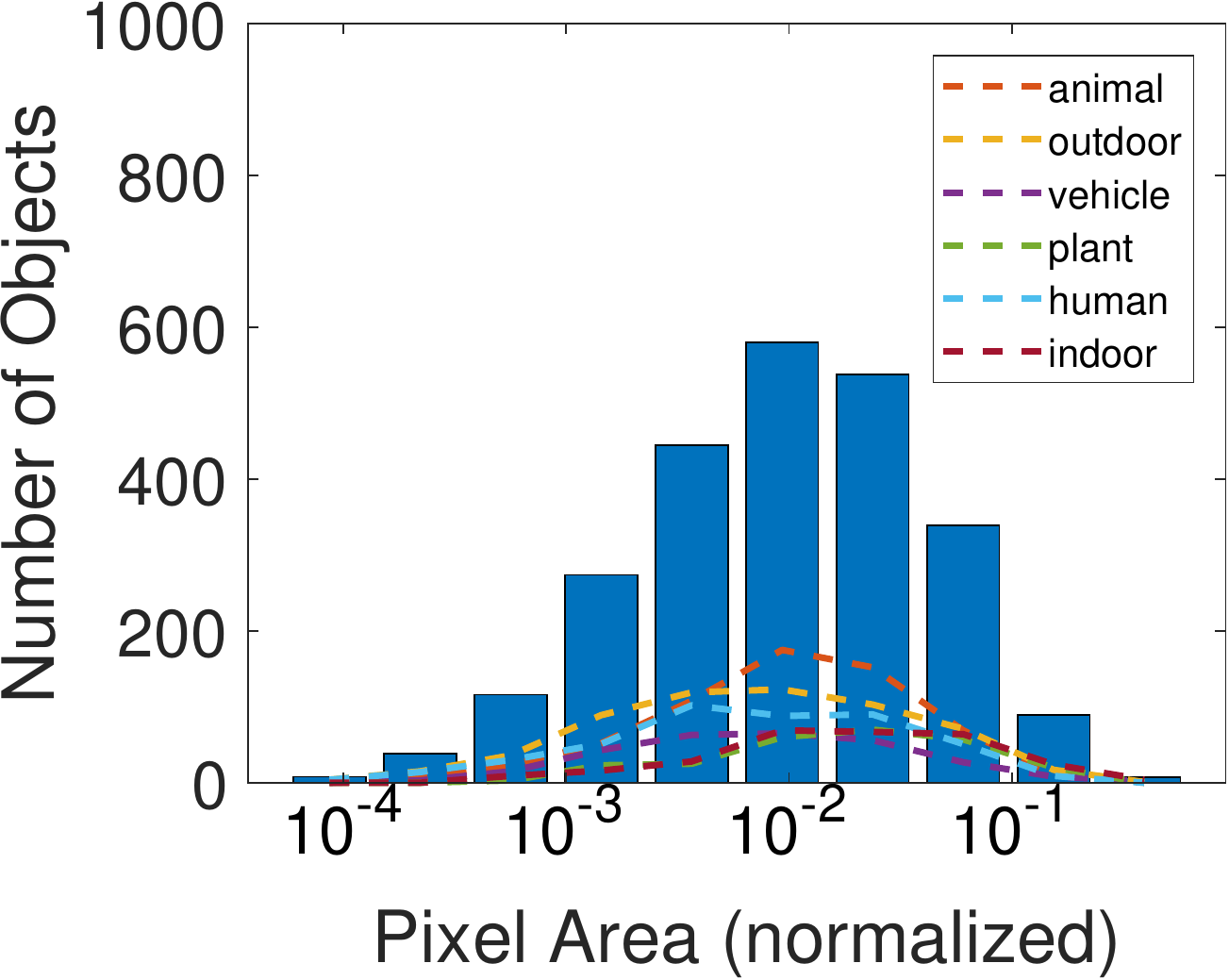}}
		\caption{Distribution of the objects in our dataset with respect to (a) object's euclidean distance to image center, (b) object width, (c) object height, (d) object aspect ratio, and (e) object area.}
		\label{ch3-objectDistributions2}
	\end{figure}

\begin{figure}[t]
		\center
		\newcommand{\imwidth}{0.49\columnwidth}
		\subfloat[]{\includegraphics[width=\imwidth]{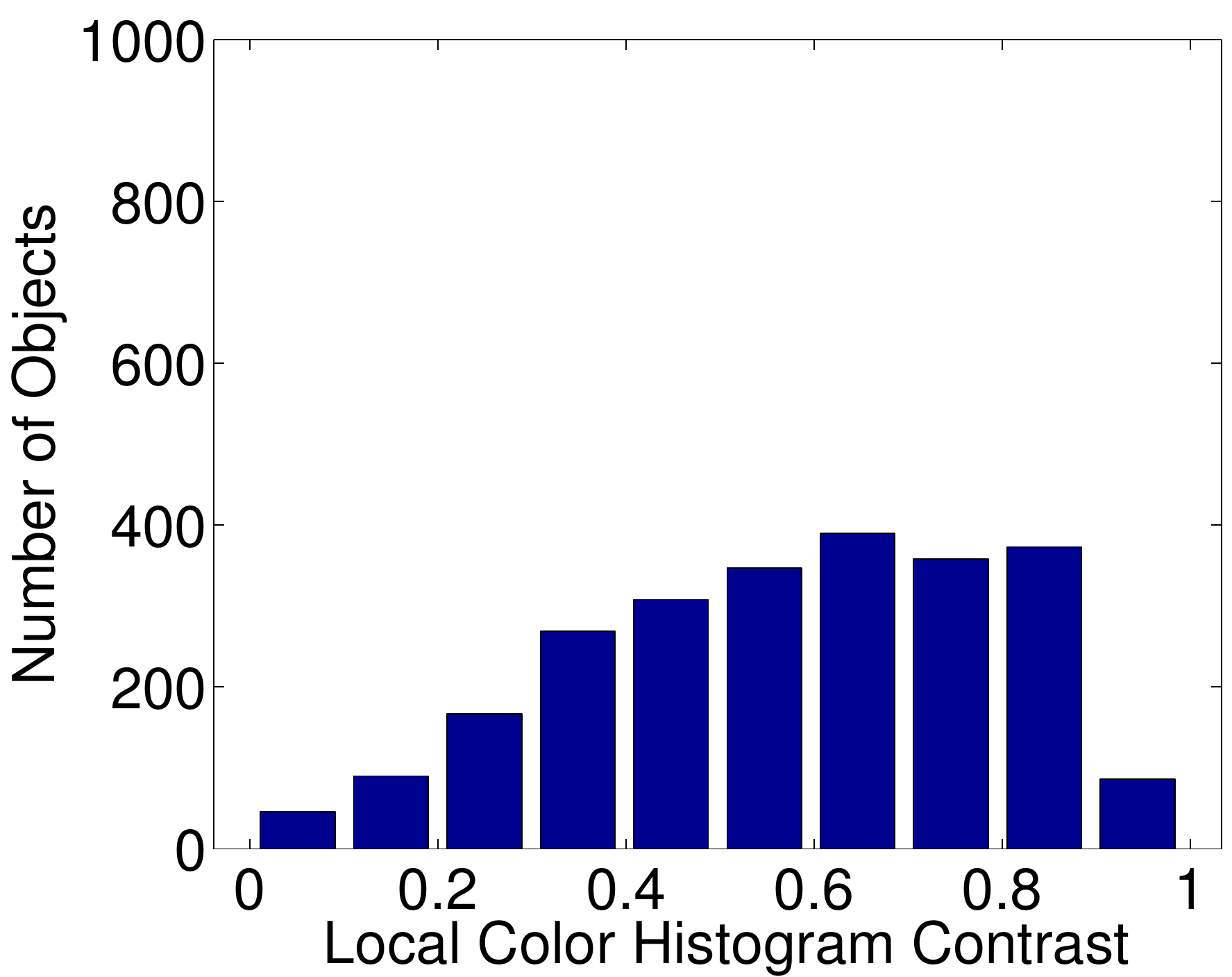}}
		\hfill
		\subfloat[]{\includegraphics[width=\imwidth]{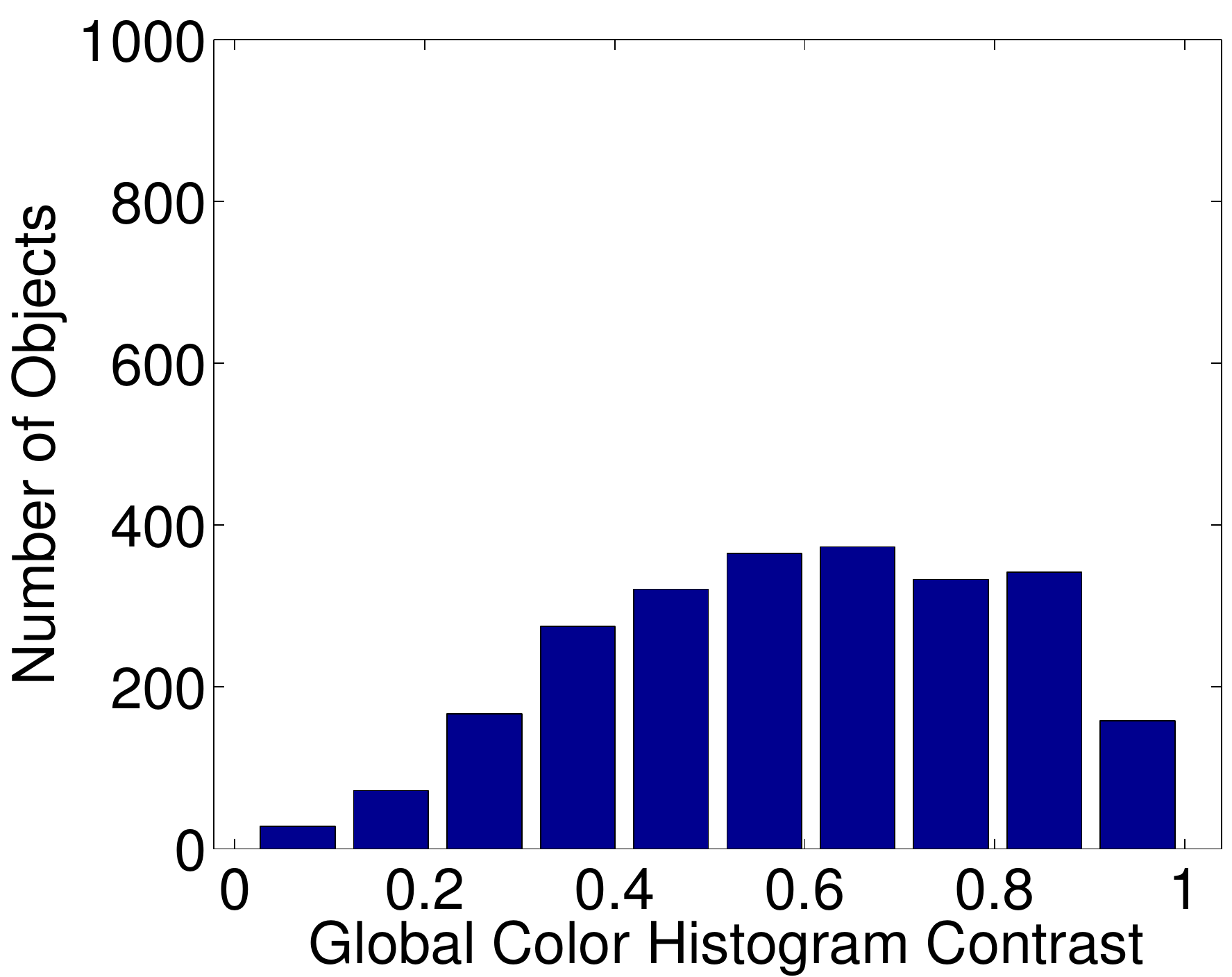}}\\
		\subfloat[Local neighborhood mask ($\mathbf{M}_l^o$)]{\frame{\includegraphics[width=\imwidth]{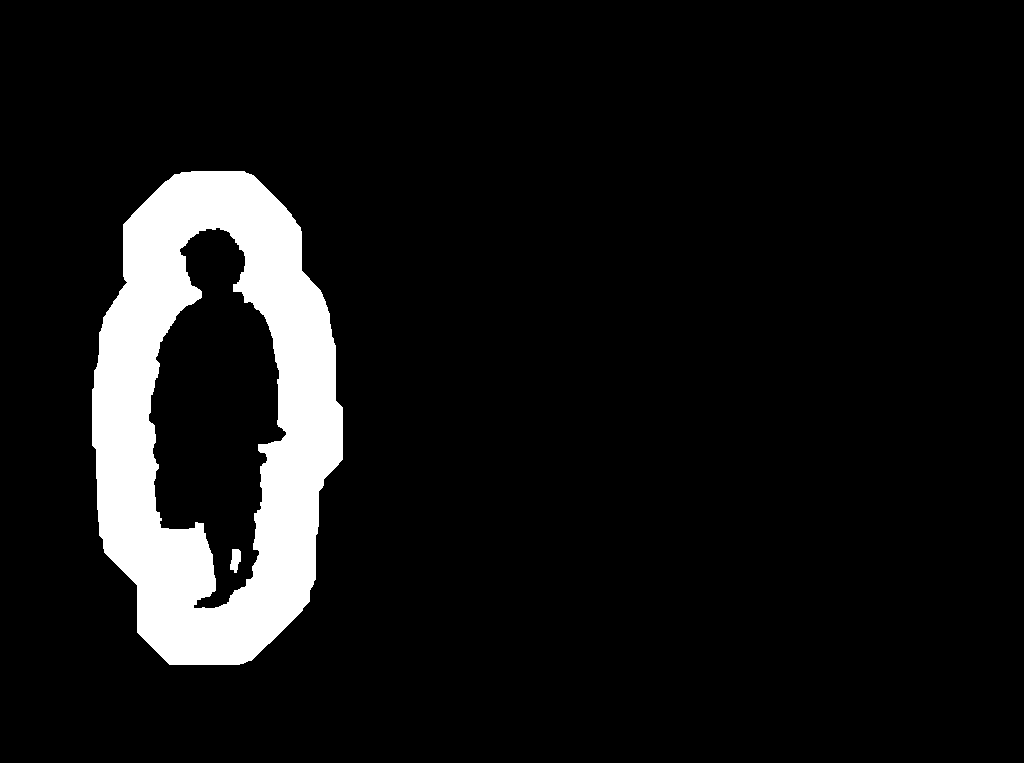}}}
		\hfill
		\subfloat[Global background mask ($\mathbf{M}_g^o$)]{\frame{\includegraphics[width=\imwidth]{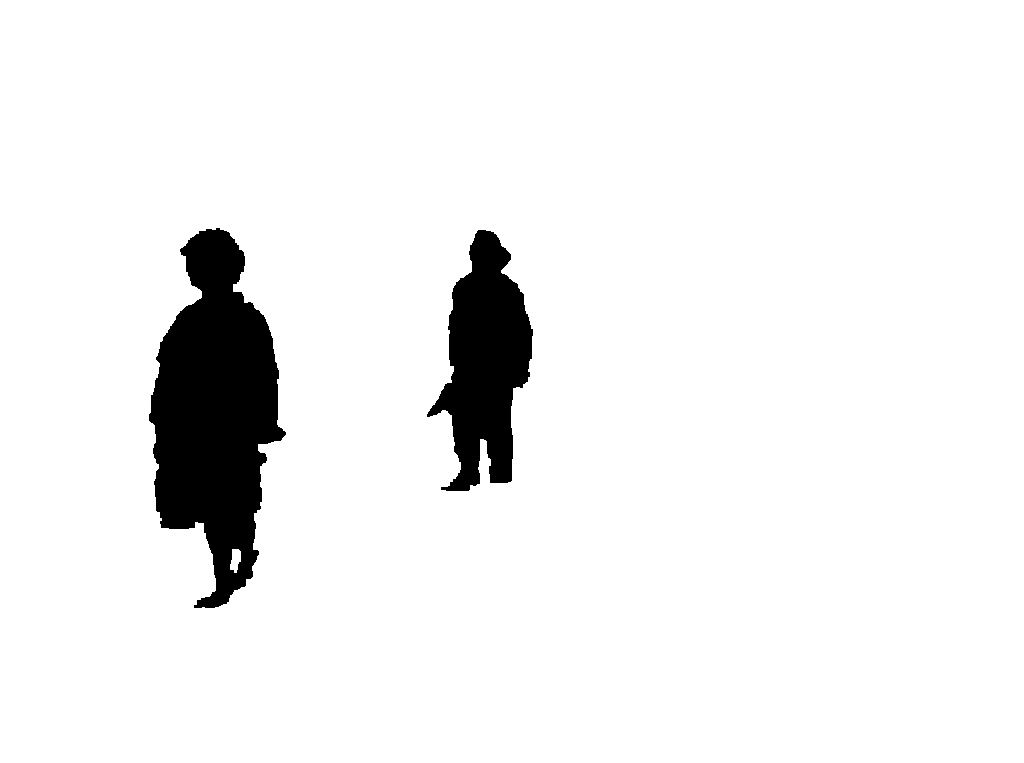}}}
		\caption{{The distributions of (a) local and (b) global color contrasts. Color contrast of an object (see Figure~\ref{fig:averageColorCalculation}) is computed using local and global pixel neighborhoods visualized in (c) and (d), respectively.}
		}
		\label{ch3-averageColorCalculation}
	\end{figure}

	In Section~III of the paper, we mentioned that our dataset is balanced ensuring sufficient variations among images in terms of the number of objects, the object sizes and positions, the object color content and the color-contrast between the object and surroundings (local and global). Here, in Figures~\ref{ch3-numberOfObjects},~\ref{ch3-objectDistributions1},~\ref{ch3-objectDistributions2} and~\ref{ch3-averageColorCalculation}, we illustrate the said characteristics of our dataset in terms of several related parameters.
	
	Consider Figure~\ref{ch3-numberOfObjects}. Figure~\ref{ch3-numberOfObjects}(a) gives the distribution of the images in our dataset in terms of the number of salient objects contained in them. It is evident that the images in our dataset have a substantially varied number of objects, which is central to the discussions in the paper. {Figure~\ref{ch3-numberOfObjects}(b) gives the distribution of different salient object categories in our dataset, where the terms `outdoor' and `indoor' stands for other generic objects found outdoors and indoors, respectively.} Figures~\ref{ch3-numberOfObjects}(c), (d) and (e) display some example images from our dataset with the salient objects marked.
	
	In Figure~\ref{ch3-objectDistributions1}, we demonstrate the variedness of the objects in our dataset with respect to their color contents. Figure~\ref{ch3-objectDistributions1}(a) gives the distribution of the objects in our dataset in terms of their color entropies. Obviously, color entropies quantify within-object color variety. To compute color entropy of an object $o$ the CIE L$^*$a$^*$b$^*$ values at pixels within its segmentation mask $\mathbf{M}^o$ are considered. Segmentation mask formation for an object is described in Section~\ref{obsegmask}. The distribution of the object color values in the 3 dimensional L$^*$a$^*$b$^*$ space is then used to compute its entropy. It is evident from Figure~\ref{ch3-objectDistributions1}(a) that the objects in our dataset vary substantially in within-object color variety. The 2D plots in Figure~\ref{ch3-objectDistributions1}(b-d) show the  L$^*$, a$^*$ and b$^*$ values of the objects distributed with respect to each other. Here, an object is represented by a single color value, which is the average of all the color vectors at pixels within the object's segmentation mask $\mathbf{M}^o$. So, an element in the 2D plots corresponds to an object and gives its average 2D color value. As can be seen from all the three figures, significant average color variation exists between the objects of the dataset.
	
\begin{figure}[h]
	\centering
	\newcommand{\imwidth}{0.95\columnwidth}
\includegraphics[width=\imwidth]{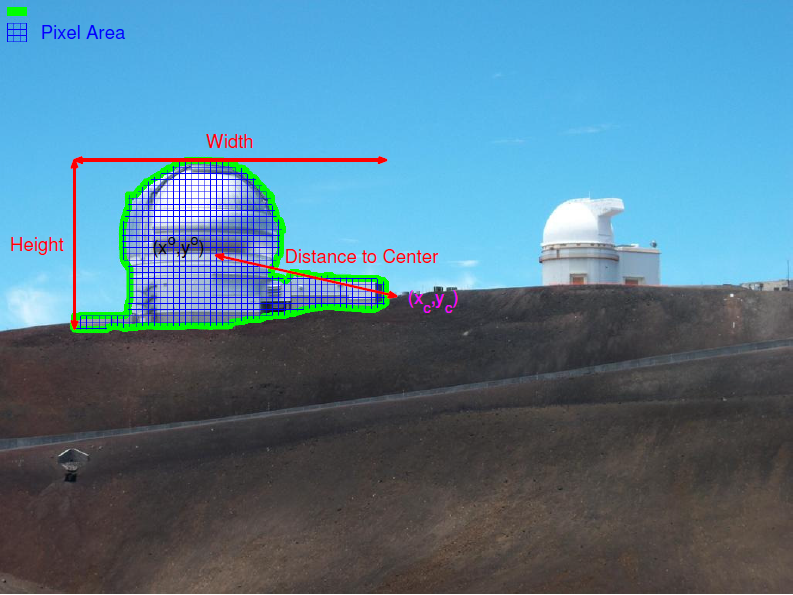}
	\caption{Depiction of computation of various parameters related to an object's size and location in the image.}
	\label{param}
\end{figure}

	In Figure~\ref{ch3-objectDistributions2}, we demonstrate the variedness of the objects in our dataset with respect to their positions in images, and their shapes and sizes in terms of various parameters. {The objects are also categorized into six classes, namely, animal, plant, human, vehicle, other outdoor and indoor objects, and category-wise variedness with respect to the parameters are also presented.}	To understand the parameters, consider Figure~\ref{param}. Figure~\ref{ch3-objectDistributions2}(a) gives the distribution of the objects in our dataset in terms of their Euclidean distances to the image centers normalized with respect to underlying images' diagonal lengths. Figures~\ref{ch3-objectDistributions2}(b) and (c) respectively show the distributions of the objects in our dataset in terms of their widths and heights, which are computed as shown in the example of Figure~\ref{param}. Note that, the widths and heights have been computed in terms of the number of pixels, and then normalized with respect to their maximum values in the dataset. Figure~\ref{ch3-objectDistributions2}(d) shows the distribution of the objects in our dataset in terms of their aspect ratios, which are calculated as the ratio of width and height. Finally, Figure~\ref{ch3-objectDistributions2}(e) shows the distribution of the objects in our dataset in terms their areas, which are the number of pixels in the irregular object regions, normalized by the maximum value in the dataset. It is evident from these figures that the objects in our dataset vary substantially in terms of size and image position. {It is also evident that the said observation is also valid if we consider the object category-wise distributions shown in Figure~\ref{ch3-objectDistributions2}(a)-(e).}
		
	In Figure~\ref{ch3-averageColorCalculation} the variedness in contrasts of the objects in our dataset from their local neighborhood and global background is demonstrated. For an object in an image within a region given by the object segmentation mask $\mathbf{M}^o$, regions of local neighborhood and global background in the image are defined. For example, the local neighborhood mask $\mathbf{M}_l^o$ and the global background mask $\mathbf{M}_g^o$ for the object mask $\mathbf{M}^o$ in Figure~\ref{fig:averageColorCalculation}(b) corresponding to an object in the image of Figure~\ref{fig:averageColorCalculation}(a) are shown in Figures~\ref{ch3-averageColorCalculation}(a) and (b), respectively. The local and global contrasts are then calculated as the $\chi^2$-distances between the distributions of the L$^*$a$^*$b$^*$ color values at the pixels within the object segmentation mask, and the local neighborhood and the global background masks, respectively. The distribution of the objects in our dataset in terms of the said distance values are shown in Figures~\ref{ch3-averageColorCalculation} (c) and (d) with the former depicting the global contrast distribution and the latter local contrast distribution. As can be seen, significant variations in local and global contrasts exist between different objects in our dataset.
	
\end{document}